%% file: main.tex
\documentclass[journal]{IEEEtran}

\input{sections/0-preamble.tex}

\input{sections/0-notation.tex}

\begin{document}

\title{6DLS: Modeling Nonplanar Frictional Surface Contacts for Grasping using 6D Limit Surfaces}
\author{Jingyi~Xu,~\IEEEmembership{Student~Member,~IEEE,}
        Tamay~Aykut,
        Daolin~Ma,~\IEEEmembership{Member,~IEEE,}
        and~Eckehard~Steinbach,~\IEEEmembership{Fellow,~IEEE}
\thanks{J. Xu, T. Aykut, and E. Steinbach are with the Chair of Media Technology, Department of Electrical and Computer Engineering, Technical University of Munich. E-mail: \{jingyi.xu,tamay.aykut,eckehard.steinbach\}@tum.de}
\thanks{D. Ma is with the Manipulation and Mechanism Lab, Department of Mechanical Engineering, Massachusetts Institute of Technology. E-mail: daolinma@mit.edu}}

\markboth{}%
{ }

\maketitle
\copyrightnotice 

\input{sections/1-abstract}


\IEEEpeerreviewmaketitle

\section{Introduction}
\input{sections/2-introduction}

\section{Related Work}
\input{sections/3-related-work}

\input{sections/5-problem-statement}

\section{Contact Wrench for a Nonplanar Surface Contact}
\label{sec:friction_computation}

\input{sections/6-friction-param-surface}

\input{sections/8-limit-surface}

\section{Prediction of Multicontact Grasp Success}
\label{sec:multi_contacts}
\input{sections/9-multi-contacts}

\input{sections/11-simulations}

\input{sections/12-experiments}

\section{Discussion}
\input{sections/13-discussion}

\section{Conclusion}
\input{sections/14-conclusion}

\section*{Acknowledgment}
\input{sections/15-acknowledgement}

\ifCLASSOPTIONcaptionsoff
  \newpage
\fi


\bibliographystyle{IEEEtran}
\bibliography{myabrv, IEEEfull,main}

\input{sections/16-biography}



\appendices

\section{Frictional Wrench for a Planar Surface}
\label{appendix:background}
\input{sections/17-background}

\section{Frictional Wrench for a Discrete Nonplanar Surface}
\label{appendix:friction_discrete}
\input{sections/7-friction-discrete-surface}

\section{Selecting the Local Contact Frame for Contact Wrench Computation}
\label{sec:app_local}
\input{sections/20-wrench-example}

\input{sections/10-data-acquisition}

\end{document}

%% file: sections/0-preamble.tex
\usepackage[printwatermark]{xwatermark}
\usepackage{subfiles}
\usepackage{cite}
\usepackage{graphicx}
\usepackage{amssymb}
\usepackage{mathtools}
\usepackage[section]{placeins}
\usepackage{tabularx}
\usepackage{array}
\usepackage{relsize}
\usepackage{flushend} 
\usepackage{balance}

\usepackage{algorithm}
\usepackage{algorithmic}
\usepackage{array}
\usepackage[caption=false]{subfig}
\usepackage{float}
\usepackage{url}
\usepackage{xcolor}

\usepackage[utf8]{inputenc}
\usepackage{amsfonts}
\usepackage{xfrac}
\usepackage{fix-cm}  
\usepackage{wrapfig}
\usepackage{tikz}
\usetikzlibrary{calc,arrows,positioning,matrix,fit,shapes,tikzmark,scopes} 
\usepackage{tikz-3dplot}
\usepackage{pgfplots}
\pgfplotsset{compat=newest}
\usepackage{makecell}
\usepackage{multirow}
\usepackage{booktabs}
\usepackage{siunitx}
\usepackage{esint}

\usepackage{tikz}
\newcommand\copyrighttext{\footnotesize The paper will appear in the \emph{IEEE Transactions on Robotics}.  \textcopyright~\the\year~IEEE. Personal use of this material is permitted. Permission from IEEE must be obtained for all other uses, in any current or future media, including reprinting/republishing this material for advertising or promotional purposes, creating new collective works, for resale or redistribution to servers or lists, or reuse of any copyrighted component of this work in other works.}
\newcommand\copyrightnotice{%
	\begin{tikzpicture}[remember picture,overlay]
	\node[anchor=north,yshift=-15pt] at (current page.north) {\fbox{\parbox{\dimexpr0.97\textwidth-\fboxsep-\fboxrule\relax}{\copyrighttext}}};
	\end{tikzpicture}
	\vspace{-0.3cm}
}

\def\myfinal{yep}  
\newif\ifremove   
\removetrue     

\def\mysum#1{\ifx\myfinal\undefined\textcolor{gray}{Summary: #1}\par\fi}

\def\mydo#1{\ifx\myfinal\undefined\textcolor{blue}{#1}\par\fi}
\def\mydel#1{\ifx\myfinal\undefined\textcolor{red}{#1}\par\fi}

\def\sref#1{Section~\ref{#1}}
\def\aref#1{Appendix~\ref{#1}}
\def\fref#1{Figure~\ref{#1}}
\def\eref#1{Equation~\eqref{#1}}
\def\eqsref#1{Eq.~\eqref{#1}}
\def\tref#1{Table~\ref{#1}}

\newlength\w
\newlength\h

\newcommand{\eq}{Equation}

\definecolor{ceditor}{RGB}{0,0,0}

\DeclarePairedDelimiterX{\norm}[1]{\lVert}{\rVert}{#1}

\newcommand*\diff{\mathop{}\!\mathrm{d}}
\newcount\colveccount
\newcommand*\colvec[1]{
	\global\colveccount#1
	\begin{pmatrix}
		\colvecnext
	}
	\def\colvecnext#1{
		#1
		\global\advance\colveccount-1
		\ifnum\colveccount>0
		\\
		\expandafter\colvecnext
		\else
	\end{pmatrix}
	\fi
}

\makeatletter
\@for\next:={int,iint,iiint,iiiint,dotsint,oint,oiint,sqint,sqiint,
	ointctrclockwise,ointclockwise,varointclockwise,varointctrclockwise,
	fint,varoiint,landupint,landdownint}\do{%
	\expandafter\edef\csname\next\endcsname{%
		\noexpand\DOTSI
		\expandafter\noexpand\csname\next op\endcsname
		\noexpand\ilimits@
	}%
}
\makeatother

\newcommand{\ldotsc}{,\ldots,}

\makeatletter
\newcommand*\bigdot{\mathpalette\bigcdot@{.5}}
\newcommand*\bigcdot{\cdot}
\newcommand*\bigcdot@[2]{\mathbin{\vcenter{\hbox{\scalebox{#2}{$\m@th#1\bullet$}}}}}
\makeatother

\def\V#1{\ensuremath{\mathbf{#1}}}
\def\mym#1{\ensuremath{#1}}

\def\sX{\@ifnextchar[{\@sXFrame}{\mym{\V{X}}}}



%% file: sections/0-notation.tex
\newcommand{\R}{\mathbb{R}}
\def\R#1{\mathbb{R}^{#1}}
\renewcommand{\vec}[1]{\boldsymbol{{#1}}} 
\newcommand{\mat}[1]{\boldsymbol{{#1}}}

\newcommand{\trans}{{{\mathsmaller T}}}

\newcommand{\uv}{\left(u,v\right)}
\newcommand{\dudv}{\diff u \diff v}

\newcommand{\U}{\mathcal{U}}

\newcommand{\twistnobold}{{\xi}}
\newcommand{\twist}{\vec{\twistnobold}}
\newcommand{\unittwist}{\hat{\vec{\twist}}}
\newcommand{\lineISA}{l}
\newcommand{\linedir}{\vec{e}}
\newcommand{\linepoint}{\vec{q}}
\newcommand{\moment}{\vec{m}}
\newcommand{\vel}{{\vec{v}}}
\newcommand{\vpara}{{\vec{v}}_{\parallel}}
\newcommand{\vperp}{{\vec{v}}_{\perp}}
\newcommand{\angular}{{\vec{\omega}}}
\newcommand{\normang}{{\norm{\vec{\omega}}}}
\newcommand{\triplet}{\left(\linedir, \linepoint, h\right)}
\newcommand{\tripletmoment}{\left(\linedir, \moment, h\right)}

\newcommand{\wrenchresmp}{\hat{\vec{s}}^\text{LS}}
\newcommand{\verte}{\vec{s}^{\text{cube}}}
\newcommand{\wrenchresmpden}{\vec{s}^{\text{LS}}}

\newcommand{\pressure}{p}

\newcommand{\pressureunit}{\hat{\pressure}}
\newcommand{\boldf}{\vec{f}}
\newcommand{\dfperpfunc}{\diff\fperp\hspace{-0.2em}\uv}
\newcommand{\boldtau}{\vec{\tau}}
\newcommand{\arm}{\vec{r}}

\newcommand{\wrenchnobold}{{w}}
\newcommand{\wrench}{\vec{\wrenchnobold}}
\newcommand{\wrencharb}{\vec{\wrenchnobold}^{*}}

\newcommand{\normalizedwrench}{\hat{\wrench}}
\newcommand{\normalwrench}{\wrench_{\perp}}
\newcommand{\fperp}{{\vec{f}_\perp}}
\newcommand{\tauperp}{{\vec{\tau}_\perp}}

\newcommand{\origin}{{\vec{o}}}
\newcommand{\originx}{{{o}_x}}
\newcommand{\originy}{{{o}_y}}
\newcommand{\originz}{{{o}_z}}
\newcommand{\boldsigma}{\vec{\sigma}}

\newcommand{\surface}{\mathcal{S}}
\newcommand{\tangent}{T_{\surface}}

\newcommand{\velfunc}{{\vec{v}}\uv}
\newcommand{\vrfunc}{\hat{\vec{v}}_r\uv}
\newcommand{\vr}{\hat{\vec{v}}_r}

\newcommand{\normalfunc}{\vec{n}\uv}
\newcommand{\dS}{\diff\surface}

\newcommand{\normal}{\vec{n}}
\newcommand{\boldomega}{\vec{\omega}}

\newcommand{\centeri}{\vec{s}_{i}}
\newcommand{\normali}{\normal_{i}}
\newcommand{\vri}{\hat{\vec{v}}_{r_i}}

\newcommand{\areai}{a_{i}}
\newcommand{\pressurei}{\pressure_{i}}

\newcommand{\x}{\boldsymbol{x}}
\newcommand{\xnb}{x}

\newcommand{\A}{\mat{A}}
\newcommand{\Ae}{\A_e}
\newcommand{\Aq}{\A_q}
\newcommand{\fe}{f_1}
\newcommand{\fq}{f_2}

\newcommand{\nprs}{K}
\newcommand{\psd}{\mat{M}}

\newcommand{\W}{\mathcal{W}}

\newcommand{\pairsp}{\left(\surface,\pressureunit \right)}

\newcommand{\X}{\mathcal{X}}

\newcommand{\wext}{\vec{w}_\text{ext}}

\newcommand{\contactw}{\vec{c}}

\newcommand{\D}{\mathcal{D}}

\newcommand{\Area}{\mathcal{A}}
\newcommand{\dA}{\diff\Area}

%% file: sections/1-abstract.tex
\begin{abstract}
	Robot grasping with deformable gripper jaws results in nonplanar surface contacts if the jaws deform to the nonplanar local geometry of an object.
	The frictional force and torque that can be transmitted through a nonplanar surface contact are both three-dimensional, resulting in a six-dimensional frictional wrench (6DFW). 
	Applying traditional planar contact models to such contacts leads to over-conservative results as the models do not consider the nonplanar surface geometry and only compute a three-dimensional subset of the 6DFW.
	To address this issue, we derive the 6DFW for nonplanar surfaces by combining concepts of differential geometry and Coulomb friction. 
	We also propose two 6D limit surface (6DLS) models, generalized from well-known three-dimensional LS (3DLS) models, which describe the friction-motion constraints for a contact.
	We evaluate the 6DLS models by fitting them to the 6DFW samples obtained from six parametric surfaces and 2,932 meshed contacts from finite element method simulations of 24 rigid objects. 
	We further present an algorithm to predict multicontact grasp success by building a grasp wrench space with the 6DLS model of each contact.
	To evaluate the algorithm, we collected 1,035 physical grasps of ten 3D-printed objects with a KUKA robot and a deformable parallel-jaw gripper.
	In our experiments, the algorithm achieves 66.8\% precision, a metric inversely related to false positive predictions, and 76.9\% recall, a metric inversely related to false negative predictions.
	The 6DLS models increase recall by up to 26.1\% over 3DLS models with similar precision\footnote{Code and dataset are available at \url{https://github.com/martinajingyixu/non-planar-surface-contact}. We also provide a video summarizing the algorithm at https://youtu.be/mBS30kPqrw4}{}.
\end{abstract}

%% file: sections/2-introduction.tex
\mysum{why need friction analysis for grasping and planar pushing }

\IEEEPARstart{R}{obot} grasping remains an active area of research and has wide applications in industry and home robotics such as bin picking and decluttering. 
When the physical properties of objects and gripper jaws are known, frictional contact models are typically used to plan grasps~\cite{ferrari.1992,bicchi2000robotic,okamura2000overview,miller.2004,danielczuk2019reach} or combined with learning techniques to detect grasps \cite{mahler2019learning} and to predict the success of manipulation tasks~\cite{xu2018learning}.


\begin{figure}[t]
	\centering
	\subfloat[]{\includegraphics[height=16em]{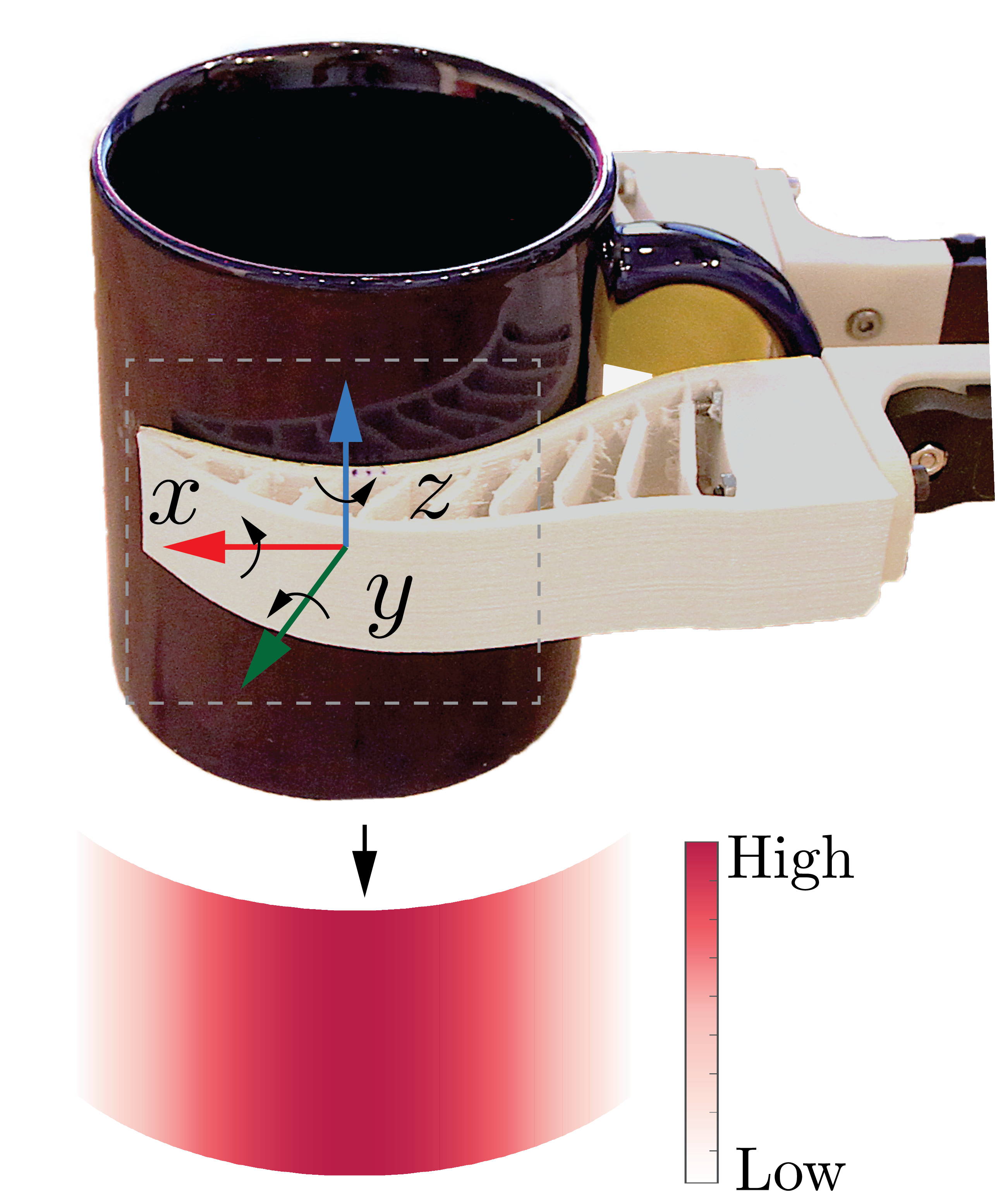} }%
	\subfloat[]{\includegraphics[height=16em]{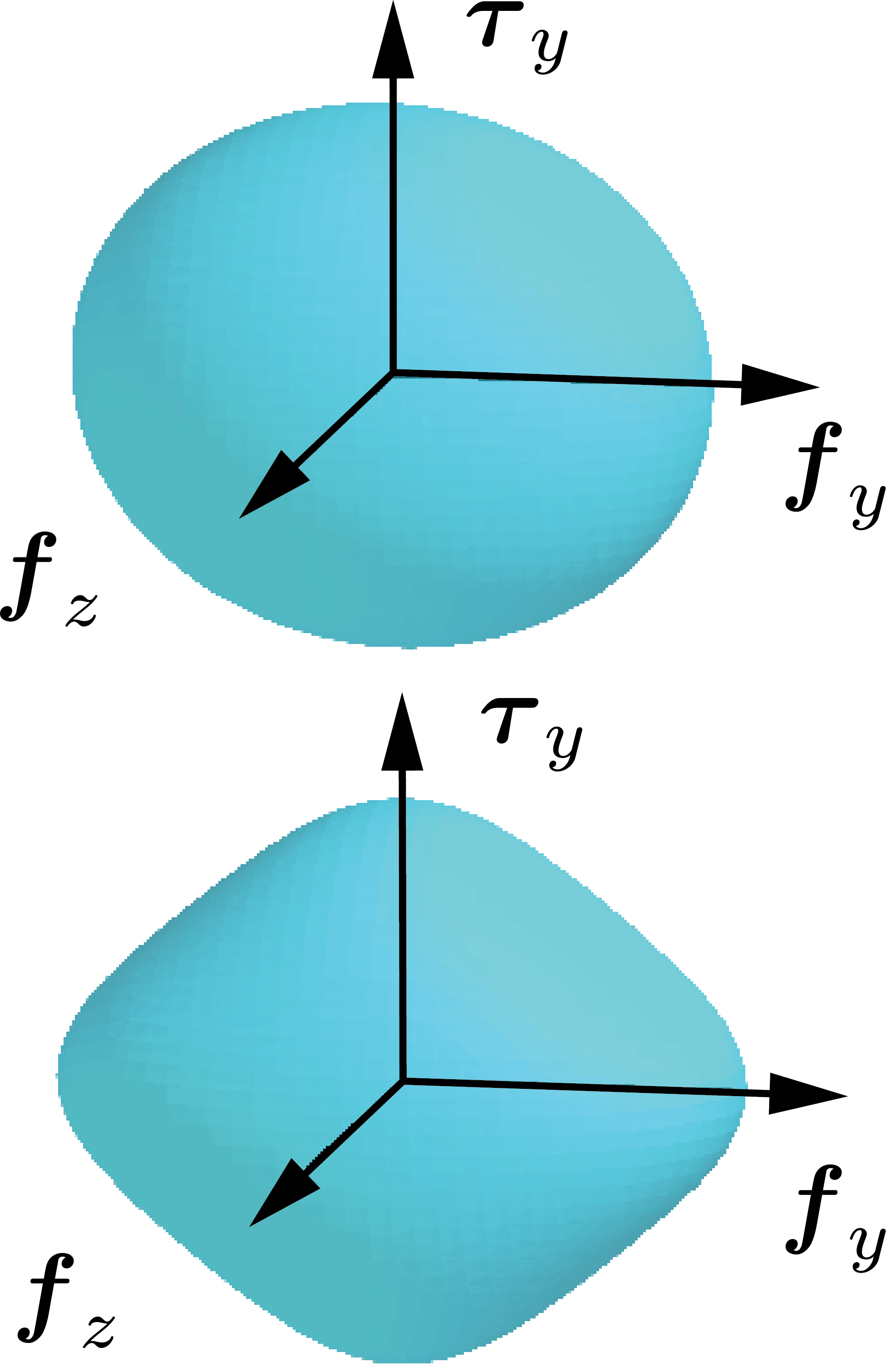} }%
	\caption{\small (a) A representative nonplanar surface contact is created when a soft parallel-jaw gripper deforms to a cup while grasping. Bottom: the extracted nonplanar contact surface and the pressure distribution, where redder colors represent higher pressure. (b) A 3D projection of the proposed elliptical (upper) and quartic (bottom) 6D limit surface model, which approximates the 6D frictional wrench limit of a nonplanar surface contact.}
	\label{fig:headshot}
\end{figure}

Deformable jaws or grippers covered with compliant materials \cite{wall2017method} are widely deployed in grasping applications as they deform to the local geometry of the object and can better resist external disturbances.
Such soft-finger grasps result in nonplanar surface contacts if the local geometry of the object is nonplanar, as shown in \fref{fig:headshot}(a), and the frictional wrench, a vector that is composed of the frictional force and torque, is 6D as the frictional force and torque are in three dimensions, respectively.  
Existing work in grasp planning~\cite{ciocarlie.2005,ciocarlie.2007,tsuji2014grasp,harada2014stability} typically assumes a planar contact area and uses a so-called \emph{limit surface (LS)}~\cite{goyal.1991}, which describes all possible 3D frictional wrenches that can be transmitted through a planar contact. 
Such planar contact models neglect the nonplanar surface geometry and only consider a 3D subset of the 6D frictional wrench, potentially lead to an overly conservative friction estimation.
Danielczuk~et~al.~\cite{danielczuk2019reach} modeled the frictional wrench for curved surface contacts by discretizing the contact surface to planar elements and fit a 3DLS model for each element, which is less efficient for fine surface geometry.

To address this issue, we model the 6D frictional wrenches that can be transmitted through the nonplanar surface contact between a deformable gripper jaw and a grasped object.
As friction depends on the relative motion between two bodies in contact, we derive the 6D frictional wrench (6DFW) for a given instantaneous motion of the grasped object.
However, for many robot grasping applications, the relative motion caused by external disturbances during the manipulation is unknown at the time of grasp planning. 
Therefore, we propose the 6D limit surface (6DLS), generalized from the 3DLS, to represent the 6D frictional wrench limit for the nonplanar surface contact.
We further present an ellipsoid and a quartic model as a low-dimensional representation to approximate a 6D limit surface. 
\fref{fig:headshot}(b) illustrates a 3D projection of each of the above 6DLS models.
We apply the 6DLS models to predict physical grasp success by building a \emph{grasp wrench space (GWS)} and compare the prediction results with two 3D planar contact models~\cite{lee1991fixture,zhou2018convex} and two 3D nonplanar surface contact models from our previous work~\cite{xu.2017}.

This paper makes the following main contributions:
\begin{enumerate}
    \item A concept of the 6D limit surface (6DLS), generalized from the 3DLS, to represent the 6D frictional wrench limit for a nonplanar deformable contact. 
    \item Two models to approximate a 6DLS and a pipeline to compute the models, based on the given contact surface and pressure distribution.
    \item An algorithm that synthesizes the grasp wrench space based on the proposed 6DLS models to predict multicontact grasp success.
\end{enumerate}

%% file: sections/3-related-work.tex
We summarize related work in isotropic frictional contact analysis for robot grasping.
Excellent surveys for contact modeling can be found in \cite{bicchi2000robotic,rimon_burdick_2019,kao2008contact,prattichizzo2016grasping} and for soft-fingered manipulation in \cite{li2001review,inoue2008mechanics}.

\subsection{Grasp Contact Models}
\mysum{General description}
For a point contact between two rigid objects, a so-called \emph{friction cone} is commonly used to describe the set of possible forces that can be transmitted through the contact.
When multiple point contacts are involved, the friction cone of each contact is typically approximated with a convex polyhedral cone \cite{kerr1986analysis,ferrari.1992} or an ellipsoid \cite{tsuji2009easy} to be efficiently formulated into a convex optimization problem for multicontact grasp analysis~\cite{miller.2004}.

Grasping with deformable jaws leads to a non-negligible contact area due to the additional torsional friction. 
The frictional wrench for such area contacts is, therefore, in three dimensions.
Goyal~et~al.~\cite{goyal.1991} proposed a limit surface to describe 3D frictional constraints for planar contacts. 
Lee and Cutkosky~\cite{lee1991fixture} approximated the LS with a 3D ellipsoid for computational efficiency and Zhou~et~al.~\cite{zhou2018convex} modeled the LS with a convex fourth-order polynomial to improve fitting accuracy. 

Building a limit surface requires a contact profile, including a contact area and a pressure distribution. 
The Hertzian contact~\cite{hertz1882uber} is a linear-elastic model, which describes the profile for planar contacts between two elastic bodies. 
Xydas and Kao~\cite{xydas.1999} proposed a general power-law pressure distribution function for anthropomorphic soft-finger modeling that captures different material properties and contact geometries.  
The limit surfaces for the power-law model with different exponents are calculated by numerical integration and the results validated the elliptical LS approximation proposed in~\cite{lee1991fixture}.
Tiezzi and Kao~\cite{tiezzi2007modeling} studied the viscoelastic contacts and the time-dependent evolution of the limit surfaces for soft fingers.
Fakhari~et~al.~\cite{fakhari2016development} introduced the asymmetry of a pressure distribution caused by the disturbance of tangential forces, which leads to a smaller LS and potentially a less robust grasp.
Arimoto~et~al.~\cite{arimoto2000dynamics,arimoto2002stable} proposed a radially-distributed model of soft-tip fingers for rigid object manipulation.
Inoue and Hirai~\cite{inoue2008mechanics} introduced a parallel-distributed model of hemispherical soft fingertips.
The model assumes that the fingertip consists of an infinite number of linear springs and is also suitable to the fingertips with tangential deformation.

Other work has used the finite element method (FEM) or local geometry approximation to acquire contact profiles.
Ciocarlie~et~al.~\cite{ciocarlie.2005} simulated the contacts between soft fingers and a rigid cube with FEM simulations and used the obtained contact profiles to build the ellipsoidal LS.
They extended the previous work in \cite{ciocarlie.2007} and computed the local geometry of two contact bodies based on the elastic contact theory~\cite{johnson.1987} for real-time grasp score computation. 
The algorithm assumed an elliptical contact area and used the Hertzian model and the Winkler elastic foundation to compute the pressure distribution. 
Tsuji~et~al.~\cite{tsuji2014grasp} generalized the elliptical contact area approximation to 2D quadric surfaces.
Harada~et~al.~\cite{harada2014stability} analyzed the contacts between rigid objects and a parallel-jaw gripper with a deformable pad attached to each jaw. 
The authors estimated the contact region by clustering the object model and obtained the contact area by projecting the contact region onto the plane that contains the undeformed jaw pad.  
Danielczuk~et~al.~\cite{danielczuk2019reach} proposed a model to approximate the contact area between soft jaw pads and rigid objects using constructive solid geometry and a linear model to estimate the pressure distribution. 
The contact area is decomposed into triangles. 
The combination of the LS of each triangle defines the frictional constraints of the contact. 

While existing work either assumes a planar contact surface or decomposes the surface into planar elements, we model the frictional wrenches for a nonplanar surface contact with a single limit surface.

\subsection{Wrench-based Multicontact Grasp Analysis}
A grasp wrench space describes possible wrenches that a grasp can act on an object and is commonly approximated with a convex hull of the union or Minkowski sum of the linearized 3D limit surface model of each contact.
The GWS is widely used to determine grasp quality; an excellent survey can be found in \cite{roa2015grasp} by Roa and Su{\'a}rez.
Nguyen~\cite{nguyen1988constructing} proposed the force-closure property, 
which indicates that an arbitrary external disturbance can be countered with the grasp configuration. 
To quantify the grasp quality, Ferrari and Canny~\cite{ferrari.1992} proposed the shortest distance between the origin and the facets of the GWS. 

While the algorithm for the GWS construction introduced in this work is similar to \cite{ferrari.1992}, we build the GWS with the proposed linearized 6DLS models, which capture the full 6D frictional constraints of each contact.

%% file: sections/5-problem-statement.tex
\section{Problem Statement}
\label{sec:problem_statement}

Given a contact surface and a pressure distribution, we compute a model that approximates the 6D limit surface, which is a surface that bounds the set of all possible 6D frictional wrenches that a deformable gripper jaw can exert on the object at the nonplanar surface contact.

To compute the limit surface of a contact, we make the following assumptions:
\begin{enumerate}
	\item We assume Coulomb friction with a known constant friction coefficient. 
	\item We assume that the contact profile, including the contact surface and the pressure distribution, or their estimations are known. A contact profile can be captured by a tactile sensor~\cite{ma.2019,romero2020soft} or estimated by a contact model~\cite{xydas.1999}. 
\end{enumerate}

We apply the limit surface to predict grasp success for a vertical lifting task with a parallel-jaw gripper. 
We compute one LS model for each gripper jaw and predict the grasp success with these two LS models by building a grasp wrench space.
We make the following assumptions for the prediction:
\begin{enumerate}
    \item The change of the contact profile is minor during the manipulation. Therefore, we do not recompute the limit surface models. Once a slip occurs, we consider the grasp failed. 
    \item We neglect inertial terms (quasi-static physics). Specifically, we assume that the robot arm lifts the object in a slow manner, so that the acceleration is negligible. The prediction can be too optimistic as the acceleration is not considered. For the scenarios with fast robot arm movement, the acceleration can be modeled as an additional external disturbance.
\end{enumerate}

\subsection{Notation and Definitions}

We adapt definitions from \cite{goyal.1991}\cite[pp.~67--93]{pressley.2010.elementary}\cite[pp.~45--50]{murray.1994.mathematical} and divide them into five categories. 
\subsubsection{Contact Profile}
\begin{itemize}
	\item $\mu$: friction coefficient 
	\item $\surface$: object surface that is in contact with a gripper jaw, defined as \emph{contact surface } 
	\item $\boldsigma\uv \in \R{3}$: parametric form of $\surface$ with $\uv \in \U$ being the parameters in the parametric space $\U \subseteq \R{2}$
	\item $\pressure\uv$:  contact pressure distribution 
	\item {\color{ceditor}$\pressureunit\uv=\pressure\uv/\int_\surface \pressure\uv\dS$:  normalized pressure distribution with $\int_\surface \pressureunit\uv\dS= 1$}
	\item $\normalfunc \in \R{3}$: surface normal 
	\item $\origin \in \R{3}$: contact pressure center
\end{itemize}

\subsubsection{Normal Force, Torque, Wrench}
\begin{itemize}
    \item $\dfperpfunc \in \R{3}$: local normal force impressed by $\pressure\uv$. $\diff\fperp\hspace{-0.2em}\uv$ is antiparallel to $\normalfunc$
	\item $F = \int_\surface \norm{\diff\fperp\hspace{-0.2em}\uv} = \int_\surface \pressure\uv\dS$:  sum magnitude of the local normal forces
	\item $\fperp = \int_\surface \diff\fperp\hspace{-0.2em}\uv$: \emph{normal force}
	\item $\tauperp\in \R{3}$: torque impressed by the pressure, defined as \emph{ normal torque}
	\item $\normalwrench = \left[\fperp^{\trans},\tauperp^{\trans}\right]^{\trans}$:   wrench impressed by the pressure, defined as \emph{normal wrench} 
\end{itemize}
\subsubsection{Instantaneous Motion}

\begin{itemize}
	\item $\lineISA$: instantaneous screw axis (ISA) 
	\item $\twist = \left[\vel^{\trans},\boldomega^{\trans}\right]^{\trans}$: instantaneous motion of the grasped object in a  three-dimensional space, defined as \emph{twist}.  $\boldomega  \in \R{3}$ and $\vel \in \R{3}$ are the angular and linear velocity, respectively.
	\item $\unittwist$: unit twist. $\unittwist = \left[\vel^{\trans},\boldomega^{\trans}\right]^{\trans}$ with $\normang=1$ or $\unittwist = \left[\vel^{\trans},\boldsymbol{0}^{\trans}\right]^{\trans}$ with $\norm{\vel}=1$.
	\item $h = \vel^{\trans} \boldomega / \normang^2$: pitch of a twist
\end{itemize}
\subsubsection{Frictional Force, Torque, Wrench}

\begin{itemize}
	\item $\wrench = \left[\boldf^{\trans},\boldtau^{\trans}\right]^{\trans}$:  frictional wrench of a contact, where $\boldf \in \R{3}$ and $\boldtau \in \R{3}$ are the frictional force and torque, respectively. 
    \item $\{\wrench_1 \ldotsc \wrench_\nprs\}$:  set of $\nprs$   frictional wrenches 
	\item {\color{ceditor}$\wrench_\text{max} \in \R{6}$: wrench composed of the maximal magnitudes of  $\{\wrench_1 \ldotsc \wrench_\nprs\}$ in the six dimensions}
	\item $\left\{\normalizedwrench_i \mid \normalizedwrench_i = \wrench_i \oslash \wrench_{\text{max}}, i\in\left\{1 \ldotsc \nprs\right\}\right\}$:  set of $\nprs$ normalized frictional wrenches, where $\oslash$ is the Hadamard division 
\end{itemize}
\subsubsection{Limit Surface and Frictional Constraints}
\begin{itemize}
	\item $\A$:  limit surface model of a contact. $\Ae$ and $\Aq$ denote the ellipsoidal and the quartic limit surface model, respectively.
	\item $\varepsilon\geq0$:   wrench fitting error of a limit surface model
	\item $\X$: set of linearized frictional constraints of a contact
	\item $\D$: dataset of linearized frictional constraints
\end{itemize}

Note that the units used in the example of wrench computation  (\sref{subsec:units}) and in the experiments are from the metric system. 
Particularly, we use meters (m) for length, Pascal (Pa) for pressure, Newtons (N) for force, (N$\cdot$m) for torque, seconds (s) for time, meters per second for linear velocity, radians per second for angular velocity. 
We omit these units from now on.


\subsection{Metrics}
\label{sec:metrics}
We evaluate a 6D limit surface model with the wrench fitting error, which is measured as the mean distance of sampled frictional wrenches to the LS.

Additionally, we apply the proposed 6DLS to predict multicontact grasp success. 
We consider a grasp to be successful if it lifts the object and there is no relative motion between the object and the gripper jaws; a failure otherwise.
We seek to maximize the precision and recall of binary predictions on physical experiments.
Denoting $n_\text{tp}$ as the number of true positive predictions, $n_\text{fp}$ as false positives, $n_\text{fn}$ as false negatives, precision and recall are $n_\text{tp}/(n_\text{tp}+n_\text{fp})$ and $n_\text{tp}/(n_\text{tp}+n_\text{fn})$, respectively. 
A combination of high precision and recall indicates that the algorithm predicts few false positives and false negatives.
Both metrics are commonly used for datasets with unequal class distributions in robot grasping applications~\cite{mahler2019learning,danielczuk2019reach} and describe if the algorithm is overconfident or too conservative~\cite{everingham2010pascal}.

%% file: sections/6-friction-param-surface.tex
In this section, we compute the contact wrench for a nonplanar surface contact.
A contact wrench consists of a normal and a frictional wrench.
While the former only depends on the contact profile, the latter also depends on the relative motion between the grasped object and the gripper jaw.
We start with formulating the motion, followed by computing the frictional wrench that resists the motion, and finally compute the normal force and torque of the contact. 
The contact wrench computation is a generalization of a planar surface contact~\cite{goyal.1991,howe.1996}\cite[pp.~130--134]{mason2001mechanics}. 
Details on the theoretical background are given in \aref{appendix:background}.

\begin{figure}[!t]
	\centering
	\subfloat[]{{\includegraphics[height=12.5em]{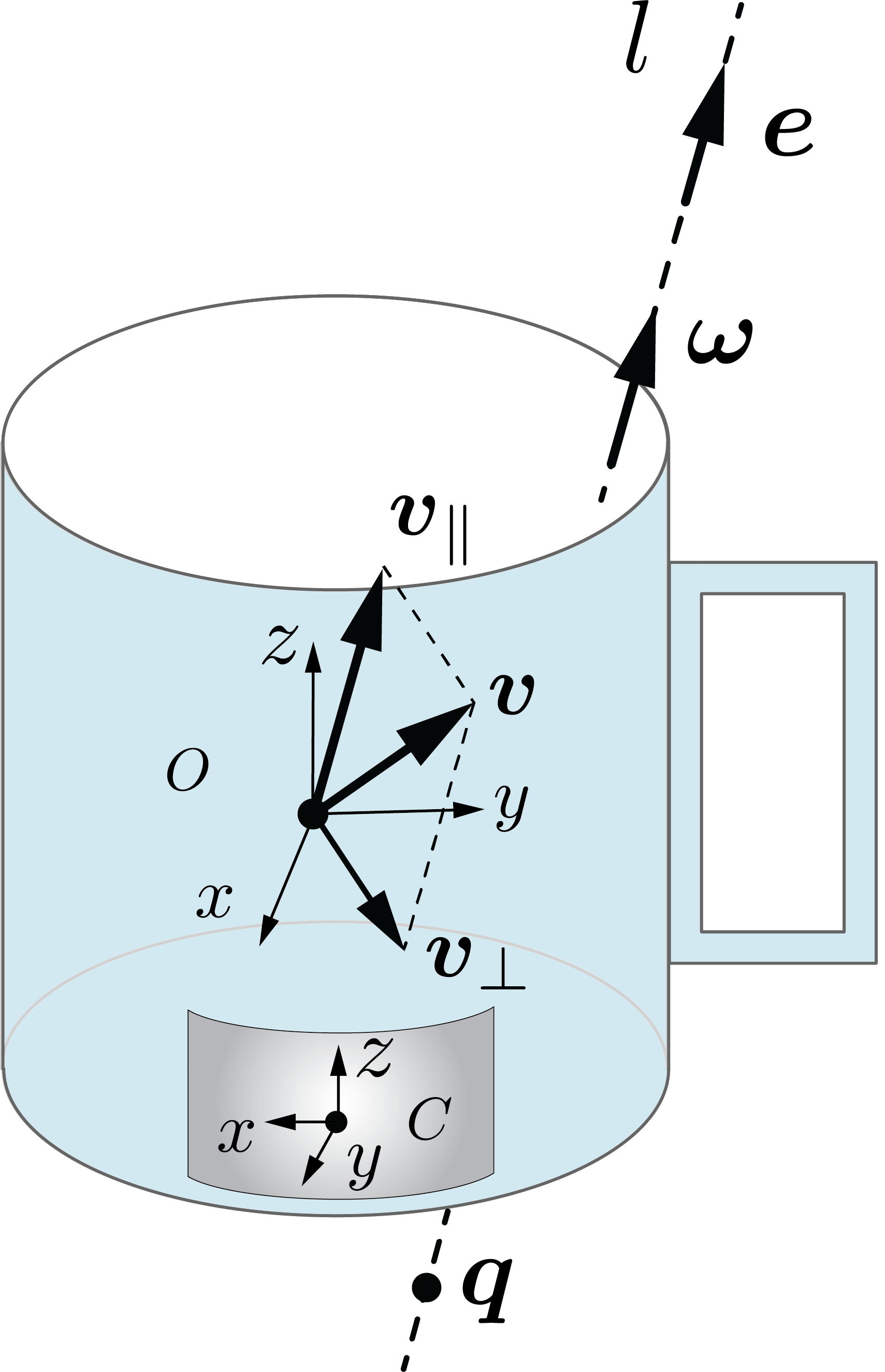} }}%
	\hfill
    \subfloat[]{{\includegraphics[height=12.5em]{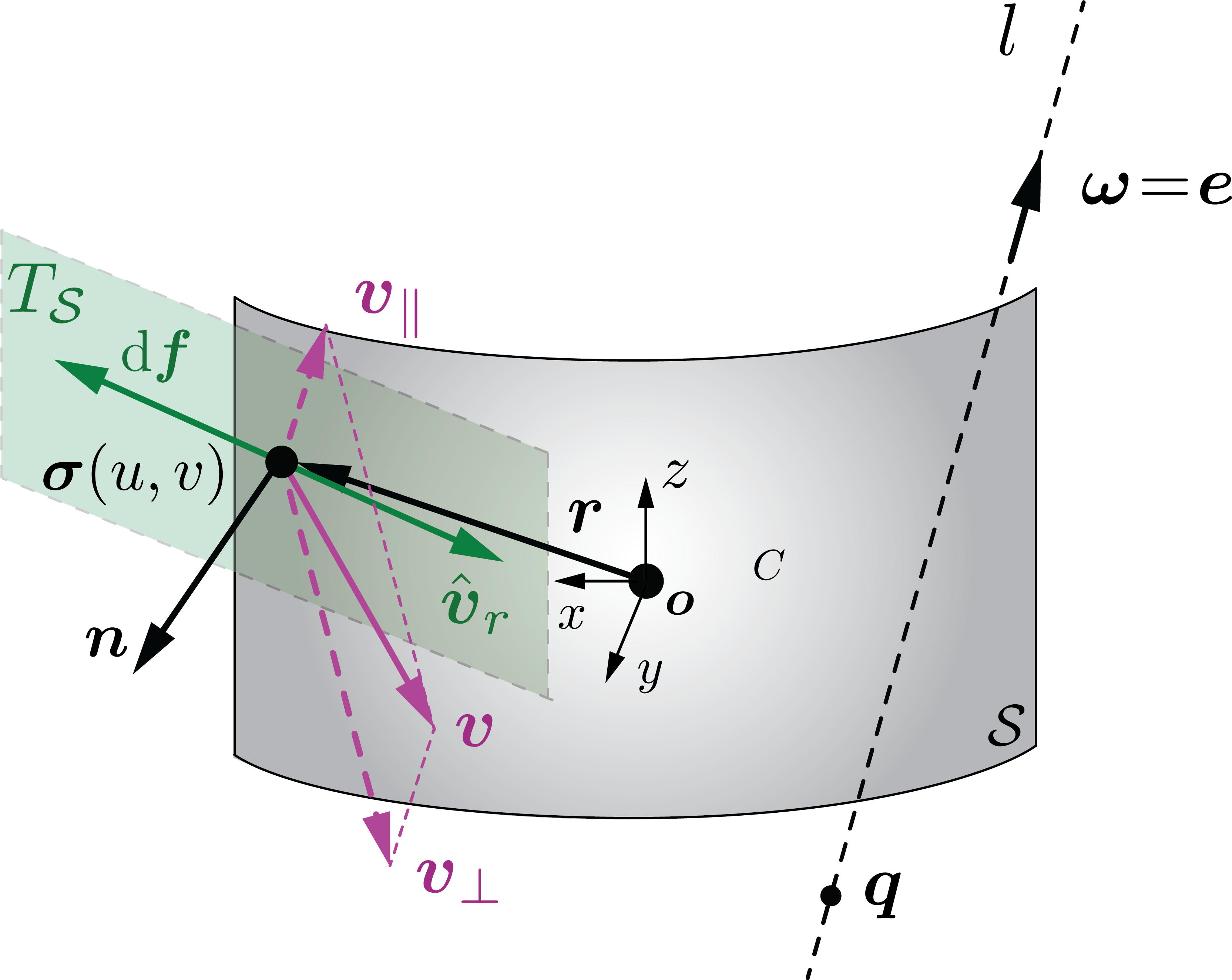} }}%
	\caption{\small (a) Instantaneous motion of a grasped object (cup) described with the instantaneous screw axis $\lineISA$ and the pitch $h$. The gray surface is the extracted contact surface between the object and a deformable gripper jaw. 
	(b) An enlarged view of the nonplanar contact surface $\surface$ and the local frictional force $\diff \vec{f}$ at a representative point on $\surface$ given $\lineISA$ and $h$.
	}
	\label{fig:curved-friction}
\end{figure}

\subsection{Instantaneous Motion}
\label{subsec:twist}
With an external disturbance acting on an object during a successful grasp, there is no relative motion between the object and the gripper jaws.
The impact of the external disturbance can be interpreted as a tendency of a relative motion.
Once the disturbance overcomes the maximal static friction, slip occurs and kinematic friction arises. 
As we model the static friction for grasping, we only consider the tendency of the object motion or the tendency of a relative motion.
We model the tendency of the object motion the same way as modeling an instantaneous body motion, as we consider the maximal static friction, which happens if the object is about to move.

The instantaneous motion in a three-dimensional space is defined as the \emph{twist}. 
A twist is a combination of a translation along an axis and a rotation about the same axis, defined as the \emph{instantaneous screw axis} (ISA).

Given an ISA, the linear velocity $\vel~\in~\R{3}$ at a point can be represented as the sum of a velocity parallel to and a velocity perpendicular to the ISA, denoted as $\vpara$ and $\vperp$, respectively. 
\fref{fig:curved-friction}(a) shows a representative grasped object (cup).
The origin of the object frame $O$ is located at the center of mass (COM) of the cup.
Given a representative ISA ($\lineISA$) illustrated as a dashed line, \fref{fig:curved-friction}(a) depicts the linear velocity at $O$.
{\color{ceditor}The instantaneous screw axis is represented as the line $\lineISA$ with direction $\linedir~\in~\R{3}$ that goes through a point $\linepoint~\in~\R{3}$.
Note that a line is completely defined by $\linedir$ and $\linepoint$, where $\norm{\linedir}=1$ and $\linepoint$ is an arbitrary point.
The \emph{Pl\"{u}cker coordinates}~\cite[pp.~60--68]{mason2001mechanics} of $\lineISA$ are defined as $(\linedir,\moment)$, where $\moment=\linepoint \times \linedir$ is the \emph{moment vector}. 
Using the Pl\"{u}cker coordinates to describe a line is beneficial since $\moment$ remains the same, regardless of which point $\linepoint$ on the line is chosen to compute $\moment$.
Note that a line in the space has four degrees of freedom (DoF), since a translation along the line or a rotation about itself leads to the same line.
In addition to the constraint $\norm{\linedir}=1$, the Pl\"{u}cker coordinates $(\linedir,\moment)$ reflect the four DoF of $\lineISA$ by satisfying $\linedir^T\moment=\linedir^T\left(\linepoint \times \linedir\right)=0$.}

We denote $\boldomega~\in~\R{3}$ as the angular velocity about $\lineISA$ with $\boldomega=\normang\linedir$ and denote $h= \vel^{\trans} \boldomega / \normang^2$ as the \emph{pitch of a twist}, which is the ratio of translational to rotational motion. 
{\color{ceditor}The linear velocity at $O$ is $\vel = \vpara + \vperp = h \boldomega +\linepoint\times\boldomega= h \boldomega +\normang\moment$. 
The twist $\twist$ at the origin is defined as
\begin{equation}
\twist = \begin{bmatrix}{\vel}\\{\boldomega}\end{bmatrix}  = \normang\begin{bmatrix}{h \linedir + \moment}\\{\linedir}\end{bmatrix}.
\label{eq:twist}
\end{equation}}\leavevmode
There are two important special cases. 
If $\normang \neq 0$ and $h=0$, a twist is a pure rotation about $\lineISA$. 
If $\normang = 0$ and $h=\infty$, a twist is a pure translation along $\lineISA$ and $\twist = \norm{\vel} \left[\linedir^{\trans}, \boldsymbol{0}^{\trans}\right]^{\trans}$.

Since the magnitude of $\vel$ or $\angular$ does not affect the frictional wrench due to the Coulomb friction assumption, we define a \emph{unit twist}, adapted from~\cite[p.~49]{murray.1994.mathematical}: a unit twist is a twist such that either $\normang=1$, or $\normang=0$ and $\norm{\vel}=1$. 
Specifically, the unit twist $\unittwist$ consists of two cases: for the motion that contains a rotational component, one obtains $\unittwist$ by substituting $\normang=1$ in \eref{eq:twist}; for a pure translation, i.e.,  $\normang=0$, one obtains $\unittwist$ by substituting $\norm{\vel}=1$.
{\color{ceditor}In summary, $\unittwist$ is 
\begin{equation}
 \unittwist =  
\begin{dcases*}
\begin{bmatrix}{h\linedir +\moment} \\ {\linedir} \end{bmatrix}& 
if $ \normang\neq 0$\\
\begin{bmatrix}{\linedir} \\ {\vec{0}} \end{bmatrix} &if $ \normang= 0$. 
\end{dcases*}
\label{eq:unittwist}
\end{equation}
Since the triplet $\tripletmoment$ or the direction vector $\linedir$ alone completely defines $\unittwist$ depending on $\normang$, we describe $\unittwist$ with $\tripletmoment$ or $\linedir$ to compute the relative motion and the frictional wrench.}

\subsection{Frictional Wrench}
\label{subsec:frictionParametricSurface}
Given the unit twist $\unittwist$ of the grasped object, we compute the linear velocity of the object at a  point on the contact surface and derive the direction of the relative velocity between the object and the jaw at this point. 
The gray surface illustrated in \fref{fig:curved-friction}(a) is a nonplanar contact surface between the cup and a deformable gripper jaw. 
\fref{fig:curved-friction}(b) depicts the enlarged view of the surface.

As we compute the frictional wrench with respect to the contact pressure center $\origin$, we can (without loss of generality) define a local contact frame $C$ with a rectilinear coordinate system, whose origin is located at $\origin$ and axes are arbitrarily chosen.
To compute $\origin$, we denote the pressure distribution as $\pressure$ and the contact surface as $\surface$ whose parametric form is $\boldsigma(u,v)$ with $(u,v)~\in~\R{2}$ being the parameters. 
We define $\origin$ as 
\begin{equation}
\begin{aligned}
\vec{\origin} = \begin{bmatrix}{\originx}\\{\originy}\\{\originz}\end{bmatrix} &= \frac{\int_ \surface \pressure(u,v) \bigcdot \boldsigma(u,v) \dS}{\int_ \surface \pressure(u,v)\dS}
\label{eq:origin}
\end{aligned}
\end{equation}
where the integral of a 3D vector function is defined as three individual integrals of each component.
Note that $\origin$ may not be on the contact surface.

{\color{ceditor}Similar to computing the linear velocity component in \eref{eq:unittwist}, 
given the Pl\"{u}cker coordinates $(\linedir,\moment)$ of the ISA and the scalar pitch $h$, the linear velocity $\vel$ at a point on the contact surface is
\begin{equation}
 \velfunc =  
\begin{dcases*}
h\linedir + \moment -\boldsigma(u,v) \times \linedir & 
if $ \normang\neq0$\\
\linedir &if $ \normang= 0$.
\end{dcases*}
\label{eq:velocity}
\end{equation}}\leavevmode
\fref{fig:curved-friction}(b) illustrates the linear velocity (pink), which is the sum of $\vel_\parallel$ and $\vel_\perp$, at a point on $\surface$ for the given $l$ and $h$.

The direction vector $\vrfunc$ of relative velocity at a point depends on the velocity of the gripper jaw and the object in contact, and should be tangential to $\surface$ at this point. 
Since the gripper jaw is static after the grasp and prior to the manipulation, we compute $\vrfunc$ by projecting the linear velocity $\vel\uv$ of the grasped object at a point onto the tangent plane of $\surface$ at that point.
\fref{fig:curved-friction}(b) illustrates $\vr$ as a green vector, which is parallel to the projection of $\vel$ in the tangent plane $\tangent$, depicted as a green parallelogram.
Let $\boldsigma_u$, $\boldsigma _v$ be the first-order derivatives of $\boldsigma(u,v)$ with respect to $u$ and $v$, respectively. The surface normal of $\tangent$  is  $\normal={\boldsigma _u \times \boldsigma _v}/{\norm{\boldsigma _u \times \boldsigma _v}}$.
We compute $\vrfunc$ with
\begin{equation}
\vrfunc = \frac{\left(\mat{I} -  \normalfunc \hspace{0.1em} \normalfunc^{\trans}\right)\hspace{0.1em} \velfunc}{\norm{\left(\mat{I} -  \normalfunc \hspace{0.1em} \normalfunc^{\trans}\right)\hspace{0.1em} \velfunc}}
\label{eq:projvp}
\end{equation}
where $\mat{I}$ is a $3\times3$ identity matrix.

The local frictional force $\diff\boldf$ at a point is antiparallel to $\vr$ and  $\diff\boldf = -\mu \bigcdot \pressure \bigcdot \vr \dS$, where 
$\dS = \norm{\boldsigma_u \times \boldsigma_v}  \diff{u} \diff{v}$ is the area of an infinitesimally small piece of $\surface$.
The local frictional torque is $\diff\boldtau =  \arm \times \diff\vec{f}$, where $\arm = \boldsigma(u,v) - \vec{\origin}$ is the torque arm.
By integrating $\diff\boldf$ and $\diff\boldtau$ over $\surface$, the frictional wrench $\wrench$ of the contact surface acting on the object is 
\begin{equation}
\begin{aligned}
\wrench &= \begin{bmatrix}{\boldf}\\{\boldtau}\end{bmatrix} = \begin{bmatrix}{- \mu\int _\surface \pressure(u,v) \bigcdot  \vrfunc \dS} \\ 
{ - \mu\int _\surface \pressure(u,v) \bigcdot [\arm(u,v)  \times \vrfunc] \dS}\end{bmatrix}.
\end{aligned}
\label{eq:frictionArea}
\end{equation}
Computing $\wrench$ for a parametric surface can be inefficient because of the integral operation.
Therefore, we also introduce the frictional wrench computation for a discrete surface with convex polygonal elements. The method is also beneficial for a surface whose parametric form is nontrivial to determine.
Due to the similarity of the computation for a parametric and a discrete surface, we put the latter in \aref{appendix:friction_discrete}.
We analyze the runtime and the error of the frictional wrench due to the surface discretization in Section~\ref{sec:discretizationEffect}.

\subsection{Normal Wrench}
\label{subsec:normal_wrench}
We define the \emph{normal wrench} $\normalwrench$ as the wrench impressed by the pressure distribution $\pressure$ with respect to the pressure center $\origin$.
By integrating the local normal force $\diff\fperp$ and torque $\diff\tauperp$ at each point on the contact surface $\surface$, we obtain 
\begin{equation}
\begin{aligned}
\normalwrench = \begin{bmatrix}{\fperp}\\{\tauperp}\end{bmatrix} =  \begin{bmatrix}{-\int _\surface    \pressure(u,v) \bigcdot \normal(u,v) \dS} \\
{{-\int _\surface \pressure(u,v) \bigcdot [(\vec{r}(u,v)  \times \normal(u,v)] \dS}}\end{bmatrix}
\end{aligned}
\label{eq:normal_wrench}
\end{equation}
where $\fperp$ and $\tauperp$ are the normal force and torque impressed by the pressure acting on the object, respectively. 
The minus sign comes from the fact that the force acting on the object is towards the object, whereas the normal vector $\normal$ points outward.
In contrast to the frictional wrench, which depends on the unit twist, $\normalwrench$ is uniquely defined by $\surface$ and $\pressure$. 

We also compute the sum magnitude $F$ of the local normal forces with $F=\int_\surface \pressure\dS$. Note that $F > \norm{\fperp}$ for the representative contact surface depicted in \fref{fig:curved-friction}(b) with an axisymmetric pressure distribution,  since the integral of $\diff\boldf_{\perp,x}$ over the left half surface cancels the integral of $\diff\boldf_{\perp,x}$ over the right half surface. We use $F$ to compute the normalized pressure distribution $\pressureunit = p/F$ and use $\normalwrench$ to construct the grasp wrench space, as presented in \sref{sec:multi_contacts}.

\subsection{Example of Contact Wrench Computation}
\label{subsec:units}
We provide an example to compute the contact wrench for the representative contact surface shown in \fref{fig:curved-friction}.
We consider an elliptic cylinder with the parametric form $\boldsigma(u,v) = \left[0.02\cos u,0.02\sin u,v\right]^\trans - \left[{0},{0.018},{-0.04}\right]^\trans ,u\in\left[0.25\pi,0.75\pi\right],v\in\left[-0.05,-0.03\right]$ in the local contact frame $C$, whose origin is located at the pressure center $\origin$. 
Given the contact profile, including $\boldsigma(u,v)$ and the uniform pressure distribution $p(u,v)=10^3$, we first compute the normal wrench with respect to $\origin=\left[0,0,0\right]^T$ in $C$. 
Details about finding the frame $C$ can be found in \aref{sec:app_local}.

\subsubsection{Normal Wrench}
By substituting $\diff \surface = \norm{\boldsigma_u \times \boldsigma_v}  \diff{u} \diff{v} = 0.02\diff{u} \diff{v}$, $\normalfunc ={\boldsigma _u \times \boldsigma _v}/{\norm{\boldsigma _u \times \boldsigma _v}} = \left[\cos u,\sin u, 0\right]^\trans$, $p(u,v)=10^3$, $\boldsymbol{r}(u,v) = \boldsigma(u,v)-\origin$ into \eref{eq:normal_wrench}, we have $\normalwrench = \left[\fperp^\trans,\tauperp^\trans\right]^
\trans \approx \left[0,-0.5657,0,0,0,0\right]^\trans$. 
The sum magnitude of local normal forces $F = \int_\surface p \dS \approx 0.6283 > \norm{\fperp}$.
The normalized pressure distribution is $\pressureunit(u,v)  = p(u,v)/F \approx 1592.$

\subsubsection{Frictional Wrench}
In addition to the contact profile, the frictional wrench $\wrench$ also depends on $\mu$ and the unit twist.
{\color{ceditor}Given $\mu=0.3$ and a unit twist described with $\linedir = \left[0,-1,0\right]^T$, $\moment = \left[0,0,0\right]^T$, and $h=0$, which is a pure rotation around the negative $y$-axis, we compute $\wrench$ for the same contact profile.}

We first compute the linear velocity of the object by substituting $\linedir,\moment$, $\boldsigma, h$ into \eref{eq:velocity} and obtain $\velfunc = \left[-v-0.04, 0, 0.02\cos u\right]^\trans$.
Next, we determine the direction vector $\vrfunc$ of the relative velocity by substituting $\velfunc$, $\normalfunc$ into \eref{eq:projvp} and have $\vrfunc=$
\begin{equation*}
    \frac{\left[-\sin^2u(v+0.04),\sin u \cos u (v+0.04),0.02\cos u\right]^\trans}{\sqrt{\sin^2 u(v+0.04)^2 + (0.02\cos u)^2}}.
\end{equation*}
Finally, we compute the frictional wrench $\wrench$ by substituting $\mu$, $p(u,v)$, $\vrfunc$, $\diff \surface$, $\boldsymbol{r}(u,v)$ into \eref{eq:frictionArea} and obtain $\wrench \approx \left[0,0,0,0,0.0018,0\right]^\trans$. 
The local frictional force direction vectors for the representative unit twist are shown in \fref{fig:smpl_wrenches}(a).

%% file: sections/8-limit-surface.tex
\section{Six-dimensional Limit Surface}
\label{sec:LS}
So far, we have computed the frictional wrench of a nonplanar surface contact given a single unit twist of the grasped object.  
We now study the problem of modeling all possible frictional wrenches that can be transmitted through a contact by sampling the space of unit twists and finding a 6D limit surface. 
We start with the LS definition, followed by finding possible frictional wrenches of a contact and fit two 6DLS models to the wrenches. 
The 6DLS models, an ellipsoid and a convex quartic (fourth-order polynomial) model, are generalized from the corresponding 3D models proposed in \cite{lee1991fixture} and \cite{zhou2018convex} for planar surface contacts.

\begin{figure}
	\centering
	\subfloat[]{{\includegraphics[width=0.45\linewidth]{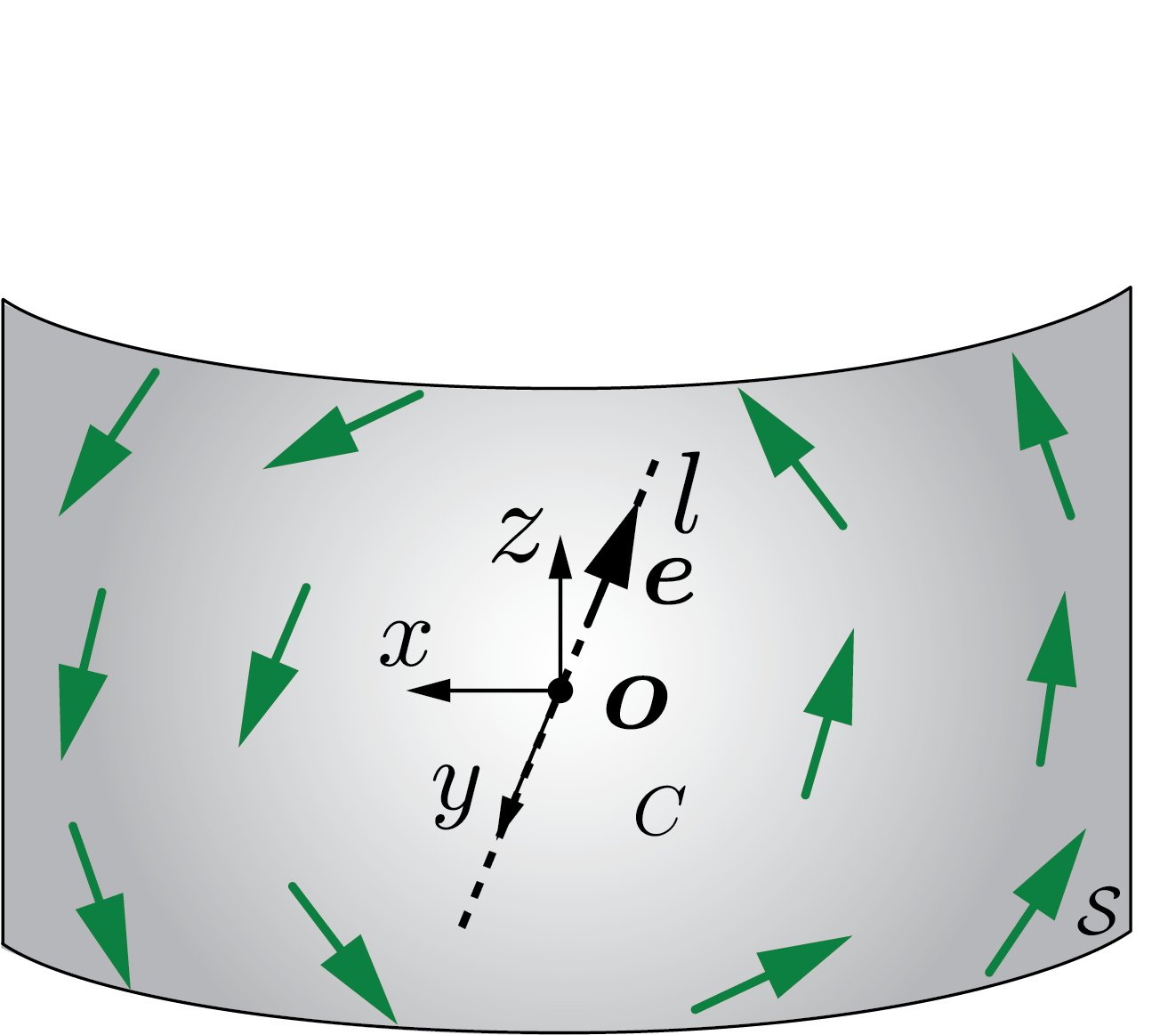} }}%
	\hspace{1em}
	\subfloat[]{{\includegraphics[width=0.45\linewidth]{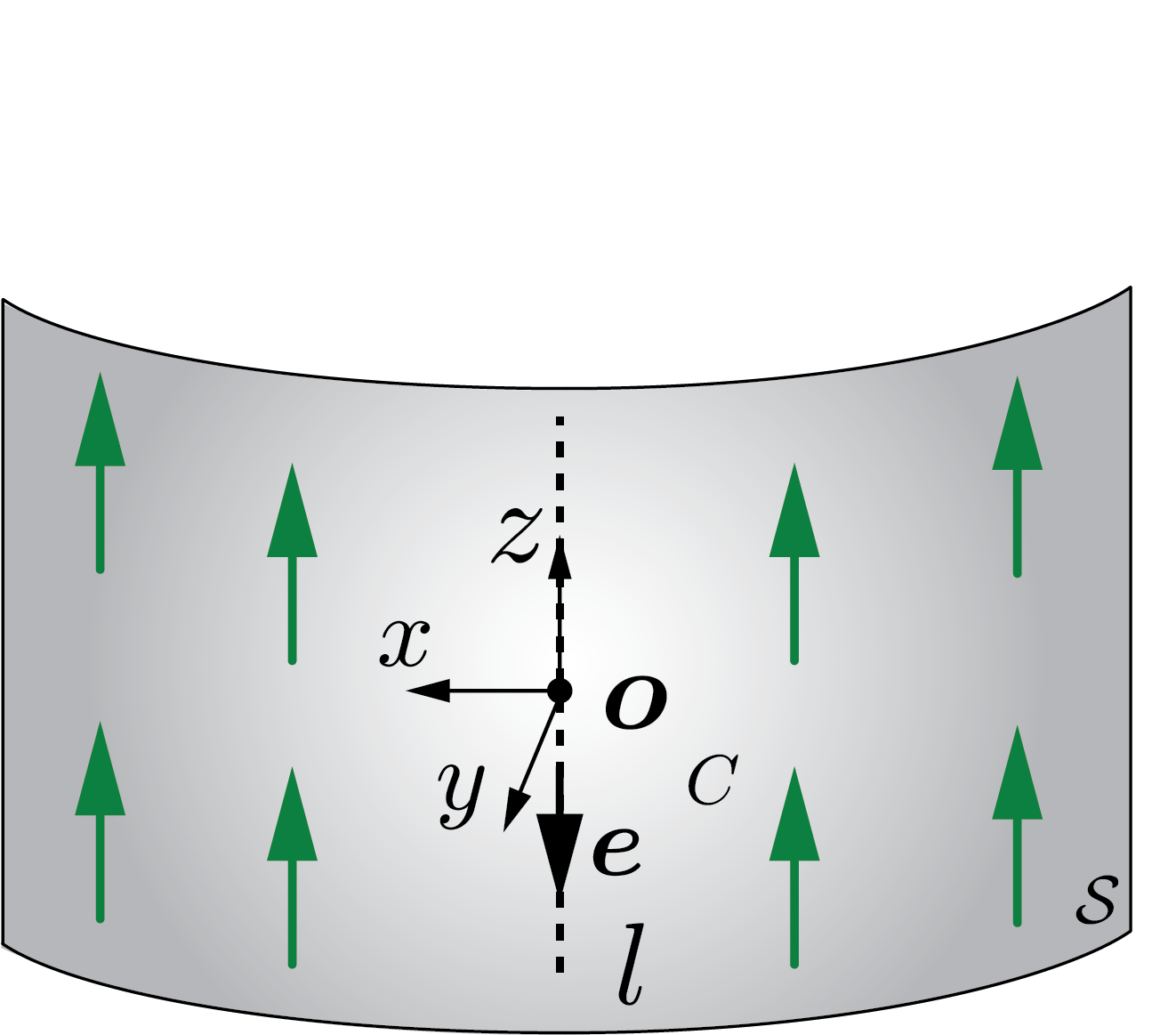} }}
	\caption{\small Local frictional force direction vectors (green) of a nonplanar surface contact acting on the grasped object, (a) if the object rotates around the negative $y$-axis, (b) if the object slides along the negative $z$-axis relative to the gripper.}
	\label{fig:smpl_wrenches}
\end{figure}

\subsection{Definition}
A limit surface is the boundary of the set of all possible frictional wrenches that can be applied through one contact or a set of contacts \cite{goyal.1991}.
Specifically, let $f(\x)=1$ with $\x \in \R{6}$ define the limit surface, an arbitrary frictional wrench $\wrencharb$ is constrained by
\begin{equation}
    f(\wrencharb) \leq 1.
\label{eq:LS_constraint}
\end{equation}

\subsection{Finding the Frictional Wrenches}
\label{sec:wrenches_LS}
One intuitive way to build a 6D limit surface consists of densely sampling the motion space and computing the frictional wrench for each motion, which is potentially a time-consuming operation.
To increase efficiency, we sample a finite number of the unit twists of the grasped object and compute the corresponding frictional wrenches using \eref{eq:velocity}--(\ref{eq:frictionArea}).
Then we fit a model to the wrenches to approximate the 6D limit surface.

{\color{ceditor}Recall that the triplet $\tripletmoment$ or the direction vector $\linedir$ alone uniquely defines a unit twist.
Due to the constraint $\linedir^T\moment=0$, one intuitive way is to sample $\linedir$ and two components of $\moment$, for instance, $m_x$ and $m_y$, and compute the third component with $m_z=-(e_xm_x+e_ym_y)/e_z$.
However, the division leads to numerical instability if $e_z=0$. 
Therefore, we sample $\triplet$ for the motions that include a rotation and compute $\moment$ with  $\moment=\linepoint\times\linedir$, and sample $\linedir$ for pure translations.}
This leads to a total of $\nprs$ motion samples. 
\fref{fig:smpl_wrenches} shows two representative sampled unit twists and the resulting local frictional force direction vectors, which are used to compute the frictional wrench of the contact.
\fref{fig:smpl_wrenches}(a) illustrates a pure rotation around the negative $y$-axis with $\linedir = [0,-1,0]^T,\linepoint=[0,0,0]^T$, $h=0$, and $\norm{\boldomega}=1$ in the local contact frame $C$, where $\tau_y$, the frictional torque component around the $y$-axis, is maximized for the representative surface.
\fref{fig:smpl_wrenches}(b) shows a pure translation along the negative $z$-axis with $\linedir=[0,0,-1]^T$ and $\norm{\boldomega}=0$, where $f_z$ reaches the maximum.
Details about the motion sampling are provided in \aref{appendix:data}.

{\color{ceditor}Let $\wrench_{i}$ be the $i$th frictional wrench with $i\in\{1 \ldotsc \nprs\}$.}
We normalize $\wrench_i$ so that each component of $\wrench_i$ is in the range of $[-1,1]$ for numerical stability. 
Let $\wrench_\text{max} = \left[f_{x,\text{max}},f_{y,\text{max}},f_{z,\text{max}}, \tau_{x,\text{max}},\tau_{y,\text{max}},\tau_{z,\text{max}}\right]^{\trans}$ be the wrench composed of the maximal magnitudes of the $\nprs$ frictional wrenches in the six dimensions. 
The $i$th normalized frictional wrench $\normalizedwrench_i$ is
\begin{equation}
	\normalizedwrench_i =  {\wrench}_i \oslash \wrench_\text{max}.
	\label{eq:normalization}
\end{equation}
Note that $\normalizedwrench$ is dimensionless.
A representative set of $\{\normalizedwrench_1 \ldotsc \normalizedwrench_\nprs \}$ is illustrated in \fref{fig:LS} as orange dots. 

\subsection{Finding an Ellipsoid}
\label{subsuc:LS_ell}
The first proposed model to approximate a 6D limit surface is a 6D ellipsoid.
An arbitrarily oriented zero-centered 6D ellipsoid is defined by $\fe(\x) = \x^{\trans} \Ae \x=1$, where $\Ae \in \mathbb{R}^{6 \times 6}$ is a positive definite matrix. 

\begin{figure}
	\centering
	\subfloat[]{{\includegraphics[height=35mm]{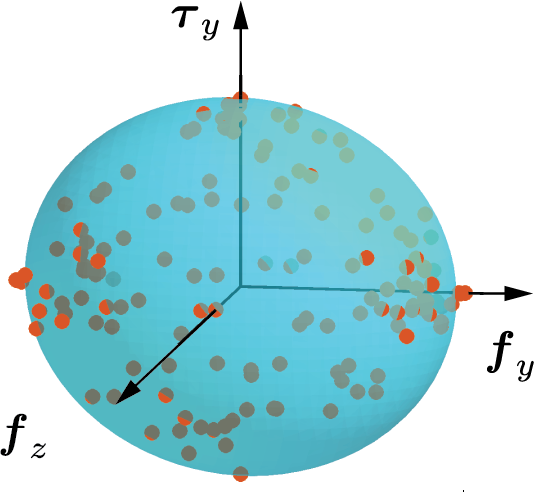} }}%
	\hspace{1em}
	\subfloat[]{{\includegraphics[height=35mm]{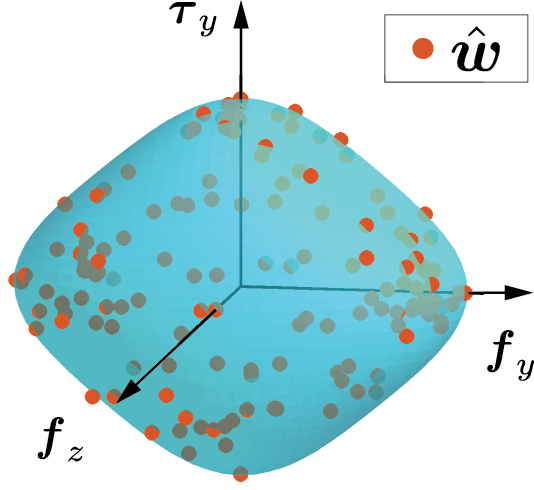} }}%
	\caption{\small A 3d projection of (a) an ellipsoid and (b) a quartic 6DLS model fit to the normalized frictional wrenches (orange dots).}
	\label{fig:LS}
\end{figure}

We fit an ellipsoid to the normalized frictional wrenches $\{\normalizedwrench_1 \ldotsc \normalizedwrench_\nprs\}$ by formulating the optimization problem
\begin{equation}
\begin{aligned}
& \underset{\Ae}{\text{minimize}}
& &  \sum_{i=1}^\nprs \left( \fe(\normalizedwrench_i) - 1 \right)^2  \\
& \text{subject to}
& & \fe(\x) = \x^{\trans} \Ae \x  \\
&&& \Ae \succ 0
\end{aligned}
\label{eq:ls_ellip}
\end{equation}
where $\Ae \succ 0$ means that $\Ae$ is positive definite. 
Since $\Ae$ uniquely defines an ellipsoid, we denote $\Ae$ as the ellipsoidal LS model.
\fref{fig:LS}(a) shows a 3D projection of the 6D ellipsoid fit to the normalized wrenches.

\subsection{Finding a Quartic}
\label{subsuc:LS_qua}
While an ellipsoid is a practical LS model for quasi-static~\cite{kao1992quasistatic} and sliding manipulations~\cite{howe.1996}, Zhou~et~al.~\cite{zhou2018convex} suggested that a convex homogeneous quartic model better captures the force-motion relation for planar sliding. 
We generalize the quartic (fourth-order polynomial)~\cite{zhou2018convex}, which describes a limit surface in 3D, to approximate a limit surface in 6D, which requires that the polynomial defining this surface is a function of all six components of $\x$.

To formulate the quartic $\fq(\x)$ with $\x\in\R{6}$, we first denote the nonnegative integer $d_j$ as the degree of the $j$th component of $\x$ with $\sum_{j=1}^6 d_j=4$ for the homogeneous quartic. 
The quartic surface with up to $L$ terms is defined by $\fq(\x) = a_1x_1^4 + a_2x_1^3 x_2 + \ldots + a_Lx_6^4 = \sum_{\kappa=1}^L \left(a_\kappa \prod_{j=1}^6 x_j^{d_{j,\kappa}}\right)=1$, where $a_{\kappa} \in \mathbb{R}$ is the coefficient of the $\kappa$th term. 
Here, $L = \binom{4+6-1}{6-1}=126$. 
Since the coefficients uniquely define a quartic surface, we denote $\Aq = \left[a_1 \ldotsc a_L\right]^{\trans}$ as the quartic LS model. 
Note that $\Aq \in \R{126}$, whereas $\Ae~\in~\R{6\times 6}$ for an ellipsoid.  $\Aq$ is uppercase for consistency as $\A$ denotes the LS model. 

\mysum{SOS-convex}
Zhou~et~al.~\cite{zhou2018convex} showed that it is essential to enforce the convexity of the quartic. 
However, the convexity is NP-hard to determine if $d>2$ and $L>1$. 
Inspired by Magnani~et~al.~\cite{magnani.2005}, we use a relaxation technique that enforces the convexity of $f_2(\x)$ only on a region by using the concept of \emph{sum-of-squares (SOS)}.

Let $\vec{z} \in \R{6}$ be a nonzero auxiliary variable, $\vec{y}(\x,\vec{z}) = \left[{\xnb_1  \vec{z}^{\trans}}, {\xnb_2 \vec{z}^{\trans}},  {\xnb_3 \vec{z}^{\trans}},  {\xnb_4 \vec{z}^{\trans}},  {\xnb_5 \vec{z}^{\trans}}, {\xnb_6 \vec{z}}^{\trans}\right]^{\trans} \in \mathbb{R}^{36}$. 
The polynomial $\fq(\x)$ is defined as \emph{SOS convex}, if there exists a positive definite matrix $\psd \in \mathbb{R}^{36 \times 36}$ such that 
\begin{equation}
\vec{z}^{\trans}\nabla^2\fq(\x)\vec{z}  = \vec{y}(\x,\vec{z})^{\trans}\psd\vec{y}(\x,\vec{z}).
\label{eq:sos_convex}
\end{equation}
We reformulate \eq~(\ref{eq:sos_convex}) as sparse linear constraints of $\Aq$ and the vectorization of $\psd$ with
\begin{equation}
	\mat{V}_1\mathrm{vec}(\psd) = \mat{V}_2\Aq
	\label{eq:sos_linear}
\end{equation} 
where $\mat{V}_1 \in \mathbb{N}_0^{441\times1296}$ and $\mat{V}_2\in \mathbb{N}_0^{441\times126}$ with  $\mathbb{N}_0 = \{\mathbb{N} \cup \{0\}\}$ are constant sparse matrices, and $\mathrm{vec}(\cdot)$ denotes the vectorization operation.

We fit a quartic surface to $\{\normalizedwrench_1 \ldotsc \normalizedwrench_\nprs\}$ by formulating the optimization problem
\begin{equation}
\begin{aligned}
& \underset{{\Aq=\left[a_1 \ldotsc a_L \right]^{\trans},\psd}}{\text{minimize}}
& &   \sum_{i=1}^\nprs \left(\fq(\normalizedwrench_i) - 1\right)^2 \\
& \text{subject to}
& & \fq(\x) = \sum_{\kappa=1}^L \left(a_\kappa \prod_{j=1}^6 x_j^{d_{j,\kappa}}\right)\\
&&& 	\mat{V}_1\mathrm{vec}(\psd) = \mat{V}_2\Aq \\
&&& \psd \succ 0.
\end{aligned}
\label{eq:ls_quartic}
\end{equation}
\fref{fig:LS}(b) shows a 3D projection of the quartic 6DLS model fit to $\{\normalizedwrench_1 \ldotsc \normalizedwrench_\nprs\}$. 

To evaluate both models, we use the mean distance of the wrench samples to the surface as the wrench fitting error 
\begin{equation}
    \varepsilon_{1,2} = \frac{1}{\nprs}\sum_{i=1}^\nprs\norm{f_{1,2}(\normalizedwrench_i)-1}.
    \label{eq:fitting_error}
\end{equation}
Note that $\varepsilon_{1,2}$ does not have a unit as the normalized frictional wrenches are unitless.

Algorithm~\ref{algo:6dls} summarizes the process to find a 6DLS model for a nonplanar surface contact. The description after double slash (//) in the algorithm is a comment.
As the limit surface models are fit to the normalized frictional wrenches, Algorithm~\ref{algo:6dls} also outputs $\wrench_\text{max}$ to denormalize the constraints for multicontact grasp analysis in \sref{subsub:linear_constraints}.

%% file: sections/9-multi-contacts.tex
So far, we have introduced the algorithm to find the limit surface model $\A$, which approximates the upper bound of the frictional forces and torques that can be transmitted through a nonplanar surface contact. 
Given $\A$ for each contact of a grasp, we now predict if the grasp can resist an external wrench, such as the gravity of the grasped object. We construct the space of wrenches that the contacts can apply to the object, defined as the \emph{grasp wrench space} (GWS)~\cite{ferrari.1992}, to infer the grasp success. 
We start with linearizing the limit surface models, followed by constructing a GWS using these linear  frictional constraints. 

\input{algos/algo-ls}

\subsection{Linearizing the Frictional Constraints}
We linearly approximate the frictional constraints for efficiency, similar to~\cite{ferrari.1992}. 
Specifically, instead of using \eref{eq:LS_constraint}, we constrain a frictional wrench to lie inside the discrete limit surface.
We discretize the LS model by sampling the surface with $M$ vertices $\wrenchresmp_m \in \R{6},m\in\{1\ldotsc M\}$. 
This sampling process is beneficial as the frictional wrenches from the initial motion sampling can be unevenly distributed due to the geometry of the contact surface. 
In \sref{subsec:results}, we compare the prediction results and the grasp wrench spaces built with the frictional wrenches from the initial sampling and with the linearized limit surface models.

While there are multiple techniques to sample an ellipsoid, sampling a quartic surface is nontrivial. We propose the following sampling algorithm as it is applicable to both surfaces. 

\begin{figure}
	\centering
	{{\includegraphics[width=0.99\linewidth]{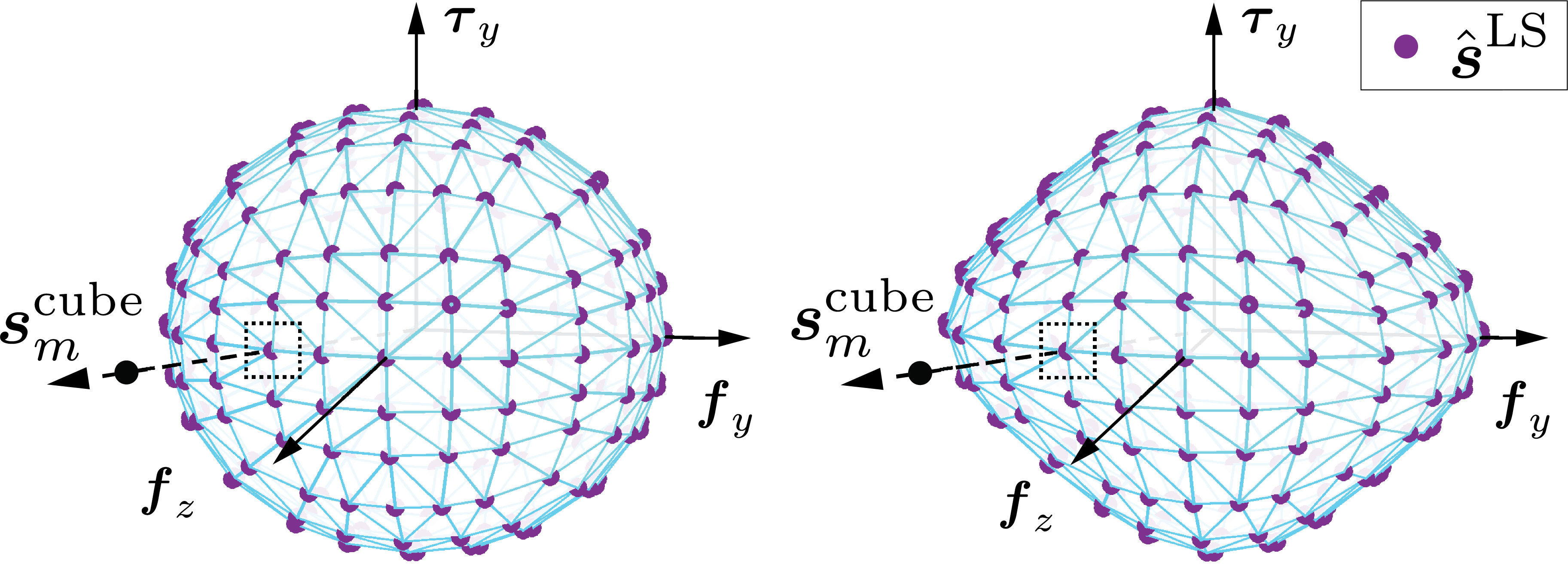} }}%
	\caption{\small The ellipsoidal (left) and the quartic (right) 6DLS models linearized by sampling the surface with vertices.}
	\label{fig:discrete_LS}
\end{figure}

We first evenly sample $M$ points that are on the surface of a 6D hypercube, where each side is in $[-1,1]$. Denoting $\verte_{m} \in \mathbb{R}^{6}$ as the $m$th vertex, we define a ray that starts from $\vec{0}\in\R{6}$ and goes through $\verte_m$. 
The intersection point of the ray and the limit surface model is the $m$th vertex $\wrenchresmp_m$ on the model.
\fref{fig:discrete_LS} illustrates a representative $\verte_m$ as the black dot and the ray that goes through $\verte_m$ as the dashed arrow.
The intersection point of the ray and the LS model is $\wrenchresmp_m$, depicted as the purple dot within the dashed rectangle. 
We parametrize the $m$th ray with $\gamma_m \bigcdot \verte_m$, where $\gamma_m$ is a positive scaling factor.
Given the LS equation $f(\x)=1$ with $f(\x)$ being $\fe(\x)$ for the ellipsoid or $\fq(\x)$ for the quartic, we find the intersection point $\wrenchresmp_m$ by solving the following equation system with Newton's method
\begin{equation}
    f(\wrenchresmp_m) = 1 \text{ and }
    \wrenchresmp_m = \gamma_m \bigcdot \verte_m \text{, } \gamma_m >0
    \label{eq:linearize_4th}
\end{equation}
where $\wrenchresmp_m$ and $\gamma_m$ are the variables to be solved in the equation system. The initial guess for the two variables are $\verte_m$ and 1, respectively, since $\wrenchresmp_m$ is near $\verte_m$. The vertex $\wrenchresmp_m$ represents the $m$th linear frictional constraint. 

\subsection{Denormalizing the Linear Frictional Constraints}
\label{subsub:linear_constraints}
Since both LS models are fit to the set of normalized frictional wrenches, described in \sref{sec:wrenches_LS}, we now denormalize the constraints. 
Given $\{\wrenchresmp_1 \ldotsc \wrenchresmp_M\}$, the set of denormalized frictional constraints $\X$ is 
\begin{equation}
    \X =  \left\{\wrenchresmpden_{m} \Bigm| \wrenchresmpden_{m}  = \wrenchresmp_m \circ \wrench_{\max}, m\in\{1 \ldotsc M\} \right\}
    \label{eq:denormailzation}
\end{equation}
where $\circ$ is the Hadamard product. 
The convex hull of $\X$ approximates the upper bound of the frictional wrenches that can be transmitted through a nonplanar surface contact. 

\subsection{Dataset of the Linear Frictional Constraints}
\label{sec:database}
Since the limit surface computation only requires the contact profile, including the contact surface $\surface$ and the pressure distribution $\pressure$, we store the computed linear frictional constraints $\X$ for the profiles in the dataset $\D$.
If we encounter the same or a close contact profile, we reuse a stored version in $\D$. 
Furthermore, we also note that $\X$ linearly scales with the sum magnitude of the local normal forces $F=\int_\surface \pressure\dS$.
One can precompute $\X$ for the pair, $\surface$ and the normalized pressure distribution $\pressureunit = \pressure/F$, and scale $\X$ with $F$ or with a force sensor reading for each jaw when constructing the GWS.  
Furthermore, one typically obtains contact profiles with tactile sensors in experiments.
As the sensors provide discrete profiles, the possible contact profiles are a finite set and the LS for each profile can be precomputed in extreme cases.

We define a \emph{profile pair} $\pairsp$ for a contact and compute $\X$ given $\pairsp$, and update $\D$ with $\D = \D \cup \X$.
When constructing the GWS, we retrieve the precomputed $\X$ for $\pairsp$ from $\D$ and denormalize $\X$ with $F$. 
If $\X \notin\D$, we compute $\X$ online.

\input{algos/algo-gws}

\input{figs/figParamSurfaces}

\subsection{Building the Grasp Wrench Space}
\label{subsec:gws}
So far, the frictional constraints are for a contact surface with respect to a local contact frame. 
We now define the contact wrench constraints $\{\contactw_1 \ldotsc \contactw_M\}$ by combining the constraints for the frictional and the normal wrench, where the latter is the wrench impressed by the pressure.
Then we express the contact wrench constraints with respect to the object frame $O$, whose origin is at the center of mass $\origin_{\text{COM}}$ of the grasped object.

Let us consider the $n$th contact with $n \in \{1 \ldotsc N\}$.
For instance, $N=2$ for a parallel-jaw gripper if both jaws are in contact with the object.
Given the profile pair $(\surface_n, \pressureunit_n)$, we compute the frictional constraints $\X_n$ and the normal wrench $\wrench_{\perp_n}$.
Next, let $\vec{t}_n\in \R{3}$ and $\mat{R}_n\in\R{3\times3}$ be the translation and rotation of the $n$th local contact frame relative to the object frame, respectively. 
Denoting $\wrenchresmpden_{n,m} \in \X_{n}$ as the $m$th frictional constraint of the $n$th contact, we compute the $m$th contact wrench constraint $\contactw_{n,m} \in \R{6}$ with respect to $\origin_{\text{COM}}$ with
\begin{equation}
\contactw_{n,m} = \begin{bmatrix} \mat{R}_n & \vec{0} \\ \hat{\vec{t}}_n \mat{R}_n & \mat{R}_n \end{bmatrix} \left(\wrenchresmpden_{n,m} +\wrench_{\perp_n} \right) \bigcdot F_n
\label{eq:contact_constraints}
\end{equation}
where $\hat{\vec{t}}_n \in \R{3\times3}$ is the cross product matrix of $\vec{t}_n$ and introduces an additional torque due to the change of frame. 
$\wrenchresmpden_{n,m}$ and $\wrench_{\perp_n}$ are multiplied by $F_n=\int_{\surface_n}\pressure_n\dS_n$ since the constraints are computed with $\pressureunit_n$.

Ferrari and Canny~\cite{ferrari.1992} proposed two ways to build the GWS: by upper bounding the magnitude of each contact force individually or by bounding the sum magnitude of the contact normal forces. 
We select the former since Krug~et~al.~\cite{krug2017grasp} suggested that the latter is over-conservative. 
The grasp wrench space $\W$ is
\begin{equation}
\W = \mbox{Conv}(\oplus_{n=1}^N\{\contactw_{n,1} \ldotsc \contactw_{n,m} \ldotsc \contactw_{n,M}\})
\label{eq:GWS}
\end{equation}
where Conv$(\bigcdot)$ denotes the convex hull and $\oplus$ is the Minkowski sum operation. 

\subsection{Prediction}
The algorithm predicts if a grasp can resist the external wrench $\wext \in \mathbb{R}^6$ by checking if the opposite wrench $-\wext$ is  inside the grasp wrench space $\W$.

Given $\W$ with $B$ facets, denoting $\hat{\vec{n}}_{i} \in \R{6}$ as the outward normal of the $i$th facet with $i\in \{1 \ldotsc B\}$, $\vec{a}_i \in \R{6}$ as a point in the hyperplane of the facet, the prediction ${y}$ is 
\begin{equation}
y = 
\begin{dcases*}
	1 & if $-\wext^T\hat{\vec{n}}_i < \vec{a}_i ^T\hat{\vec{n}}_i, \forall i \in \{1 \ldotsc B\}$ \\
	0 & otherwise.
\end{dcases*}
\label{eq:prediction}
\end{equation}
Algorithm~\ref{algo:gws_predict} summarizes the process to predict the binary grasp success.

%% file: algos/algo-ls.tex
\renewcommand{\algorithmicensure}{\textbf{Output:}}
\begin{algorithm}[t]
\caption{Finding a 6D limit surface model for a nonplanar surface contact.}
	\begin{algorithmic}

	\REQUIRE 
	\STATE - Friction coefficient $\mu$
	\STATE - Contact surface $\surface$ represented with a parametric form or a mesh
	\STATE - Pressure distribution $\pressure$ 
	\ENSURE The 6DLS model $\A$ and $\boldsymbol{w}_\text{max}$ 
	
	\STATE $\left(\linedir_i, \linepoint_i, h_i \right)$ or $\linedir_i$, $i\in\{1 \ldotsc \nprs\}\gets$ Sampling $\left(\linedir, \linepoint, h \right)$ or $\linedir$ for possible motions (unit twists)
	\STATE $\origin \gets \surface,\pressure$ // \eqsref{eq:origin}
	\FOR{$i \gets 1$ to $\nprs$}
	    \STATE $\moment_i = \linepoint_i \times \linedir_i$
	    \STATE $\wrench_i \gets  \mu,  \pressure,\surface, \origin, \linedir_i, \moment_i,h_i$ // \eqsref{eq:velocity}--(\ref{eq:frictionArea})
	\ENDFOR
	\STATE // {Normalizing the frictional wrenches}
	\STATE $\wrench_\text{max} \gets \{\wrench_1 \ldotsc \wrench_\nprs\}$
	\STATE $\{\normalizedwrench_1 \ldotsc \normalizedwrench_\nprs\} \gets \wrench_\text{max},\{\wrench_1 \ldotsc \wrench_\nprs\}$ //  \eqsref{eq:normalization}
	\STATE // Fit a 6DLS model to the normalized frictional wrenches
	\STATE $\A \gets \{\normalizedwrench_1 \ldotsc \normalizedwrench_\nprs\}$ // \eqsref{eq:ls_ellip} or \eqsref{eq:ls_quartic} for the ellipsoid or quartic 6DLS model, respectively
    \RETURN $\A$, $\wrench_\text{max}$
	\end{algorithmic}
	\label{algo:6dls}
\end{algorithm}

%

%

%% file: algos/algo-gws.tex
\renewcommand{\algorithmicensure}{\textbf{Output:}}
\begin{algorithm}[t]
\caption{Prediction of a $N$-contact grasp success.} 
	\begin{algorithmic}

	\REQUIRE 
	\STATE - Friction coefficient $\mu$
	\STATE - External disturbance $\wext$
	\STATE - Details of each contact
	\begin{itemize}
	    \item Contact surface $\surface$ 
	    \item Pressure distribution $\pressure$ 
	    \item Rotation $\mat{R}$ and translation  $\vec{t}$ between the local contact and the object frame 
	\end{itemize}
	\STATE - Dataset $\D$ of linear frictional constraints 
	\ENSURE prediction $y \in \{0,1\}$
	\FOR{$n \gets 1$ to $N$} 
		\STATE {$F_n = \int_{\surface_n} \pressure_n\dS_n, \pressureunit_n = \pressure_n / F_n$} 
		\IF{$\X_n \notin \D$} 
		\STATE // Find the 6DLS model $\A$ with Algorithm $\ref{algo:6dls}$
		    \STATE $\A,\wrench_\text{max} \gets \mu,\surface_n,\pressureunit_n$ 
			\STATE // Linearize the frictional constraints
			\STATE $\{\verte_{1} \ldotsc \verte_{M}\}\gets$ Sampling a 6D hypercube
			\STATE $\{\wrenchresmp_{n,1} \ldotsc \wrenchresmp_{n,M} \} \gets \A,\{\verte_{1} \ldotsc \verte_{M}\}$ // \eqsref{eq:linearize_4th}
			\STATE // Denormalize the linear frictional constraints
			\STATE $\X_n \gets \{\wrenchresmp_{n,1} \ldotsc \wrenchresmp_{n,M}\},\wrench_\text{max}$ // \eqsref{eq:denormailzation}
			\STATE $\D \gets \D \cup \X_n$ // $\text{ Store } \X_n \text{ to } \D$ 
		\ELSE
			\STATE $\X_n \gets D $ // $\text{Read the frictional constraints from } \D$
		\ENDIF
		\STATE {$\wrench_{\perp_n} \gets \surface_n, \pressureunit_n$} // Normal wrench with \eqsref{eq:normal_wrench}
		\STATE // Contact wrench constraints
		\STATE $\{\contactw_{n,1} \ldotsc \contactw_{n,M}\} \gets \X_n, \wrench_{\perp_n}, \mat{R}_n, \vec{t}_n,F_n$ //  \eqsref{eq:contact_constraints}
	\ENDFOR
	\STATE $\W \gets \{\contactw_{1,1} \ldotsc \contactw_{1,M}\} \ldotsc \{\contactw_{N,1} \ldotsc \contactw_{N,M}\}$ // \eqsref{eq:GWS}
	\STATE $y \gets \W, \wext$ // \eqsref{eq:prediction}
	\RETURN $y$
	\end{algorithmic}
	\label{algo:gws_predict}
\end{algorithm}

%% file: figs/figParamSurfaces.tex

\begin{figure*}
	\centering	
	\subfloat[$\surface_1$: cylinder.]{{\includegraphics[height=7em]{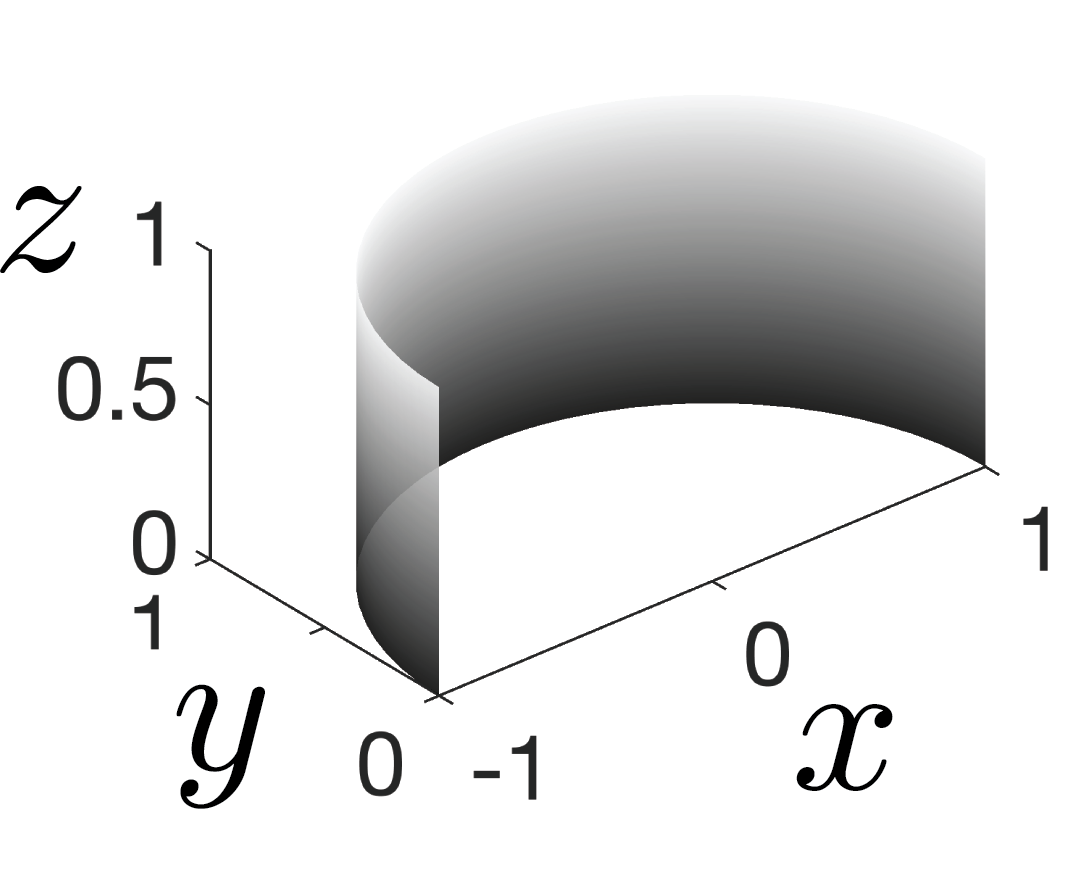} }}%
	\subfloat[$\surface_2$: elliptic cylinder.]{{\includegraphics[height=7em]{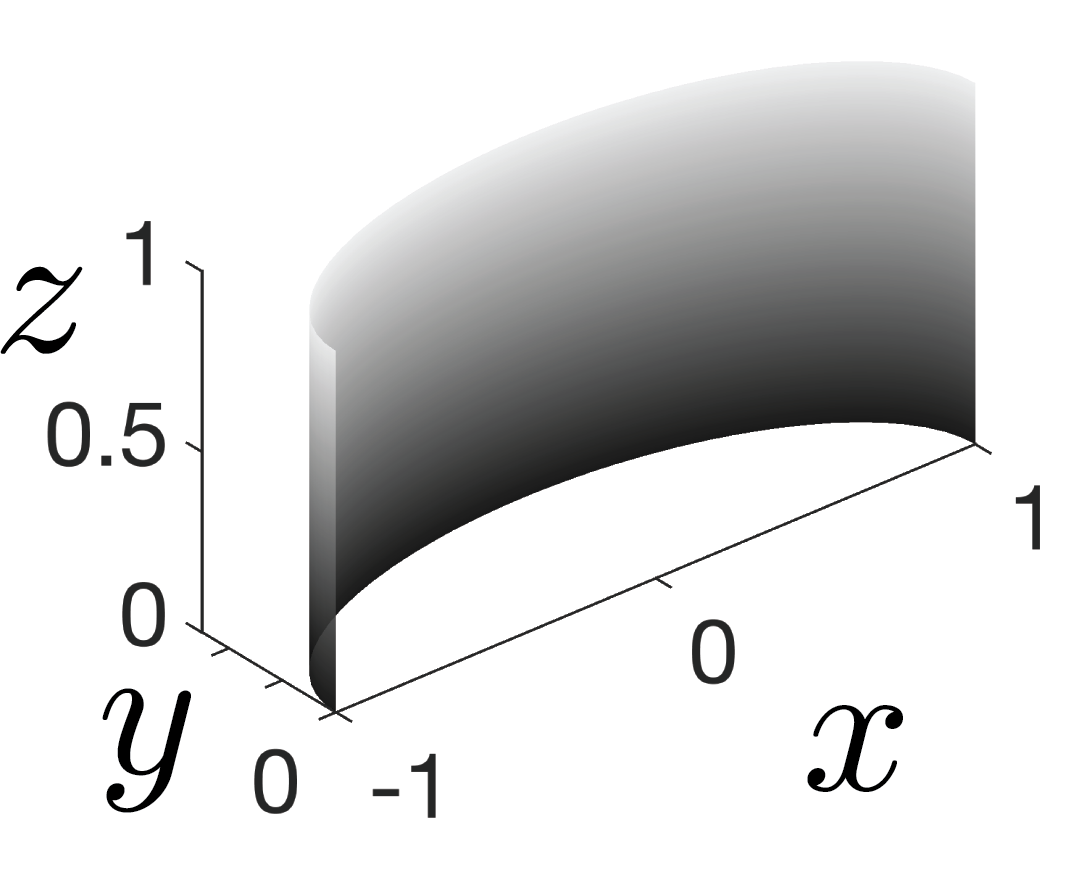} }}%
	\subfloat[$\surface_3$: sphere.]{{\includegraphics[height=7em]{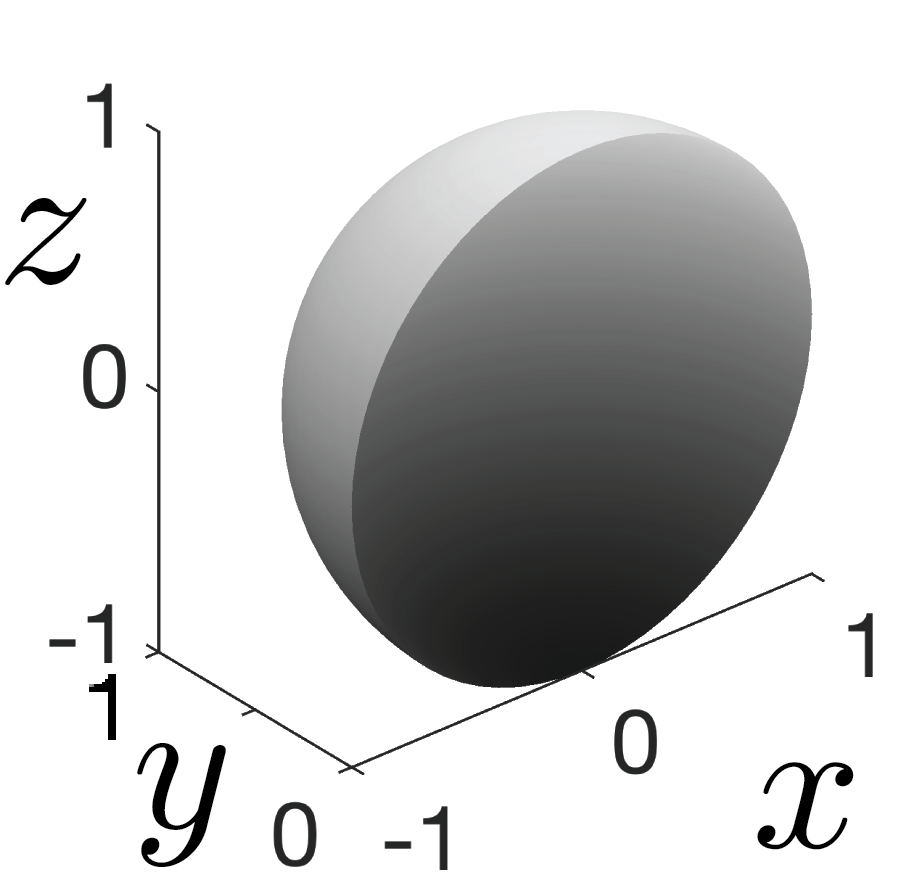} }}%
	\subfloat[$\surface_4$: ellipsoid.]{{\includegraphics[height=7em]{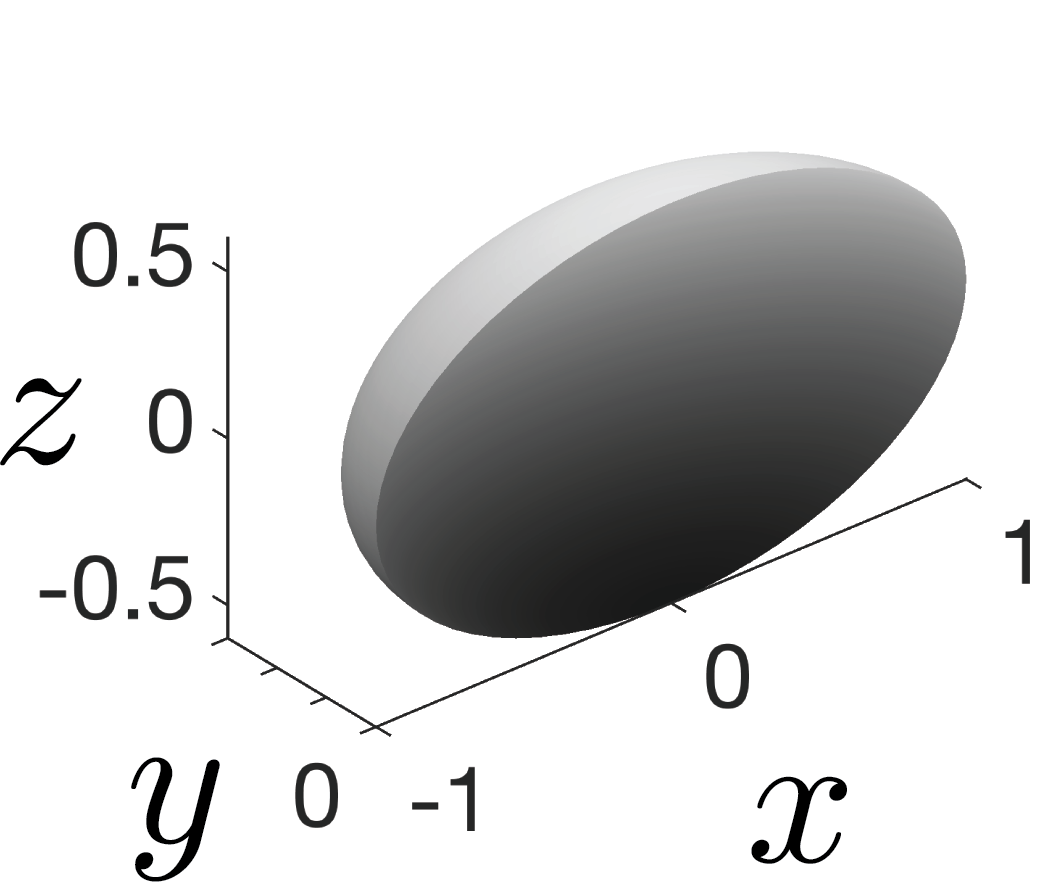} }}%
	\subfloat[$\surface_5$: paraboloid.]{{\includegraphics[height=7em]{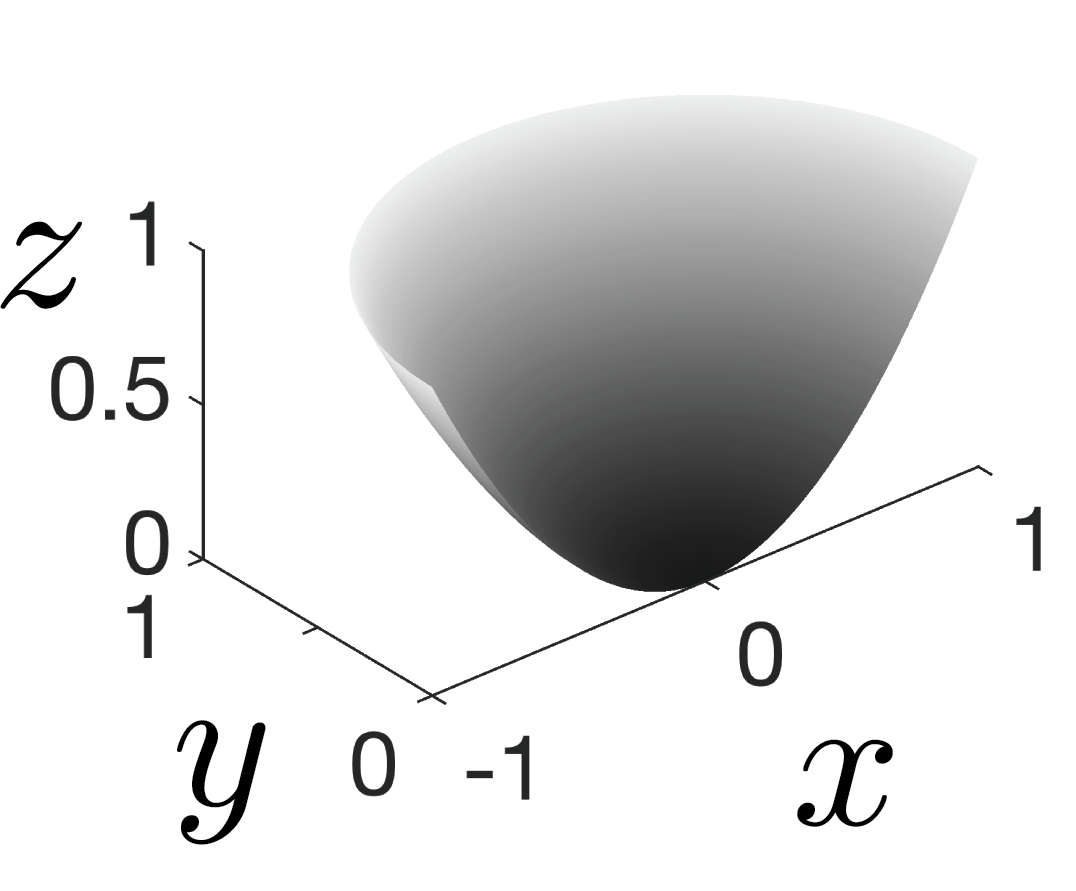} }}%
	\subfloat[$\surface_6$: elliptic paraboloid.]{{\includegraphics[height=7em]{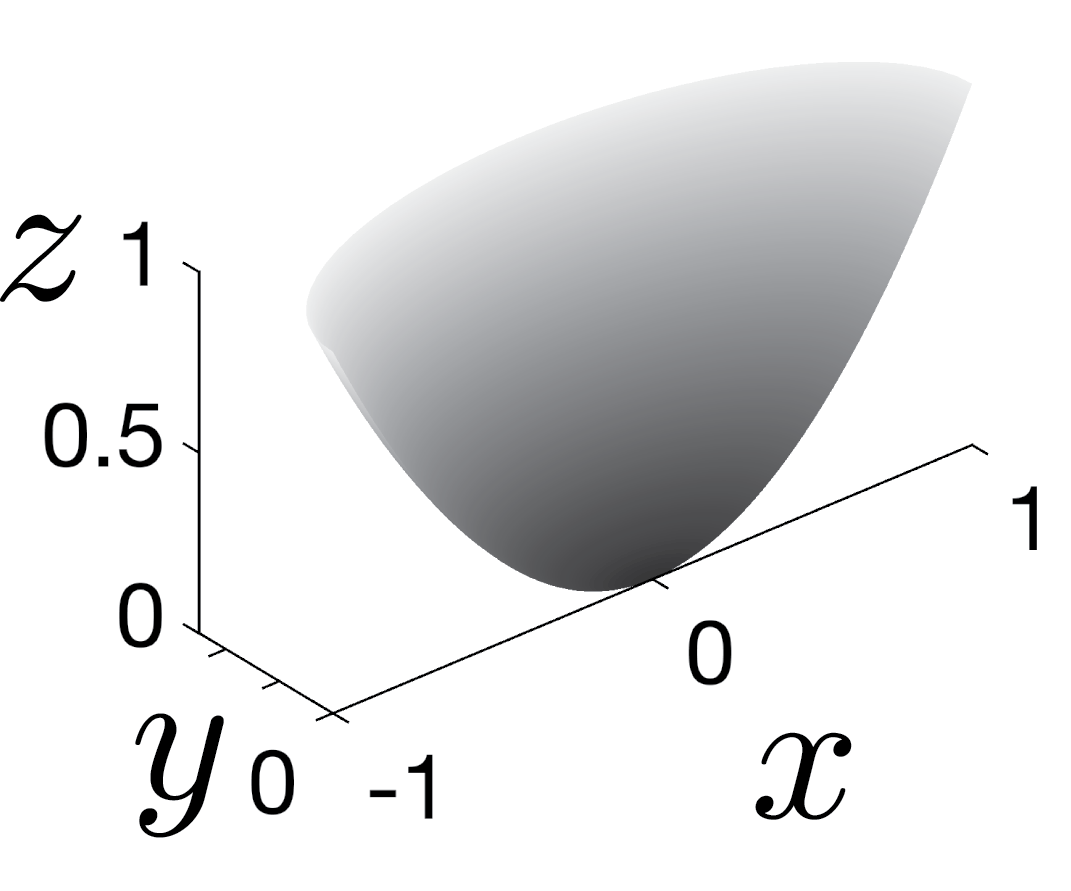} }}%
	\caption{\small Six parametric contact surfaces used to evaluate the 6D frictional wrenches and the proposed 6DLS models. 
	}
	\label{fig:parametric_surfaces}%
\end{figure*}


%% file: sections/11-simulations.tex
\section{Simulation}

In this section, we evaluate the performance of the two proposed limit surface models for nonplanar parametric and discrete contact surfaces obtained from FEM simulations.


\input{tabs/tableParametricSurfaces}

\input{tabs/tableFrictionParametricSurfaces}

\subsection{6DLS Models Evaluation with Parametric Contact Surfaces}
\label{subsec:paramSurface}
We start the evaluation with six parametric contact surfaces $\surface_1 \ldotsc \surface_6$, as illustrated in  \fref{fig:parametric_surfaces}.
\tref{tab:parametric_surfaces} summarizes the parametric form of each surface.
Although the coordinate system can be arbitrarily chosen, we selected the $x,y,z$-axes such that the surfaces are symmetric about the $yz$-plane for easier comparison. 
We used the friction coefficient $\mu=1$ and a normalized uniform pressure distribution $\pressureunit_i$ with $\int_{\surface_{i}} \pressureunit_i\dS_i=1$ for $\surface_i$, where $i\in\{1\ldotsc 6\}$. 
This selection allows for a more intuitive comparison between the frictional wrenches computed with the proposed nonplanar and the traditional planar surface contact model. 

\subsubsection{Maximal Frictional Wrench}
\label{subsubsec:sim_maximal_wrench}
For each contact surface, we computed $\wrench_\text{max}$, the wrench consists of the maximal magnitudes of the frictional wrenches in the six dimensions with respect to the friction center. 
Table \ref{tab:wrench_parametric_surface} shows $\wrench_\text{max}$ for $\surface_1$.

With the nonplanar surface contact model, the magnitude of ${f_{x}}$ reaches the maximum if the twist is a pure translation along the (negative) $x$-axis, which means the instantaneous screw axis $\lineISA$ is parallel to the $x$-axis and $\normang =0$. 
The local frictional force $\diff \vec{f}$ at each point is antiparallel to the projection of the linear velocity, or in this case the (negative) $x$-axis, onto the surface tangent plane at this point.
Similarly, the magnitudes of $f_y$ and $f_z$ are maximized if $\lineISA$ is parallel to the $y$- and $z$-axes with $\normang =0$, respectively. 
Note that $f_{z,\text{max}} = 1$ as the $z$-axis is in the tangent plane of each point since there is no curvature along the $z$-axis for $\surface_1$. 
$f_{x, \text{max}}$ and $f_{y, \text{max}}$ are equal for $\surface_1$ as the integrals of the projected $x$- and $y$-axis are identical.

We also computed $\wrench_\text{max}$ for $\surface_1$ with the 3D nonplanar model from our previous work~\cite{xu.2017} and  the traditional planar contact model.  
The former computes only the three largest components $f_x$, $f_z$ and $\tau_y$ for a nonplanar surface and sets the remaining three components to zero, resulting in an over-conservative friction estimation compared to the proposed 6D nonplanar surface model. 
For the latter, the planar surface $\surface_\text{pl}$ is obtained by projecting $\surface_1$ onto the $xz$-plane. 
We normalized the uniform pressure distribution such that $\int_{\surface_\text{pl}} \pressureunit_\text{pl}\dS_\text{pl} = 1$.
For $\surface_1$, the planar surface contact model results in an overconfident estimation of $f_x$ and an over-conservative estimation of $f_y, \tau_x, \tau_y, \tau_z$. 
Furthermore, we performed a principle component analysis (PCA) on the sampled normalized frictional wrenches $\{\normalizedwrench_1 \ldotsc \normalizedwrench_\nprs\}$ for all the studied parametric surfaces.
In all cases, we observed five or even six significant components, which is another indication that a traditional 3D limit surface is not sufficient for nonplanar surface contacts.

\input{figs/figDiscretizationEffect}

\input{figs/figLSParametricSurfaces}

\subsubsection{Surface Discretization Effect}
\label{sec:discretizationEffect}
While the friction computation for a parametric surface is less efficient than for a meshed surface due to the integral operation, the frictional wrench for a meshed surface can be less accurate depending on the number of elements. 
Therefore, we analyzed the runtime and $\wrench_\text{max}$ for the continuous surfaces $\surface_1 \ldotsc \surface_6$ and their meshes with 25--1,000 triangular elements.
We use $\wrench_\text{max}$ for the continuous surfaces as the ground truth.
To evaluate the error of $\wrench_\text{max}$ caused by the surface discretization, we define the \emph{wrench error rate}, which is the difference of $\wrench_\text{max}$ divided by the ground truth and averaged over all dimensions of $\wrench_\text{max}$.
As shown in \fref{fig:discretizationEffect}, the wrench error rate of all surfaces rapidly decreases with the number of elements since the meshed surfaces are closer to the parametric ones.
The error rate is below 4\% with 300 triangles, which is acceptable in most applications and is nearly zero with 1,000 elements.
$\surface_1$ and $\surface_2$ have the lowest error since there is no curvature along the $z$-axis.

We further measured the runtime on an Ubuntu 16.04 machine with an Intel Core i7-8700K CPU (3.7 GHz) with a MATLAB implementation without parallel computing or GPU acceleration. 
While the average runtime to integrate a single wrench is 2.62s, the computation for a discrete surface requires 1.3ms and achieves a 2,000-times speedup.

\input{figs/figLSExamples}

\input{figs/figFEM}

\subsubsection{6DLS Model Evaluation}
\label{sec:Fitting_results_parametric_surface}
\mysum{LS visualization}
We evaluate the 6DLS models by fitting them to different numbers of normalized frictional wrenches computed with the six parametric surfaces.
\fref{fig:lsErrorParametricSurface} shows the mean wrench fitting error and the corresponding runtime. 
The wrench error is measured as the mean distance of the wrenches to the LS and is computed with 20,000 normalized frictional wrenches with \eref{eq:fitting_error}.
While being less efficient, the quartic achieves a lower fitting error with more than 200 wrenches as a quartic has 126 variables and an ellipsoid has 21. 
An appropriate LS model can be selected based on the trade-off between the fitting error and the runtime required by the application. 
\fref{fig:LSExamples} shows a representative 3D cross-section of the quartic and the ellipsoidal 6DLS model, where the remaining three components are zero.
Each LS model is fit to 600 frictional wrenches (orange dots) computed with $\surface_4$ (ellipsoid). 
We observe that the wrenches are closer to the quartic surface compared to the ellipsoidal one.  
Large differences are visualized with dashed rectangles.

\subsection{6DLS Models Evaluation with Discrete Contact Surfaces}
\label{subsec:FEM}
We now evaluate the two 6DLS models with a large variety of contact profiles obtained from FEM simulations. 
We simulated contacts between a parallel-jaw gripper and thin-walled objects using the commercial software ANSYS~\cite{ansys} based on our previous work~\cite{alt.2016}.
The objects for simulations are rigid, similar to closed plastic bottles.

The object geometry is described with nonuniform rational B-Splines (NURBS), which is later meshed according to quality preferences.
The top of \fref{fig:FEM_sim}(a) depicts the object geometry generation.
An ellipse defines the base and a spline curve shapes the wall, which is controlled by the variables $p_1 \ldotsc p_{12}$.
As illustrated in \fref{fig:FEM_sim}(a) bottom, we vertically sampled the spline curve and created an ellipse at each sample.
The object surface is created by ruling the adjacent ellipses; the hatched surface at the bottom of \fref{fig:FEM_sim}(a) shows a representative ruled surface. 
The squares and circles mark the locations of the antipodal grasps with two approach directions, which are spread vertically.
By varying the 12 variables, we efficiently generated 24 objects with different geometries, as shown in  \fref{fig:FEM_sim}(b).

\input{tabs/tableFrictionFEM}
We selected the rectangular gripper jaws, where a soft silicon pad with 5mm thickness is attached to each jaw, as gripper jaws with compliant materials are widely used in robot grasping \cite{danielczuk2019reach,harada2014stability} to increase grasp robustness.
\fref{fig:FEM_sim}(c) shows the meshed bodies (left) and the nodal solution (right) of a representative FEM simulation.
We applied a displacement (orange arrow) as load to each jaw and the displacements are parallel to the grasp axis. 
For each grasp location, we applied three displacements in the range of 1mm--3mm, resulting in a total of 2,932 grasps. 
\fref{fig:FEM_sim}(d) shows representative contact profiles with interpolated pressure values.  
Each contact surface consists of 3$\times$3 rectangular elements and each element has a single pressure value.
For each contact profile, we sampled 600 frictional wrenches and fit the two 6DLS models to the normalized wrenches. 
\tref{tab:FEM} summarizes the means and standard deviations of the fitting errors. 
Similar to the results for the parametric surfaces in \sref{sec:Fitting_results_parametric_surface}, the quartic model yields a lower fitting error.
The low standard deviations of both models suggest that the proposed 6DLS models achieve consistent performance and are suitable for a large variety of contact profiles.

%% file: tabs/tableParametricSurfaces.tex
\begin{table}[t]
	\begin{center}

		\renewcommand{\arraystretch}{1.5} 
		\caption{Parametric form of the contact surfaces.}
		\begin{tabular}{| c | c | c |}
			\hline
			ID & Type &  Parametric form \\[5pt] \hline
			$\surface_1$ & cylinder & \makecell{$\left[\cos u, \sin u, v\right]^\trans$\\$(u,v) \in [0,\pi] \times  [0,1]$} \\[5pt]  \hline 
			
			 $\surface_2$ & \makecell{elliptic \\ cylinder} & \makecell{$\left[a \cdot \cos u, b\cdot\sin u, v\right]^\trans$  \\ $(u,v)  \in [0,\pi]\times [0,1]$,  $a = 1, b = \frac{1}{2} $}  \\[5pt]  \hline
			
			$\surface_3$ & sphere &  \makecell{$\left[\cos u\cdot\cos v, \cos u\cdot\sin v, \sin u\right]^\trans$ \\ $(u,v) \in [-\frac{1}{2} \pi,\frac{1}{2} \pi] \times [0,\pi]$}   \\[5pt] \hline 
			
			$\surface_4$ & ellipsoid & \makecell{$\left[a\cdot\cos u\cdot\cos v, b\cdot\cos u\cdot\sin v, c\cdot\sin u\right]^\trans$ \\ $(u,v) \in [-\frac{1}{2}\pi,\frac{1}{2}\pi] \times [0,\pi]$,  $a = 1, b = \frac{1}{2} $,  $c = \frac{3}{5} $}  \\[5pt] \hline
			
			$\surface_5$ & paraboloid & \makecell{$\left[\cos u \cdot v, \sin u \cdot v, v^2\right]^\trans$ \\ $(u,v) \in [0,\pi] \times [0,1]$}  \\[5pt] \hline
			
			$\surface_6$ & \makecell{elliptic \\ paraboloid} & \makecell{$\left[a\cdot\cos u\cdot v, b\cdot\sin u \cdot v, v^2\right]^\trans$ \\ $(u,v) \in [0,\pi]\times[0,1]$,  $a = 1, b = \frac{1}{2} $}  \\[5pt] 
			\hline
		\end{tabular}
		\label{tab:parametric_surfaces}
	\end{center}
\end{table}

%% file: tabs/tableFrictionParametricSurfaces.tex
\begin{table}
	\begin{center}
		\caption{Maximal magnitudes of the frictional wrenches in the six dimensions for $\surface_1$ computed with three contact models.}
		\begin{tabular}{|c|c|c|c|c|c|c|}
			\hline
			 \makecell{Contact \\ model}  & \multirow{1}{*}{$f_{x,{\text{max}}}$} & \multirow{1}{*}{$f_{y,{\text{max}}}$} &\multirow{1}{*}{$f_{z,{\text{max}}}$}&\multirow{1}{*}{$\tau_{x,{\text{max}}}$}&\multirow{1}{*}{$\tau_{y,{\text{max}}}$}&\multirow{1}{*}{$\tau_{z,{\text{max}}}$} \\ \hline
		
			\makecell{6D \\ nonplanar} & 0.64 & 0.64 & 1.00 & 0.34 & 0.69 & 0.59\\ \hline
			\makecell{3D \\ nonplanar \\ \cite{xu.2017}} & 0.64 & 0.00 & 1.00 & 0.00 & 0.69 & 0.00 \\ \hline
			{Planar} & 1.00 & 0.00 & 1.00 & 0.00 & 0.59 & 0.00\\ \hline
			 
		\end{tabular}
		\label{tab:wrench_parametric_surface}
	\end{center}
\end{table}

%% file: figs/figDiscretizationEffect.tex
\begin{figure}
	\centering	
	\subfloat{\includegraphics[width=0.8\linewidth]{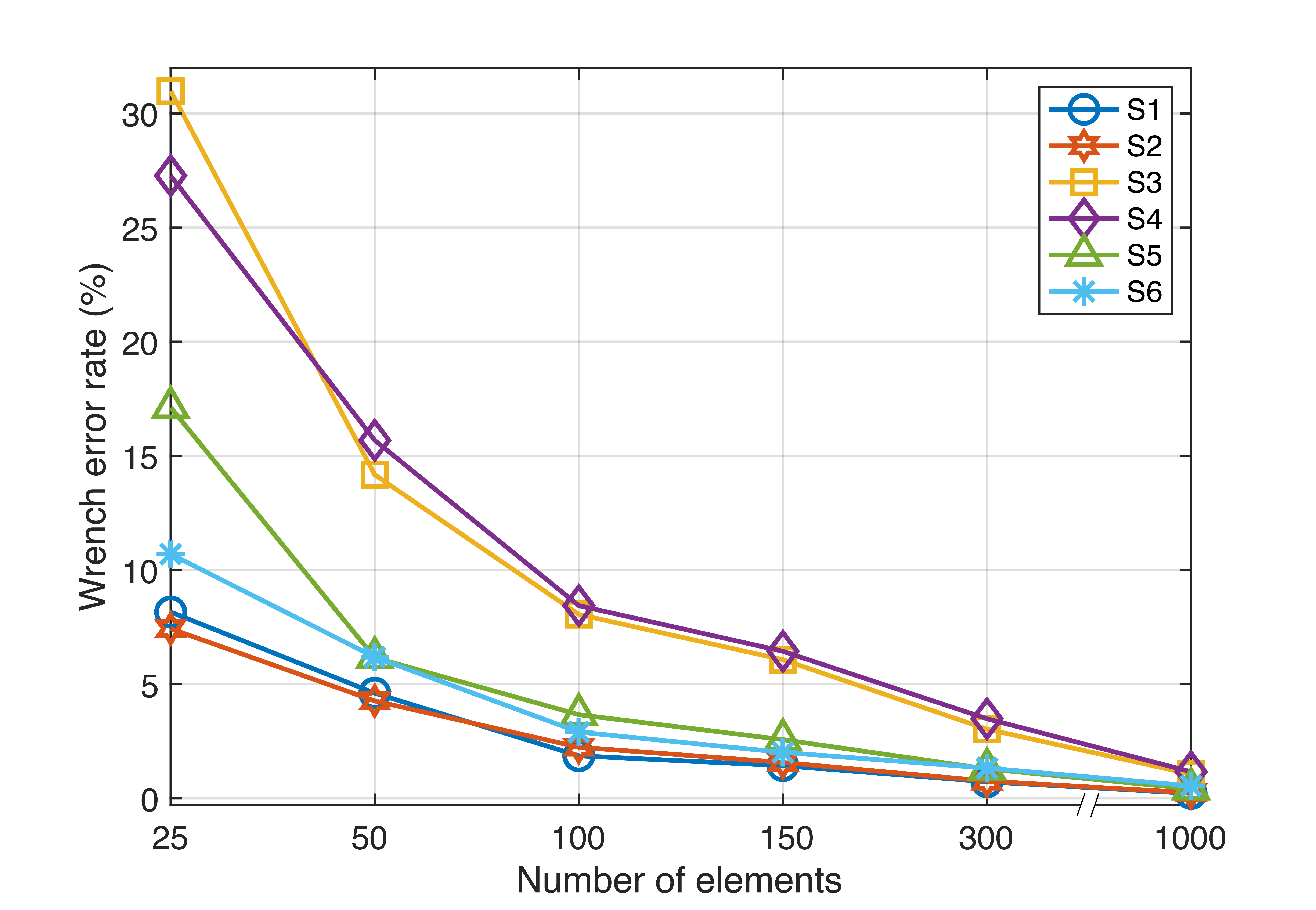}}%
	\caption{\small Wrench error rate of the six meshed surfaces with an increased number of triangular elements.} 
	\label{fig:discretizationEffect}
\end{figure}

%% file: figs/figLSParametricSurfaces.tex
%
%
%
%
\begin{figure}
	\centering	
	\subfloat[]{{\includegraphics[width=0.48\linewidth]{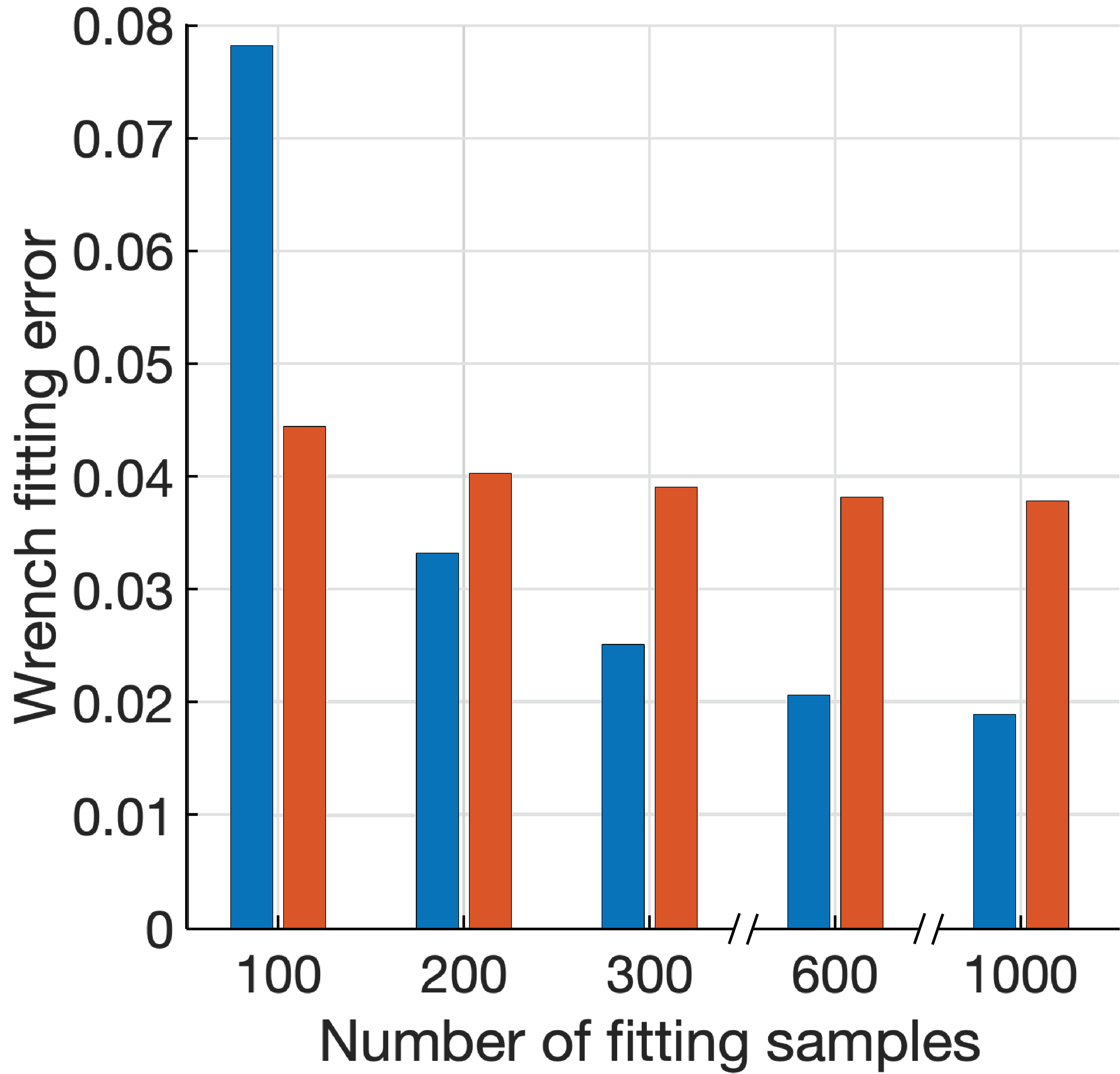}  }}%
	\subfloat[]{{\includegraphics[width=0.48\linewidth]{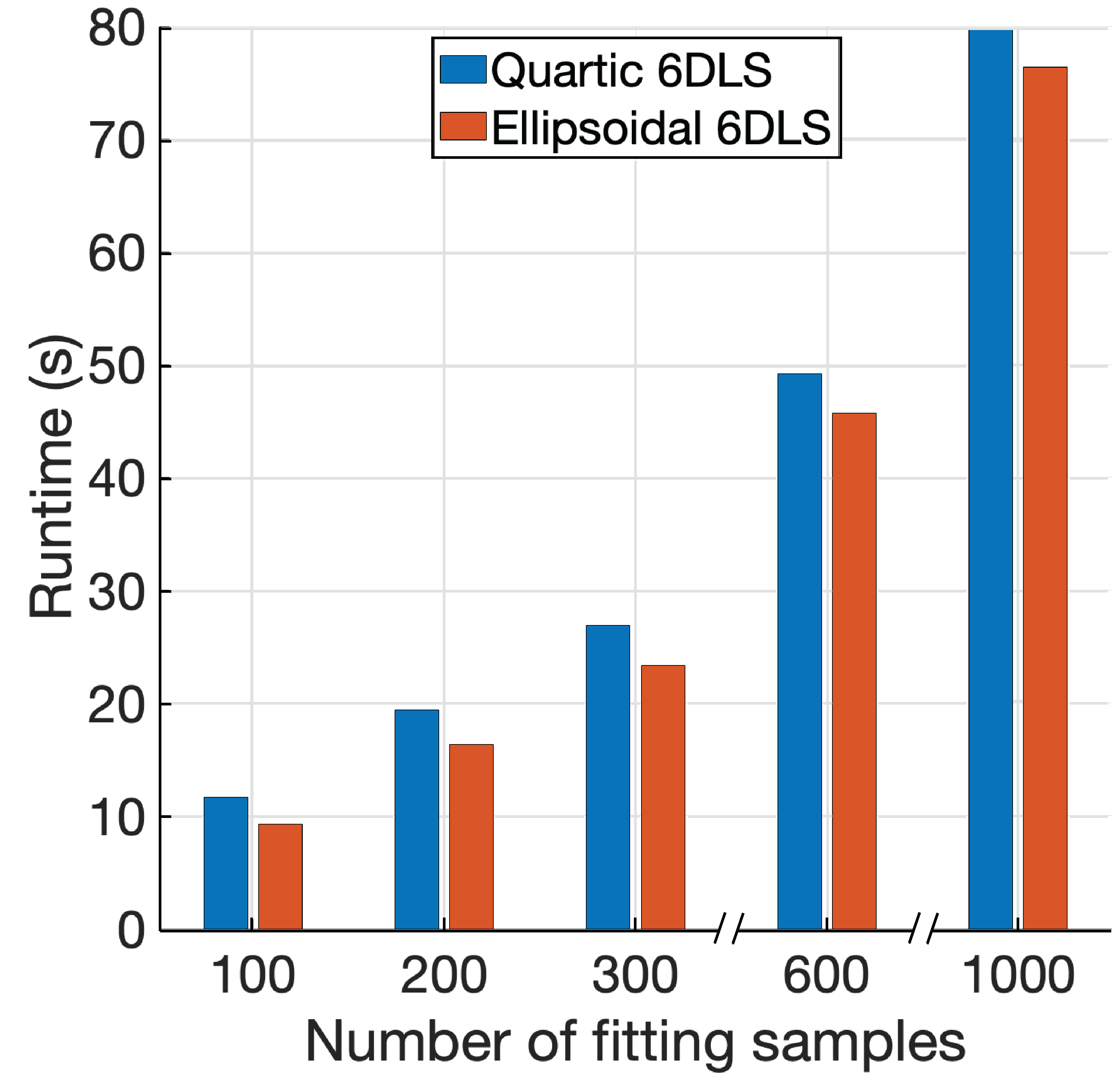}  }}%
	\caption{\small Wrench error and runtime of the quartic and the ellipsoidal 6DLS model fit to 100--1,000 frictional wrenches.}
	\label{fig:lsErrorParametricSurface}%
\end{figure}

%% file: figs/figLSExamples.tex
\begin{figure}
	\centering	
	\subfloat[]{{\includegraphics[width=0.48\linewidth]{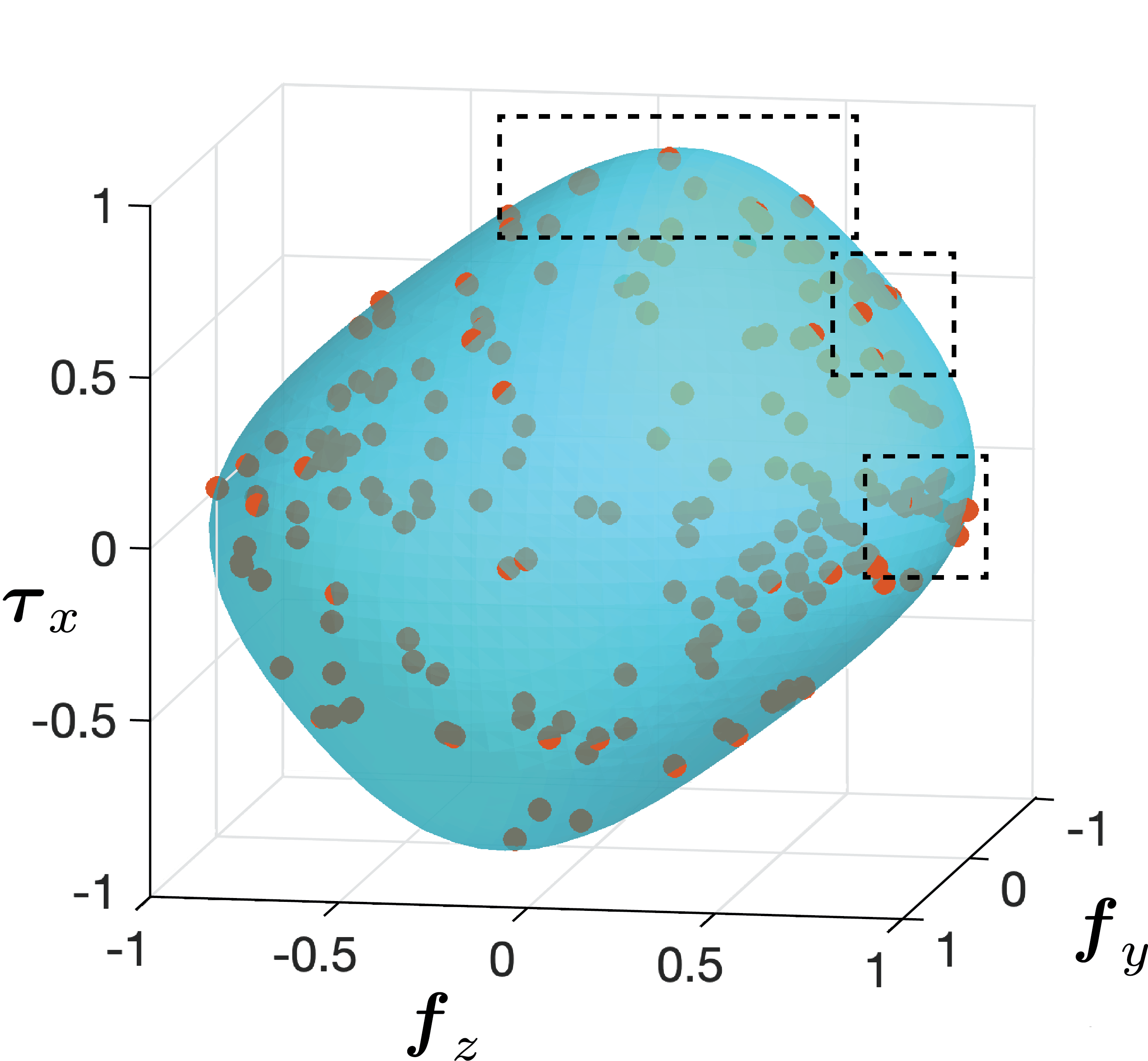} }}
	\subfloat[]{{\includegraphics[width=0.48\linewidth]{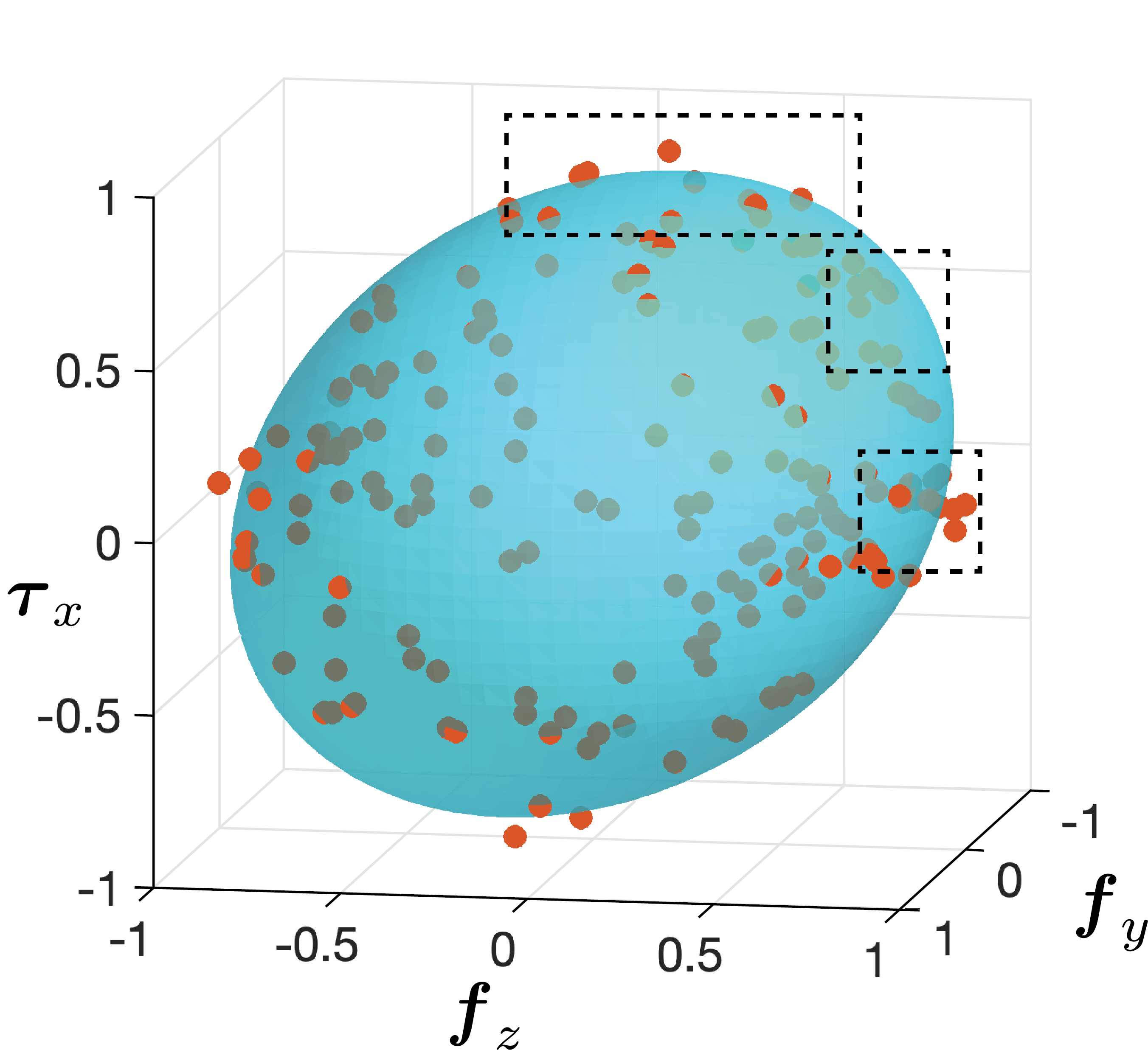} }}	
	\caption{\small A representative 3D cross-section of (a) the quartic and (b) the ellipsoidal 6DLS fit to the normalized  frictional wrenches (orange dots).} 
	\label{fig:LSExamples}%
\end{figure}

%% file: figs/figFEM.tex
\begin{figure*}
	\centering	
	\subfloat[Model generator.]{{\includegraphics[height=55mm]{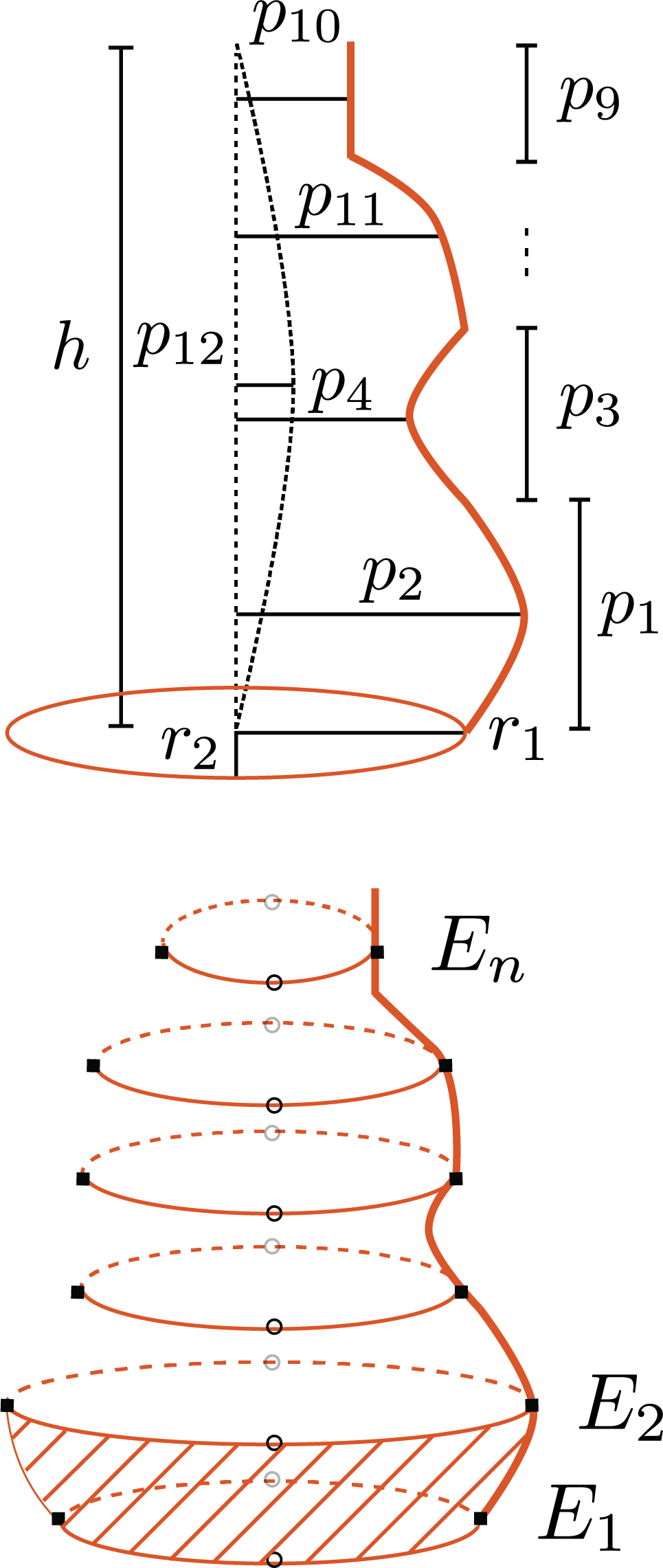} }}%
	\hfill
	\subfloat[Rigid objects.]{{\includegraphics[height=55mm]{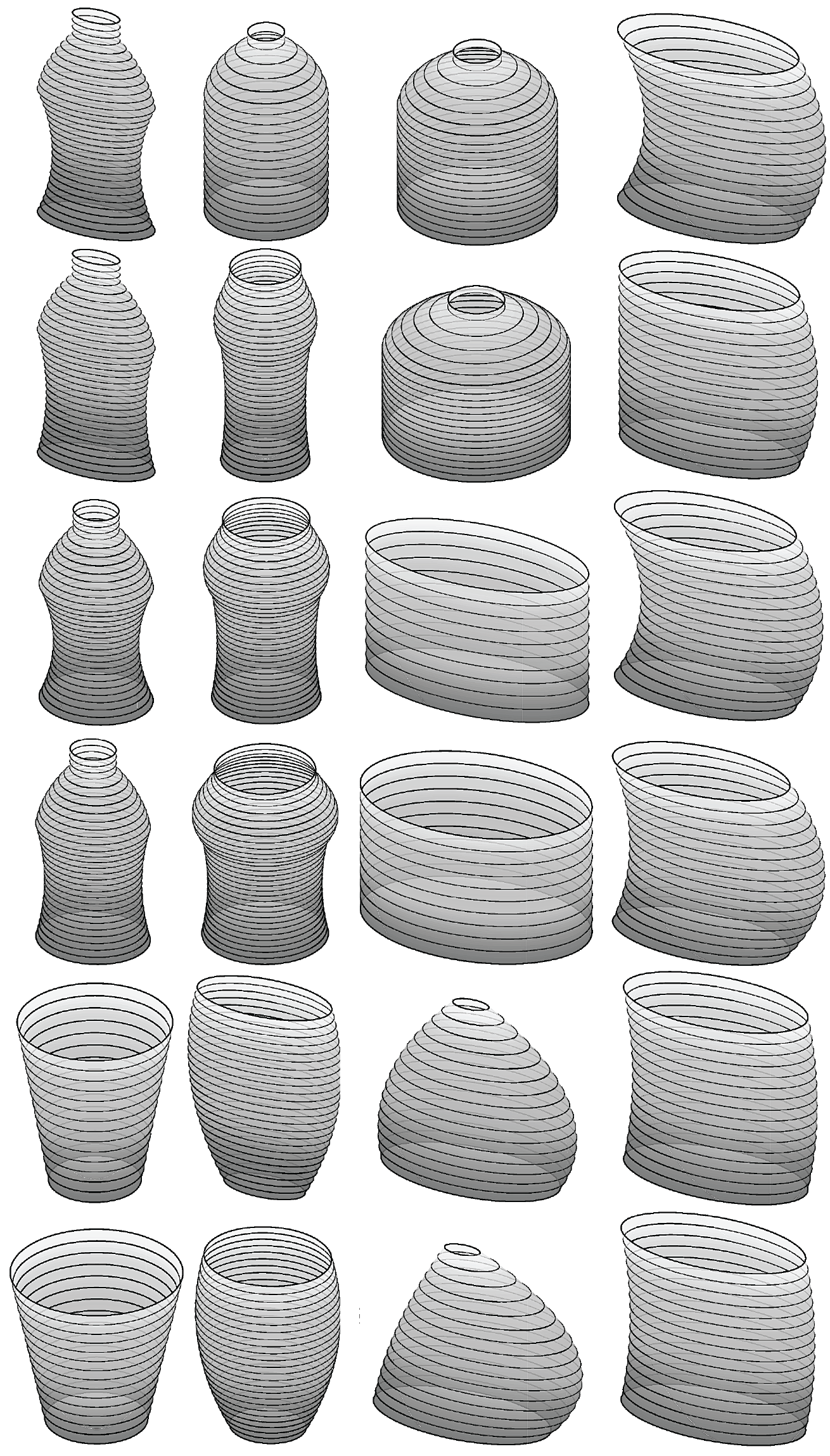} }}%
	\hfill
	\subfloat[A representative FEM simulation.]{{\includegraphics[height=55mm]{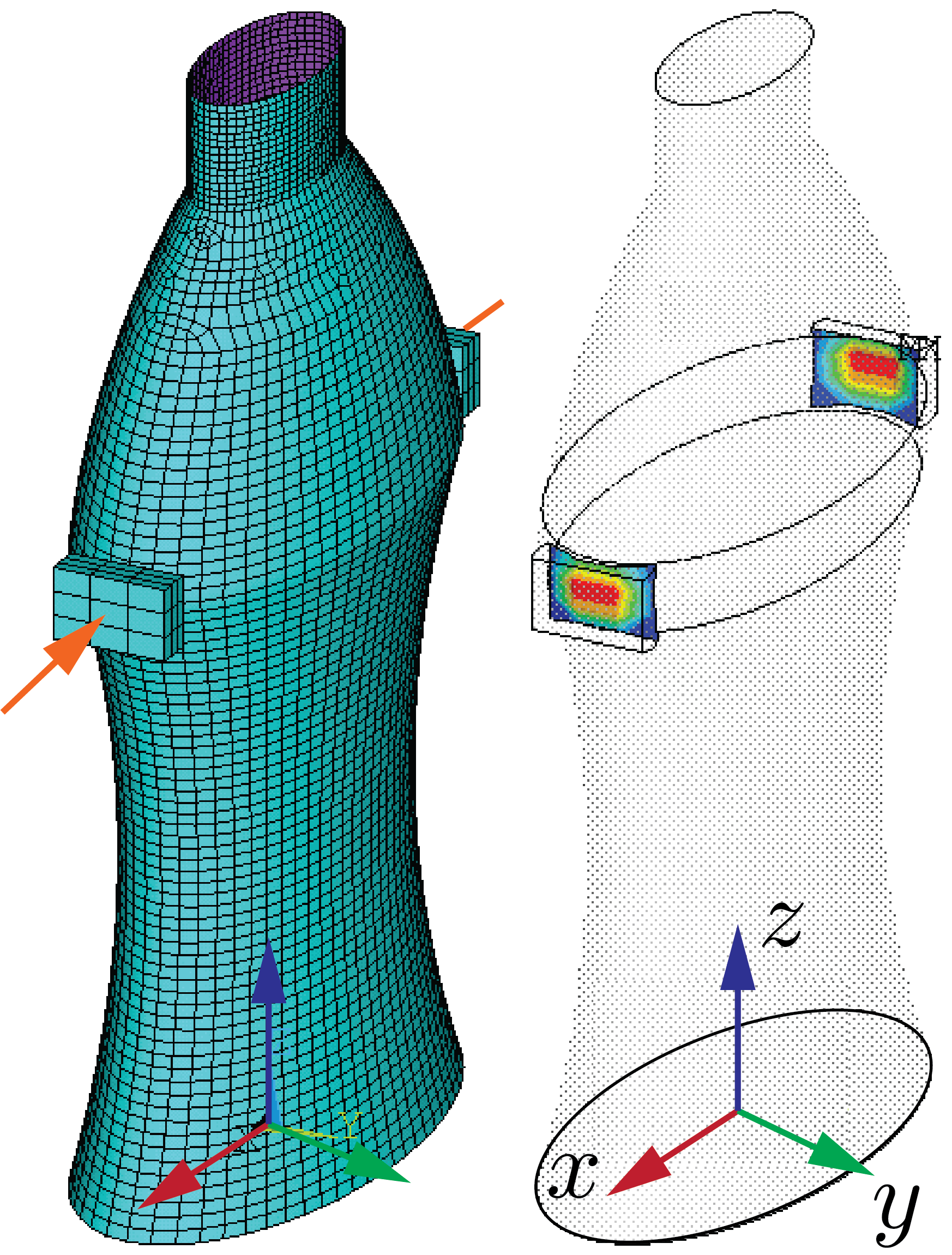} }}%
	\hfill
	\subfloat[Representative contact profiles.]{{\includegraphics[height=55mm]{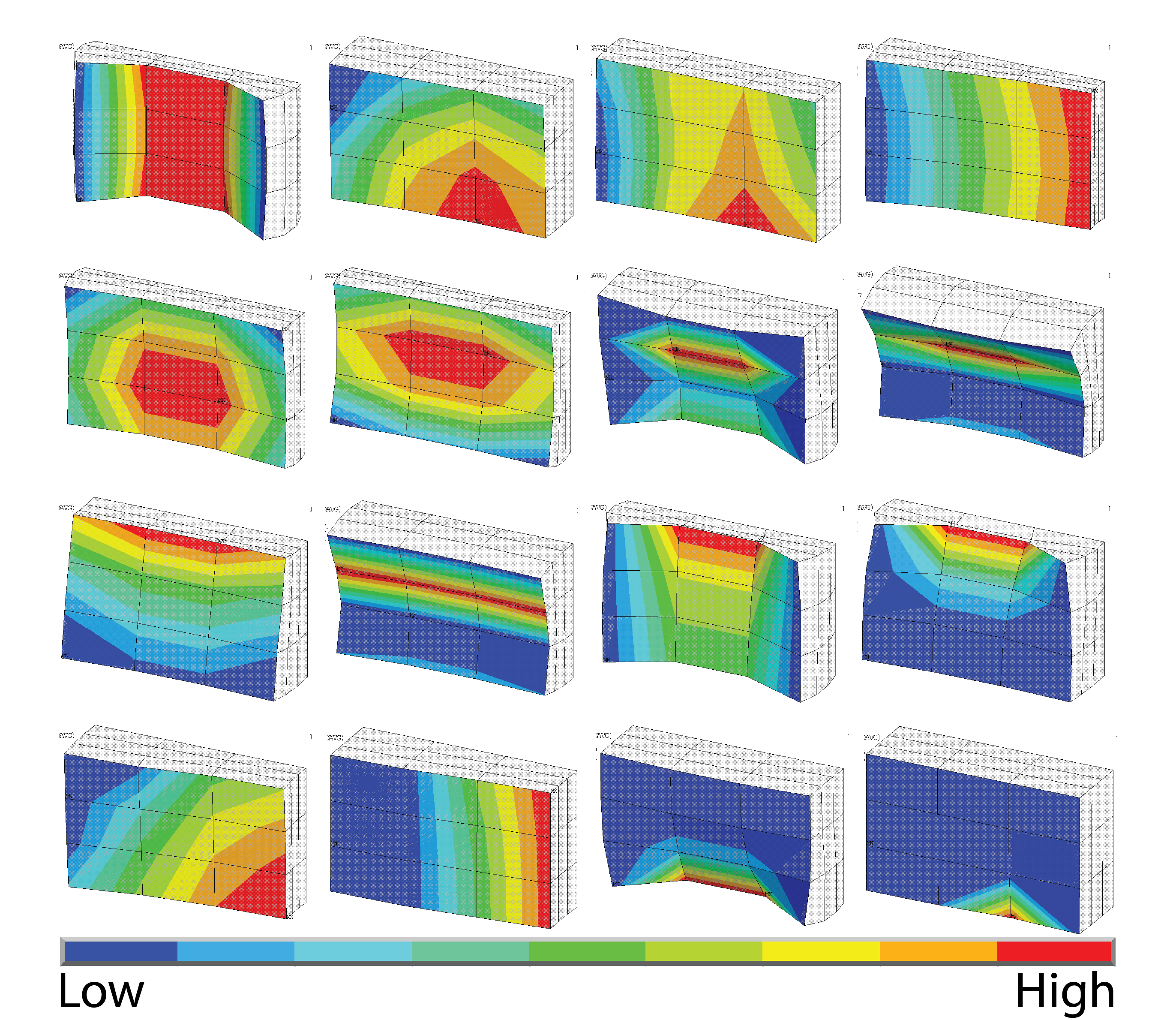} }}%
	\hfill
	\caption{\small FEM simulations to create a large variety of contact profiles to evaluate the 6DLS models. (a) Object model generator (Figure recreated from \cite{alt.2016}). (b) 24 rigid objects used for the simulations, where eight objects are asymmetric.  (c) Meshed objects and the nodal solution of a representative FEM simulation. (d) Representative contact profiles obtained from the simulations, where red means high pressure. }
	\label{fig:FEM_sim}%
\end{figure*}

%% file: tabs/tableFrictionFEM.tex
\begin{table}
	\begin{center}
		\caption{Fitting error of the 6DLS models fit to frictional wrenches computed with contact profiles from 2,932 FEM simulations.}
		\begin{tabular}{|c|c|}
			\hline
			 {Quartic model} & {Ellipsoidal model} \\ \hline
			 \textbf{0.011 $\pm$ 0.006} & 0.019 $\pm$ 0.005 \\
			\hline
		\end{tabular}
	\label{tab:FEM}
	\end{center}
\end{table}

			

%% file: sections/12-experiments.tex
\section{Experiments}
\label{subsec:roboExp}
We apply the 6DLS models to predict physical grasp success for a vertical lifting task.
Given the external wrench disturbance $\wext$, the friction coefficient $\mu$, the gripper pose, and an estimated contact profile for each gripper jaw, the algorithm predicts if the grasp can counterbalance $\wext$ by checking if the opposite of $\wext$ is in the grasp wrench space, as summarized in Algorithm~\ref{algo:gws_predict}. 

\subsection{Experiment Setup}
\label{subsec:exp_setup}
\fref{fig:setup} (left) depicts the setup of a KUKA robot arm and a SCHUNK parallel-jaw gripper mounted with customized deformable fin-ray jaws~\cite{festo} (blue).
We 3D printed rigid grasped objects to control the contact surface.
Estimation of the contact profiles is described in \sref{sec:profile}.

We further attached a 3D-printed mechanical assembly (pink) to the grasped object to create different wrench disturbances $\wext$ by mounting weight plates at various locations.
Such design allows large torques due to the long torque arms and efficient computation for $\wext$, and therefore, reduces uncertainties compared to grasping real life objects.

We define an object frame, as the GWS and $\wext$ are computed with respect to the origin of the object frame. 
We select the COM of the grasped object as the origin, instead of the COM of both the grasped object and the wrench disturbance assembly. 
This frame selection enables easier comparison between the grasp wrench spaces constructed with different contact models and does not affect the  predictions as one can select an arbitrary reference point to compute torques, and therefore, the GWS. 

\input{figs/figPhySetup}

We used two approach directions to create $\wext$.
\fref{fig:setup} (left) illustrates the $x,y,z$-axes of the object frame for a representative vertical and horizontal grasp with the $x$-axis parallel to the grasp axis.
The wrench disturbance assembly generates $\wext$ in the $(f_y,\tau_x,\tau_z)$- and $(f_z,\tau_x,\tau_y)$-space with the vertical and horizontal grasp direction, respectively.
We selected the locations of weight plates so that the disturbances are well-scattered in each space.

A force sensor is mounted on each gripper jaw to measure the grasp force along the $x$-axis. 
We also mounted an Intel RealSense SR300 RGBD camera (green) on the gripper to label the grasp success by tracking the object pose with the pcl library~\cite{pcl}. 
Specifically, we compared the object point clouds captured at two poses, when the object was grasped and when it reached the highest point of the vertical lifting task, and computed the pose change by using the Super4PCS algorithm~\cite{super4pcs}.
If the object rotation angle and translation are below a threshold pair, we label the grasp as a success.
However, we have observed that the jaws' deformation during the manipulation also leads to a changed object pose, even though there is nearly no relative motion between the object and the gripper jaws.
We selected 5$^\circ$ and 9mm as the threshold pair so that about half of the physical grasps are successful.
If the thresholds are high, the grasps will be labeled as a success even if there is a relative motion; whereas with low thresholds, the grasps will be labeled as a failure even if there is no relative motion but the gripper jaws deformed during the manipulation.
We also discuss the prediction results with different threshold pairs in \sref{subsec:sensitivity}, as robot applications have different tolerance of object motion during the manipulation.
While assembly tasks require minimal object motion, bin-picking allows larger object pose change.
In future work, we plan to use a tactile sensor to label grasp success by detecting slips.

\subsection{Contact Profile Estimation}
\label{sec:profile}
We estimate the contact profile, including the contact surface $\surface$ and the pressure distribution $\pressure$, for each gripper jaw.

\fref{fig:setup} (right) shows two types of contact surfaces created by ten 3D-printed rigid objects.
As illustrated on the top, the five object models of type I are cut from elliptic cylinders, whose horizontal radii are identical, whereas the vertical radii vary to change the surface curvature.
The cylinders are cut so that the contact surface is the same when the grasp force of each jaw is higher than a threshold (20N).
The contact surface is completely defined by the radii and the contact length $l_1$, which is depicted in \fref{fig:setup} (right).
If we directly use elliptic cylinders as the object model, we need to measure the contact surface for each trial as the  surface increases with the grasp force.
As shown on the bottom, each of the type II objects creates five or eight narrow planar contact surfaces with 3mm--5mm width.
We define the contact length $l_2$ of type II as the length of each narrow surface.  
The direction of frictional forces are constrained to lie in each planar surface, as described in  \aref{subsec:frictionDiscreteSurface}. 
Type II objects show that the contact surface can be nonplanar, even if the local contact surfaces are planar.
Such discrete nonplanar surfaces also occur, for instance, when a silicon jaw pad deforms to the corner of a rigid cube. 
Both types of objects create ten different contact surfaces in total as the grasp forces used in the experiments are larger than the threshold and each object creates one contact surface. 
We discuss the effect of contact lengths on grasp prediction results in \sref{subsec:sensitivity}.

\input{figs/figPhysFEM}

\input{tabs/tablePhysResults}

To estimate the pressure distribution, we first simulated the contacts between the fin-ray jaw and the rigid elliptic cylinders of type I using the FEM.
\fref{fig:FEMFesto}(a) shows the simulation results by applying a displacement on the jaw, as well as the interpolated pressure values of the contact. 
Although one can simulate each physical grasp for the contact profile, the system will be potentially computationally infeasible.
Furthermore, we observed a change of the pressure distribution with a small translation of the object pose along the $y$-axis; therefore, it is difficult to align the exact same object pose in the simulation and the experiment.
Hence, we approximated the pressure distribution with the power-law model proposed by Xydas and Kao~\cite{xydas.1999} based on the FEM simulation results.
As the power-law model is originally evaluated with planar circular contacts, we modified the model so that it applies to the contacts used in the experiments. 
For other types of contact surfaces, one can use the REACH model proposed by Danielczuk~et~al.~\cite{danielczuk2019reach}, which approximates the contact profile between a rigid object and a gripper jaw mounted with a deformable pad.

\fref{fig:FEMFesto}(a) shows that the pressure along the $z$-axis is nearly constant but varies along the $y$-axis. 
\fref{fig:FEMFesto}(b) shows the top view of the contact and the computation of the power-law model. 
We assume that the pressure distribution is symmetric about the center $\origin_g$ of the grasped object. 
Limitations of the assumption are discussed in \sref{sec:discussion}.
We express the pressure at a point as a function of the $y$ component of the distance between the point to $\origin_g$, denoted as $r\geq0$.
Let $r_{\text{max}}$ be the $y$ component of the maximal distance between any point on the contact surface to $\origin_g$ and $\pressureunit(r_\text{max}) = 0$, we obtain the normalized power-law pressure distribution with 
\begin{equation}
	\pressureunit(r) = \pressure_{0} \left[1- \left(\frac{r}{r_\text{max}}\right)^k\right]^{1/k}. 
	\label{eq:power}
\end{equation}
The exponent $k\in \mathbb{R}^+$ controls the shape of the pressure distribution and $\pressureunit(r)$ is a uniform pressure distribution with $k=\infty$.
$\pressure_{0}$ is a normalization constant such that $\int_\surface \hat{\pressure}(r)\dS=1$.
\fref{fig:FEMFesto}(c) illustrates the extracted contact surface  with a normalized pressure distribution $\pressureunit$ from the FEM simulation, where each element has a single pressure value.
\fref{fig:FEMFesto}(d) depicts the pressure values from \fref{fig:FEMFesto}(c) as a function of ${r}/{r_\text{max}}$ and a power-law model with $k=2.4$ fit to the pressure values. 
We observed that the pressure values are close to the curve, which suggests that the power-law model is an applicable approximation for the nonplanar contact surfaces used in these experiments.
However, we also observed that the exact $k$ value varies from 2.4 to 5.5 for the elliptic cylinders under different loads.
Therefore, we discuss the grasp success prediction results with different $k$ values in \sref{subsec:sensitivity}. 

We scaled $\pressureunit$ so that the normal force of each contact matches the force sensor reading $F_s$. 
As shown in \fref{fig:setup} (right), $F_s$ measures $\norm{{f}_{\perp_x}}$, the magnitude of the $x$ component of the normal force; therefore, $\pressureunit$ is scaled so that $\norm{{f}_{\perp_x}} = F_s$.
Thus, we computed $\norm{\hat{f}_{\perp_x}}$ with $\pressureunit$ using \eref{eq:normal_wrench} and obtained the pressure distribution $\pressure(r) = \lambda_p \bigcdot \pressureunit(r)$ with $\lambda_p = F_s/\norm{\hat{f}_{\perp_x}}$.
We precomputed the limit surface models with the normalized power-law pressure distribution $\pressureunit$ for each contact surface.
For each grasp, we scaled the contact wrench constraints for each contact with $\lambda_p$, instead of with the sum magnitude $F$ of the normal forces in \eref{eq:contact_constraints} to match the force sensor readings.

\input{figs/figPhyGWS}

\subsection{Baseline Contact Models}
We consider the following baseline contact models to predict the grasp success
\begin{itemize}
    \item \emph{3DLS-planar}: the traditional planar surface contact models. As shown in \fref{fig:setup} (right), a planar contact surface (orange line) is created by projecting the nonplanar surface (blue line) along the $x$-axis onto the $yz$-plane. We computed the frictional wrenches in the (${f}_y, {f}_z, {\tau}_x$)-space and fit the 3DLS models to the wrenches. 
    \item \emph{6DFW}: we computed 6D frictional wrenches (6DFWs) for a nonplanar surface contact and used the 6DFWs of each jaw to construct the GWS without a LS model.
    \item \emph{3DLS-nonplanar}: the 3DLS models are fit to the three major components, ${f}_y, {f}_z, {\tau}_x$, of the 6DFWs, while the remaining three components are set to zero \cite{xu.2017}.
    \item \emph{6DLS}: the proposed 6DLS models are fit to all components of the 6DFWs.
\end{itemize}
We have in a total of seven baseline contact models, as each LS model has the quartic and ellipsoid variants.

\subsection{Grasp Success Prediction Results}
\label{subsec:results}
We selected 115 well-distributed wrench disturbances for the ten grasped objects. 
For each disturbance, three grasp forces in the range of 20N--35N are randomly chosen.
By repeating each grasp three times, we collected in a total of 1,035 physical grasps. 
Each grasp is followed by a slow vertical lifting so that the acceleration affects the disturbances minimally. 
For the scenarios with a medium to high moving speed of the robot arm, one can model the acceleration of the grasped object as an additional external disturbance as the current algorithm neglects the inertial terms.
We determined the friction coefficient $\mu=0.3$ experimentally and used the power-law pressure distribution with $k=2.4$ for each contact. 
We ran the predictions with each model five times and use precision and recall to evaluate the results. 
Note that precision is inversely related to the number of false positive predictions, whereas high recall indicates low false negatives. 
We also computed the $F_1$ score and the accuracy of each model. 
\tref{tab:expRes} shows the means and the standard deviations of the seven baseline models for grasps with both object types, as we observed similar results for object type I and II.

We observed that the proposed 6DLS models outperform the 3DLS-planar, the 6DFW, and the 3DLS-nonplanar by up to 26\%, 12\%, and 9\% in recall, respectively, while maintaining a comparable precision.
High recall indicates that the 6DLS models reduce false negatives, and therefore, avoids unnecessary grasp force and grasp pose adaptations in robot manipulation.
In addition, the 6DLS models also increase $F_1$ score and accuracy by up to 16\% and 7\% compared to the remaining models.
Furthermore, the standard deviation of 6DFW is higher as the frictional wrenches are randomly sampled. 
Hence, in addition to a higher accuracy, a limit surface model also increases repeatability in predictions. 

We further observed that the ellipsoidal LS models slightly outperform the quartic ones even though the quartic achieves a lower wrench fitting error, as shown in \sref{sec:Fitting_results_parametric_surface}.
\fref{fig:GWS_compare}(a) shows the convex hulls in the first quadrant of a 2D projection of the GWSs constructed with an ellipsoidal and a quartic 6DLS linearized with 728 points. 
We observe that the difference between the two GWSs is small, as the LS models are not densely sampled for linearization due to the high computational complexity of the Minkowski sum operation.
Fewer samples on the LS model lead to an overly conservative LS approximation due to the convex hull operation. 
In future work, we plan to pose the wrench resistance as a convex optimization problem without building a GWS \cite{kerr1986analysis,Han2000Grasp,mahler2019learning} and further compare the two 6DLS models with more samples.

We compared the GWSs constructed with the 6DFW and the three ellipsoidal LS models.
\fref{fig:GWS_compare}(b) shows the 2D projections of the GWSs.
The GWS constructed with the proposed 6DLS is larger than with the 3DLS models, as it considers the full 6D frictional wrenches, and therefore, reduces false negatives. 
Note that the $\tau_y^\text{GWS}$ and $\tau_z^\text{GWS}$ components of the GWS constructed with the 3DLS-planar and the 3DLS-nonplanar models are not zero, because although the 3DLS models did not consider the frictional torques $\tau_y$ and $\tau_z$ with respect to the pressure center when computing the LS, the frictional forces introduced an additional torque when constructing the GWS due to the change of frame using \eref{eq:contact_constraints}.  
Although the 6DFW also considers 6D frictional wrenches, the constructed GWS is not symmetric, as the initial frictional wrenches are randomly sampled, resulting in worse and less repeatable prediction results.
The largest difference between the GWSs constructed with the 6DLS and the 3DLS is the $f_x^\text{GWS}$ component, as shown in the right of \fref{fig:GWS_compare}(b). 
The $f_x^\text{GWS}$ component of the GWSs constructed with both 3DLS models is zero as they do not consider the frictional force component $f_x$ and the normal forces of the left and the right jaws cancel out.
However, force disturbances along the $x$-axis are not evaluated in the experiments as the $x$-axis is the grasp axis. 
In future work, we intend to evaluate the 6DLS with full 6D wrench disturbances for the scenarios that the object is not immobilized along the $x$-axis.

\input{figs/figPhySensitivity}

\subsection{Sensitivity Analysis}
\label{subsec:sensitivity}
As each of the contact models contains several parameters, such as friction coefficient, pressure distribution, we include an analysis of each model’s sensitivity to a subset of these parameters.
\fref{fig:sensitivity} shows the precision and recall of the 6DFW and the ellipsoidal LS models with shaded error bars showing the standard deviations of five runs of each model.

\subsubsection{Effect of Threshold Pairs}
As described in \sref{subsec:exp_setup}, the grasp is labeled as a success if the object rotation and translation are below a threshold pair.
We compared prediction results with five threshold pairs in increasing order, where the second pair is used to evaluate the models in \sref{subsec:results}. 
With an increasing threshold, the precision of each model increases, as more grasps are labeled as success and the number of true positive predictions increases, whereas recall decreases because the models predict more false negatives.
\fref{fig:sensitivity}(a) further shows that the proposed ellipsoidal 6DLS has the highest recall with similar precision for all thresholds.
A large threshold pair is suitable to robot applications such as bin-picking, as the exact object pose is not critical to the manipulation success.
With precision (88\%) higher than recall (71\%), the 6DLS model becomes conservative for such applications, but still increases recall by up to 24\% over the remaining three baseline models.

\subsubsection{Effect of Pressure Distributions}
As the exponent $k$ of the power-law pressure distribution \cite{xydas.1999} changes with contact surfaces and grasp forces, we also analyzed the effect of $k$.
Note that the pressure distribution is close to uniformity with $k=10^6$.
\fref{fig:sensitivity}(b) shows that the recall of each contact model increases with $k$ as the frictional torque of each contact increases and the models predict more positives.

\subsubsection{Effect of Friction Coefficients}
\fref{fig:sensitivity}(c) illustrates the prediction results with different friction coefficient offsets. 
The symbol $\pm 0\%$ indicates that the models used the experimentally determined value $\mu=0.3$ and $+10\%$ offset represents $\mu=0.33$. 
We note that the predictions of all baseline models are relatively sensitive to $\mu$ as a LS linearly scales with $\mu$.
For scenarios with an unknown friction coefficient, one can select a lower $\mu$ value for conservative predictions as each model predicts fewer positives.

\subsubsection{Effect of Contact Lengths}
\fref{fig:sensitivity}(d) illustrates the results with different contact length offsets. 
The symbol $\pm0\%$ indicates that the models used the measured contact length $l_1$ and $l_2$ for the two contact types illustrated in \fref{fig:setup} (right).
The recall of each baseline model increases with the contact length as the surface area and the frictional torque of each contact also increases.

%% file: figs/figPhySetup.tex
\begin{figure}
	\centering	
	{{\includegraphics[width=\linewidth]{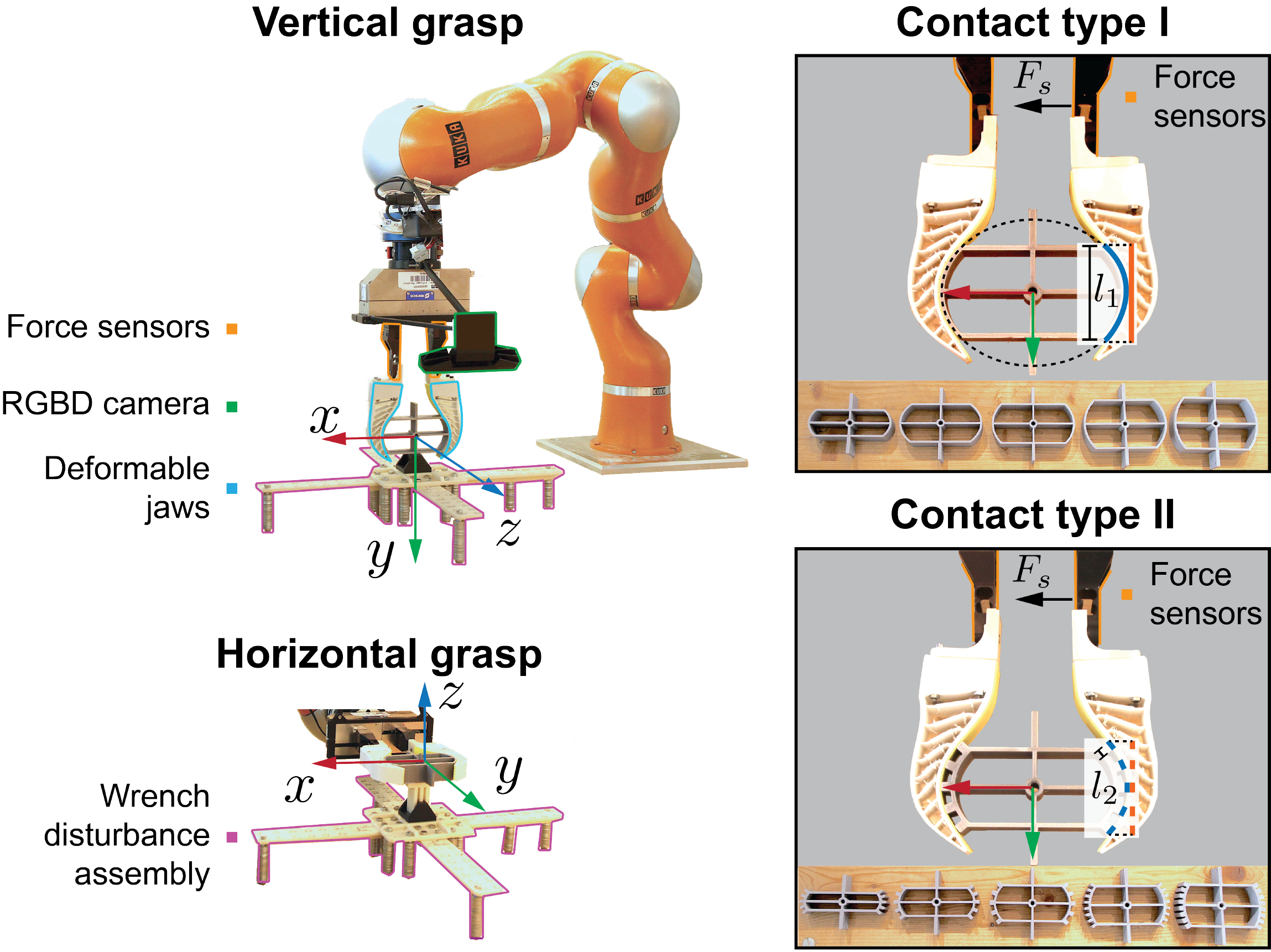} }}%
	\caption{\small Experiment setup for grasp success prediction. Left: deformable gripper jaws (blue) grasp a 3D-printed object with nonplanar surfaces. The 3D printed assembly (pink) attached to the grasped object generates external disturbances.  Right: Ten 3D-printed rigid objects that create two types of contact surfaces. }
	\label{fig:setup}%
\end{figure}

%% file: figs/figPhysFEM.tex
\begin{figure}
	\centering
    \subfloat[]{{\includegraphics[height=28.9mm]{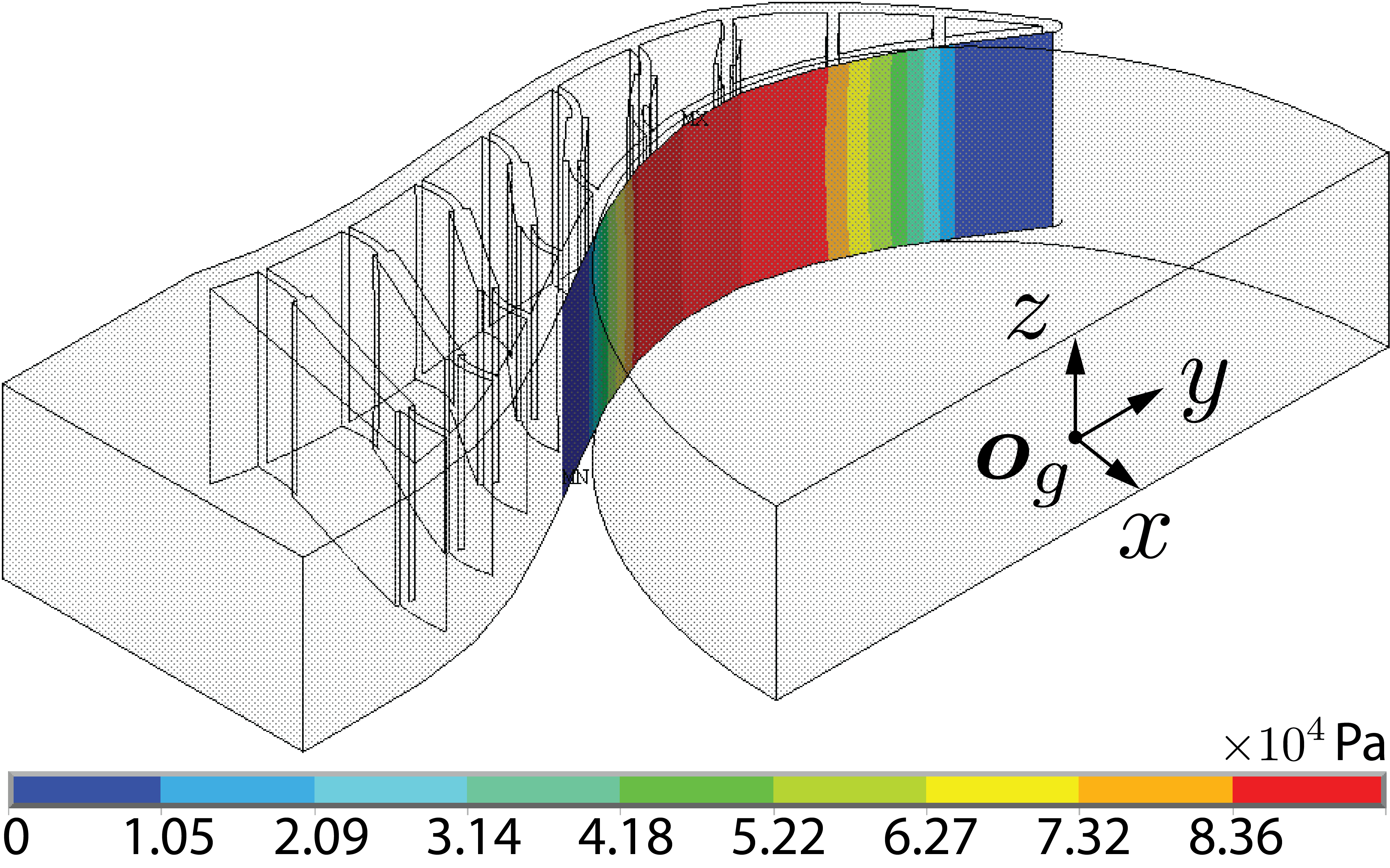} }}%
    \hspace{1em}
    \subfloat[]{{\includegraphics[height=28.9mm]{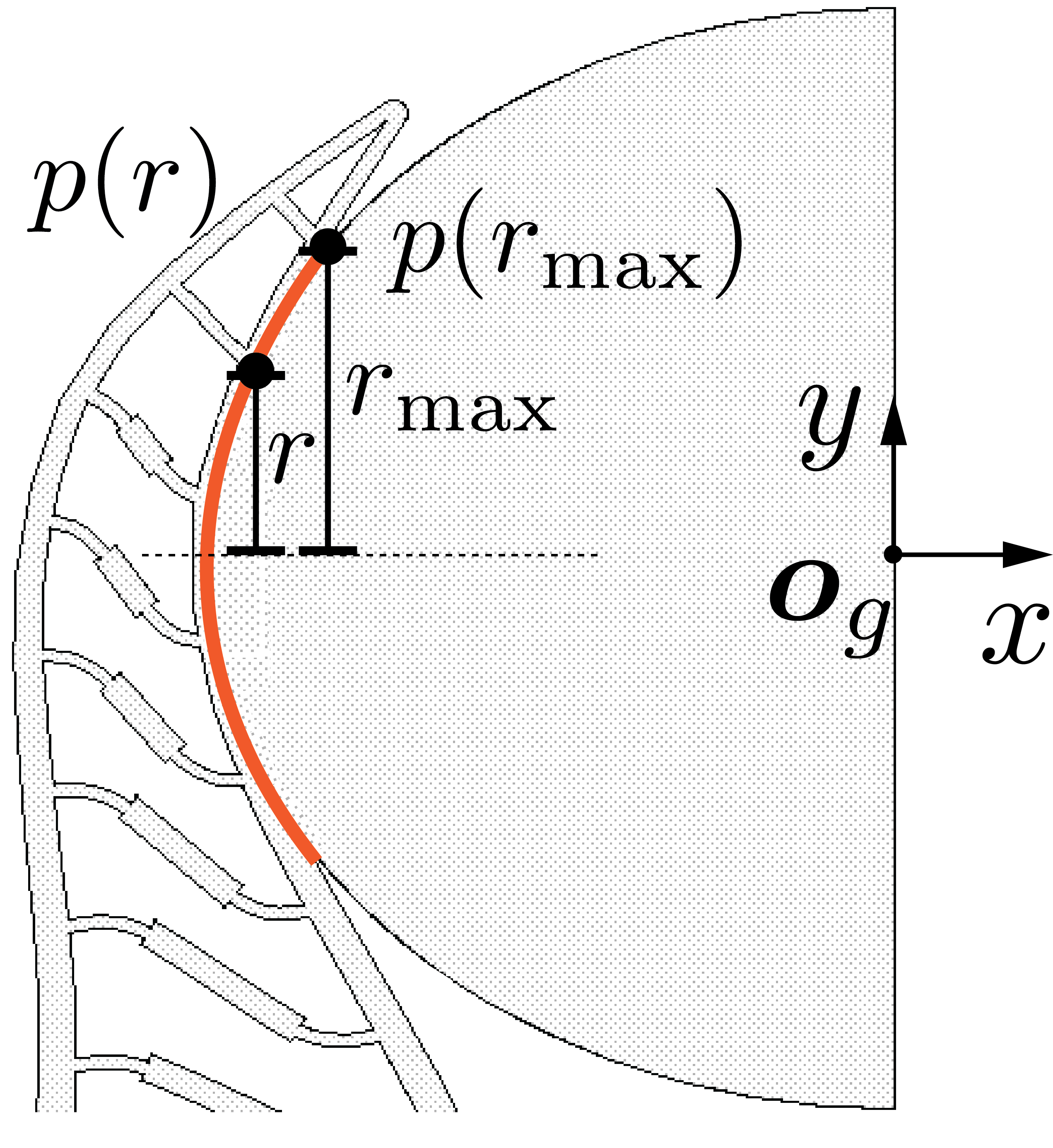} }}%
    
    \subfloat[]{{\includegraphics[height=28.9mm]{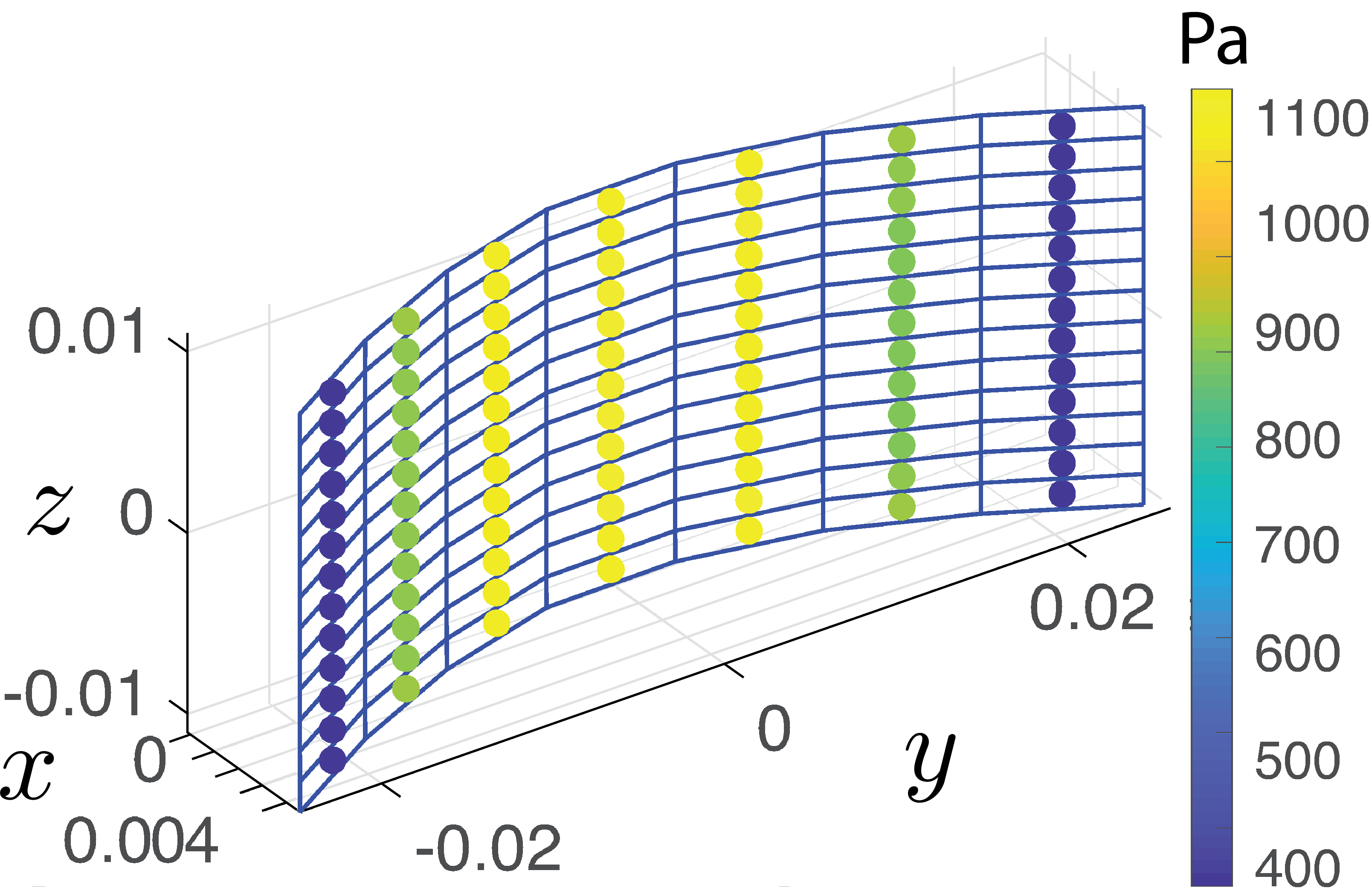} }}%
    \hspace{1em}
    \subfloat[]{{\includegraphics[height=28.9mm]{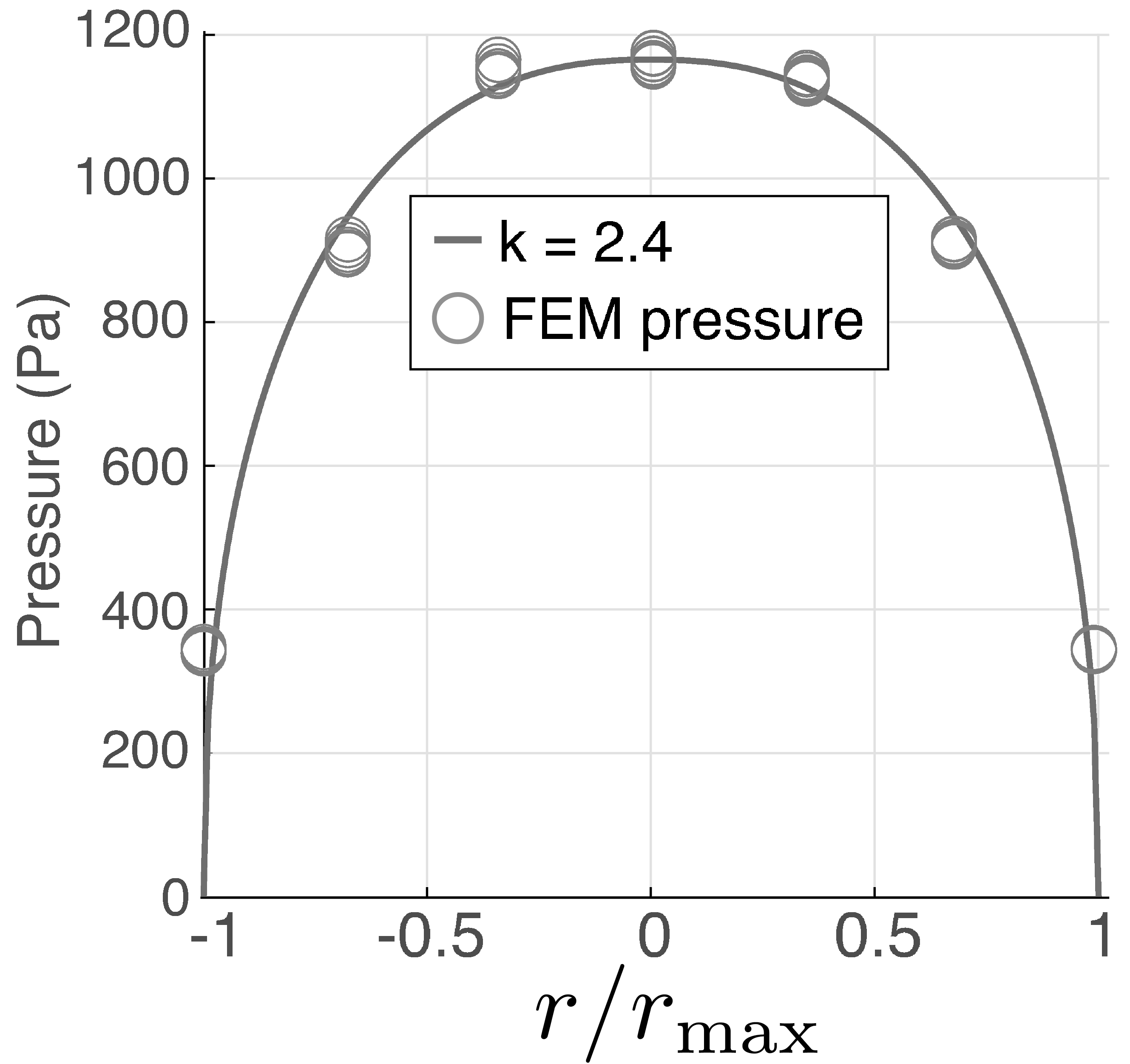} }}%
	\caption{\small {FEM simulation and the power-law pressure distribution of a nonplanar surface contact between a deformable fin-ray jaw and a rigid elliptic cylinder. 
	}}
	\label{fig:FEMFesto}
\end{figure}

%% file: tabs/tablePhysResults.tex
			
	   

			

\begin{table*}
	\begin{center}
		\caption{Prediction results of 1,035 physical grasps for the ten objects of type I and II.}
		\begin{tabular}{|c|c|c|c|c|c|}
			\hline
			Contact model & Limit surface model & Precision (\%) & Recall (\%) & $F_1$ score (\%) & Accuracy (\%) \\ \hline
			\multirow{2}{*}{3DLS-planar}  & Quartic  & 68.2$\pm$0.1 & 46.8$\pm$0.1 & 55.5$\pm$0.1 & 59.9$\pm$0.0  \\ \cline{2-6}
			& Ellipsoid  & \textbf{71.0$\pm$0.2} & 50.8$\pm$0.2 & 59.2$\pm$0.2 & 62.6$\pm$0.2  \\ \hline
			
			6DFW  & None  & 63.2$\pm$0.5 & 64.5$\pm$1.2 & 63.9$\pm$0.8 & 61.0$\pm$0.7  \\ \hline
	   
			\multirow{2}{*}{\makecell{3DLS-nonplanar \cite{xu.2017}}} & Quartic  &  66.0$\pm$0.1& 64.4$\pm$0.2 & 65.2$\pm$0.1 & 63.3$\pm$0.1  \\ \cline{2-6}
			& Ellipsoid  & 67.6$\pm$0.3 & 68.4$\pm$0.3  & 68.0$\pm$0.2 & 65.6$\pm$0.2 \\ \hline

			\multirow{2}{*}{\makecell{Proposed  6DLS} }  & Quartic & 65.7$\pm$0.1 & 73.5$\pm$0.4  & 69.4$\pm$0.2 & 65.3$\pm$0.1 \\ \cline{2-6}
			& Ellipsoid & 66.8$\pm$0.0 & \textbf{76.9$\pm$0.2} & \textbf{71.5$\pm$0.1} & \textbf{67.3$\pm$0.1}  \\ \hline
			
		\end{tabular}
	\label{tab:expRes}
	\end{center}
\end{table*}


			
	   

			

%% file: figs/figPhyGWS.tex
\begin{figure*}
	\centering
    \subfloat[]{{\includegraphics[height=12.4em]{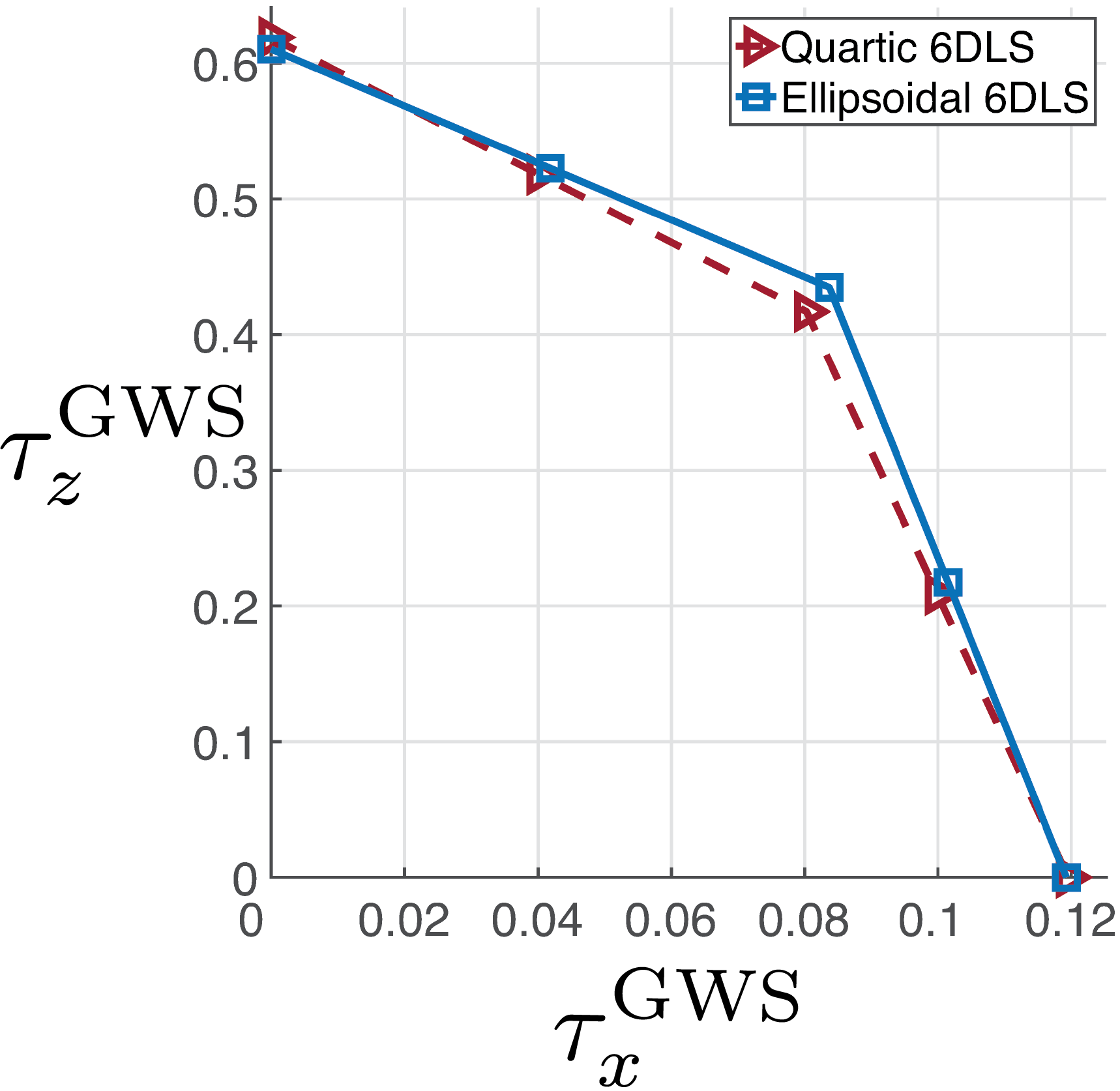} }}
    \subfloat[]{{\includegraphics[height=12.4em]{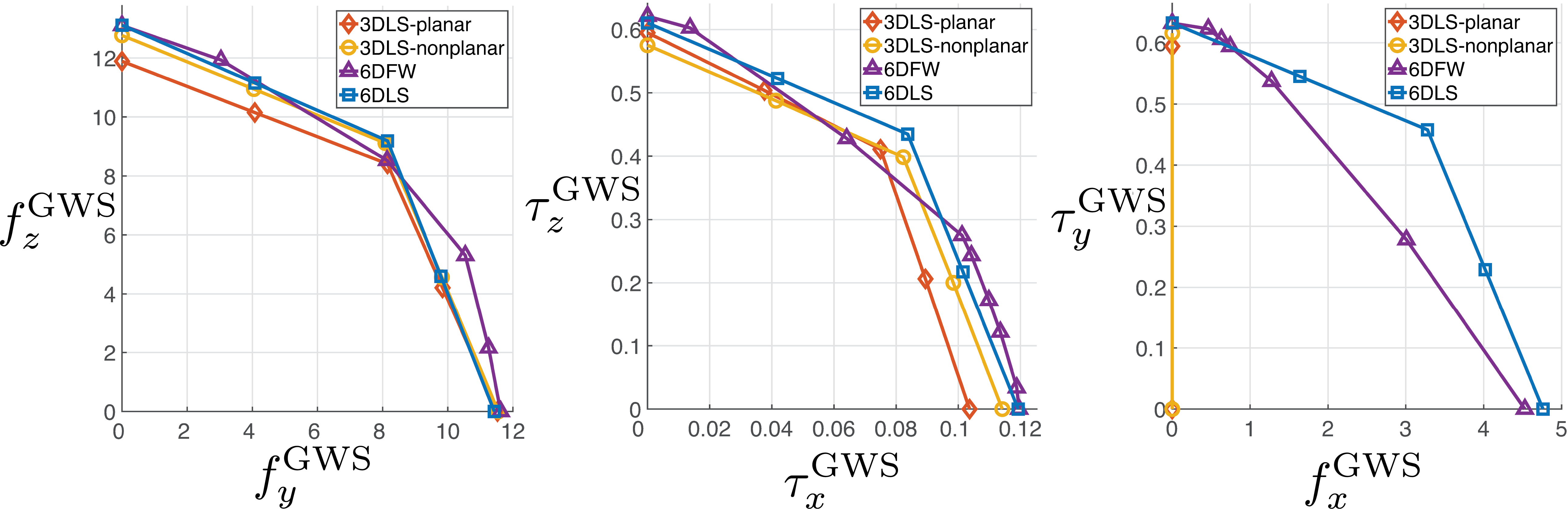} }}
	\caption{{\small 2D projections of a representative 6D grasp wrench space constructed (a) with a quartic and an ellipsoidal 6DLS model, (b) with the 6DFW and the ellipsoidal limit surface models.}}
	\label{fig:GWS_compare}
\end{figure*}

%% file: figs/figPhySensitivity.tex
\begin{figure*}
	\centering
	\subfloat[]{\includegraphics[width=0.25\linewidth]{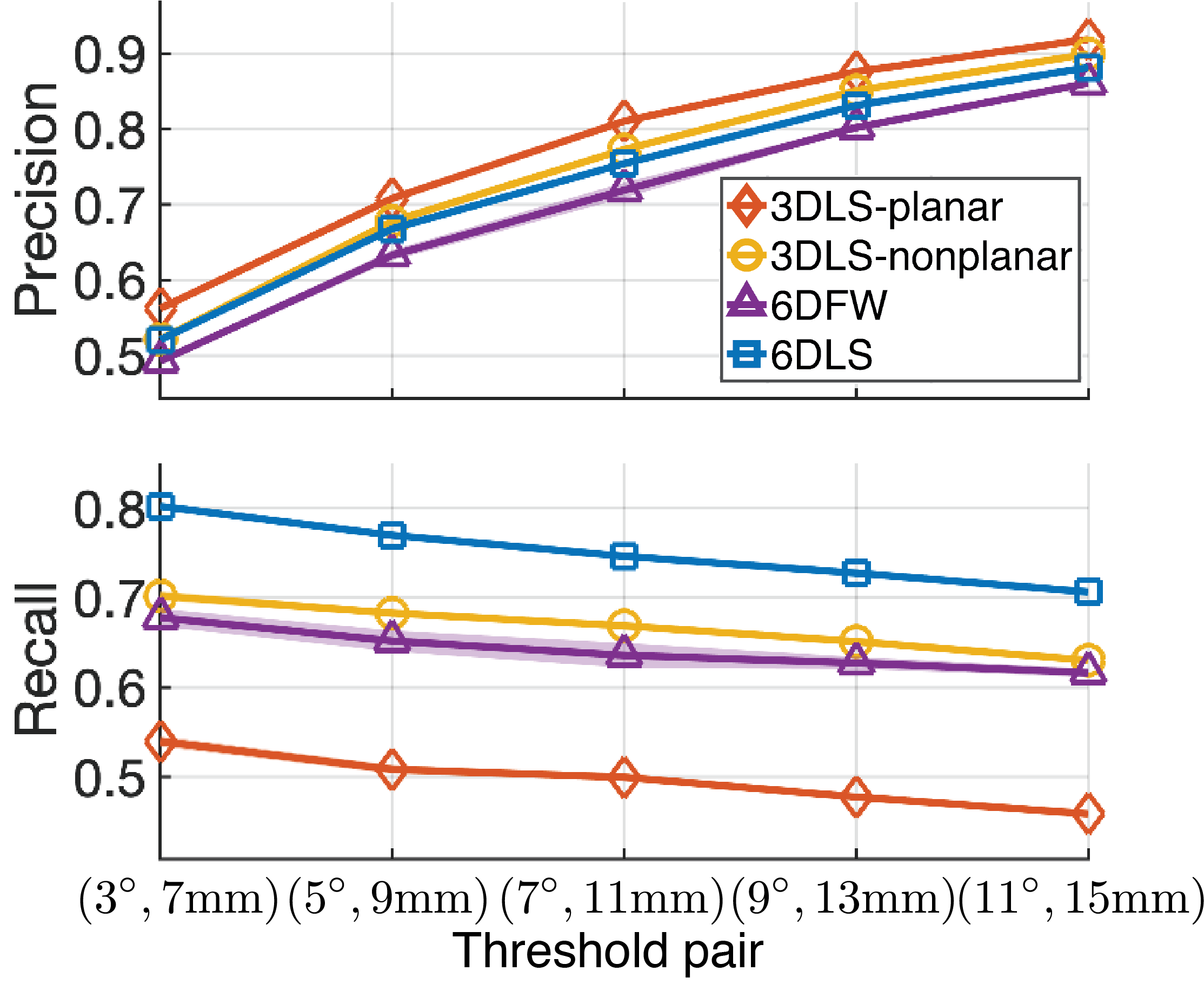}}
	\subfloat[]{\includegraphics[width=0.25\linewidth]{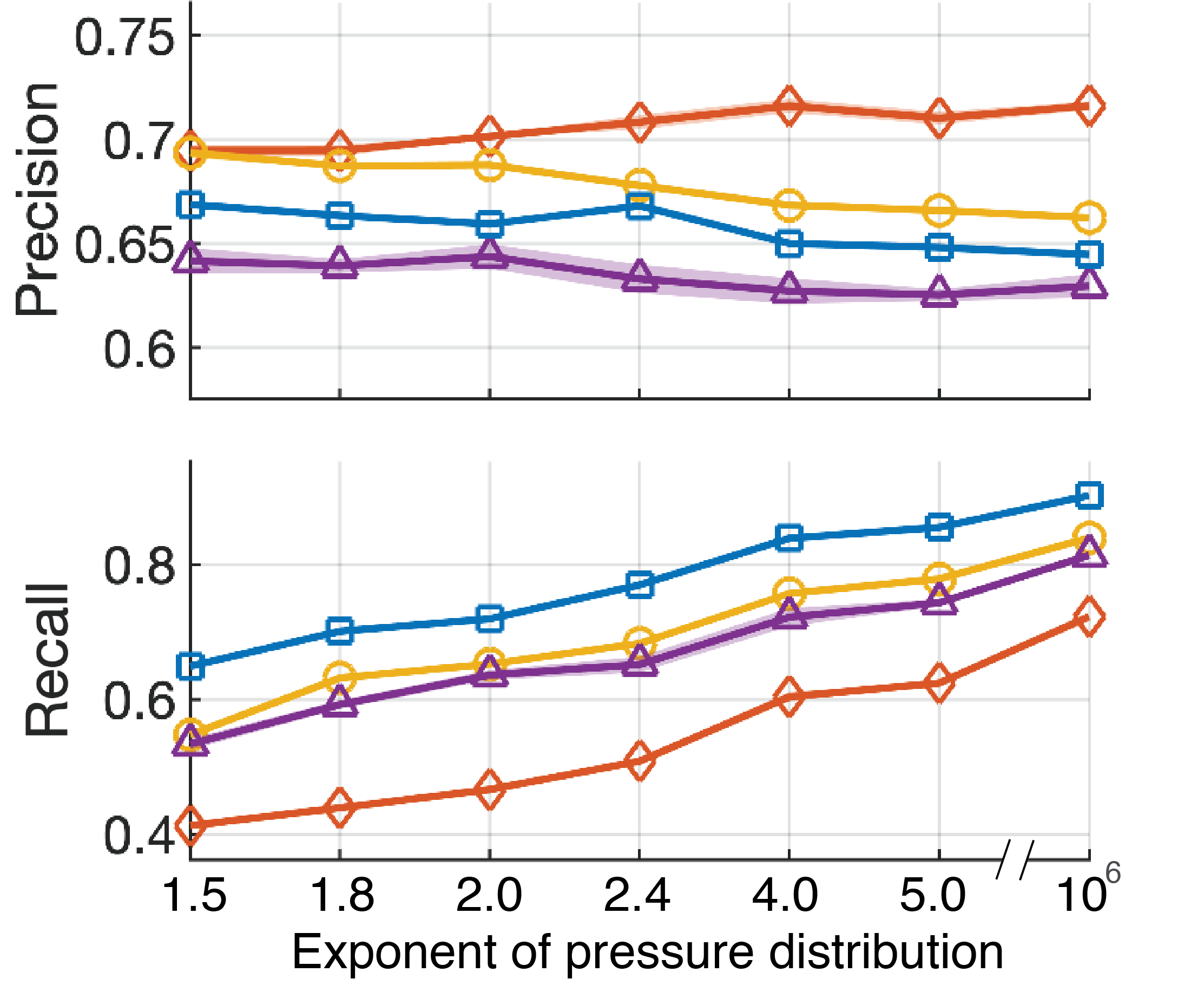} }
	\subfloat[]{\includegraphics[width=0.25\linewidth]{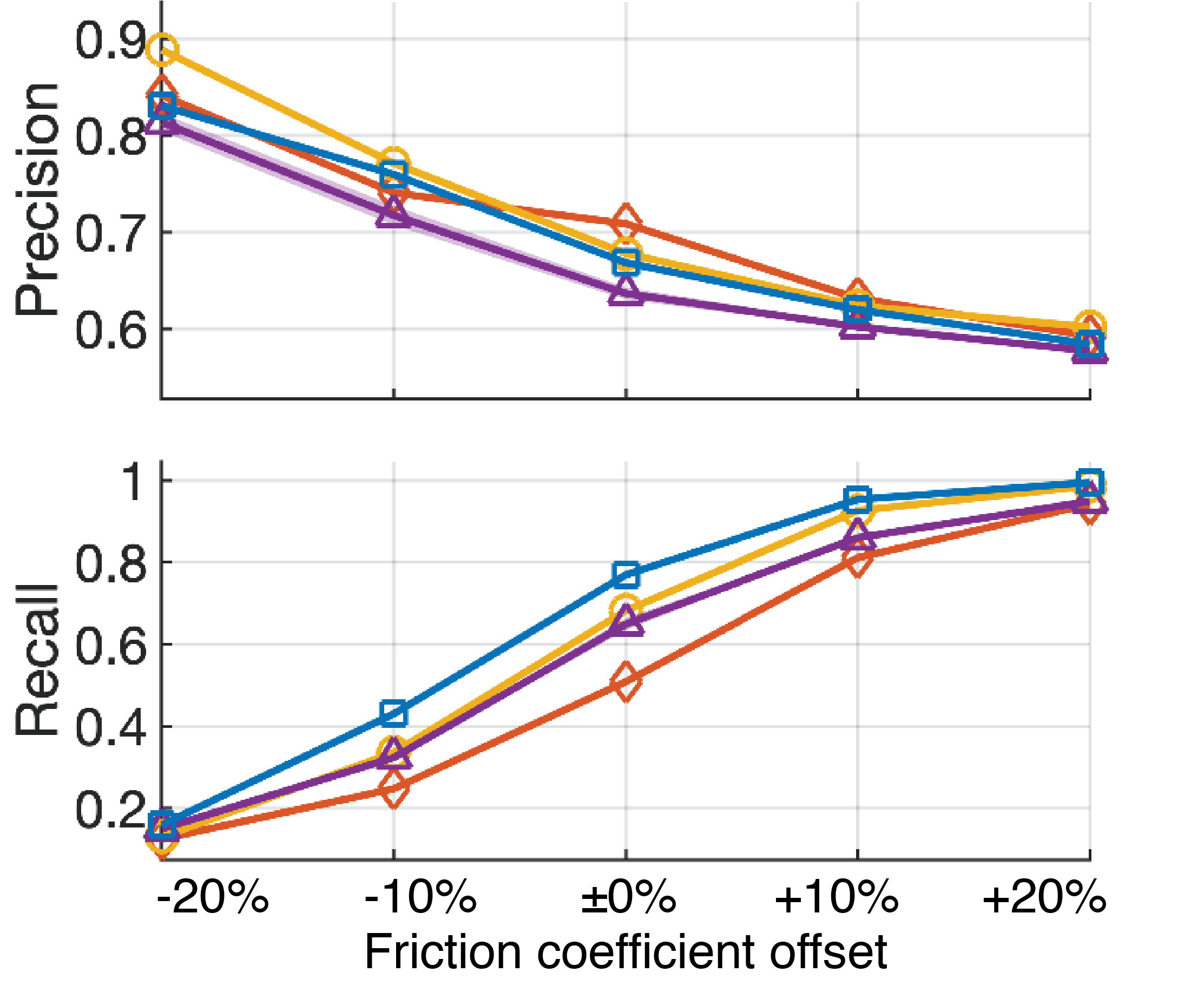} }
	\subfloat[]{\includegraphics[width=0.25\linewidth]{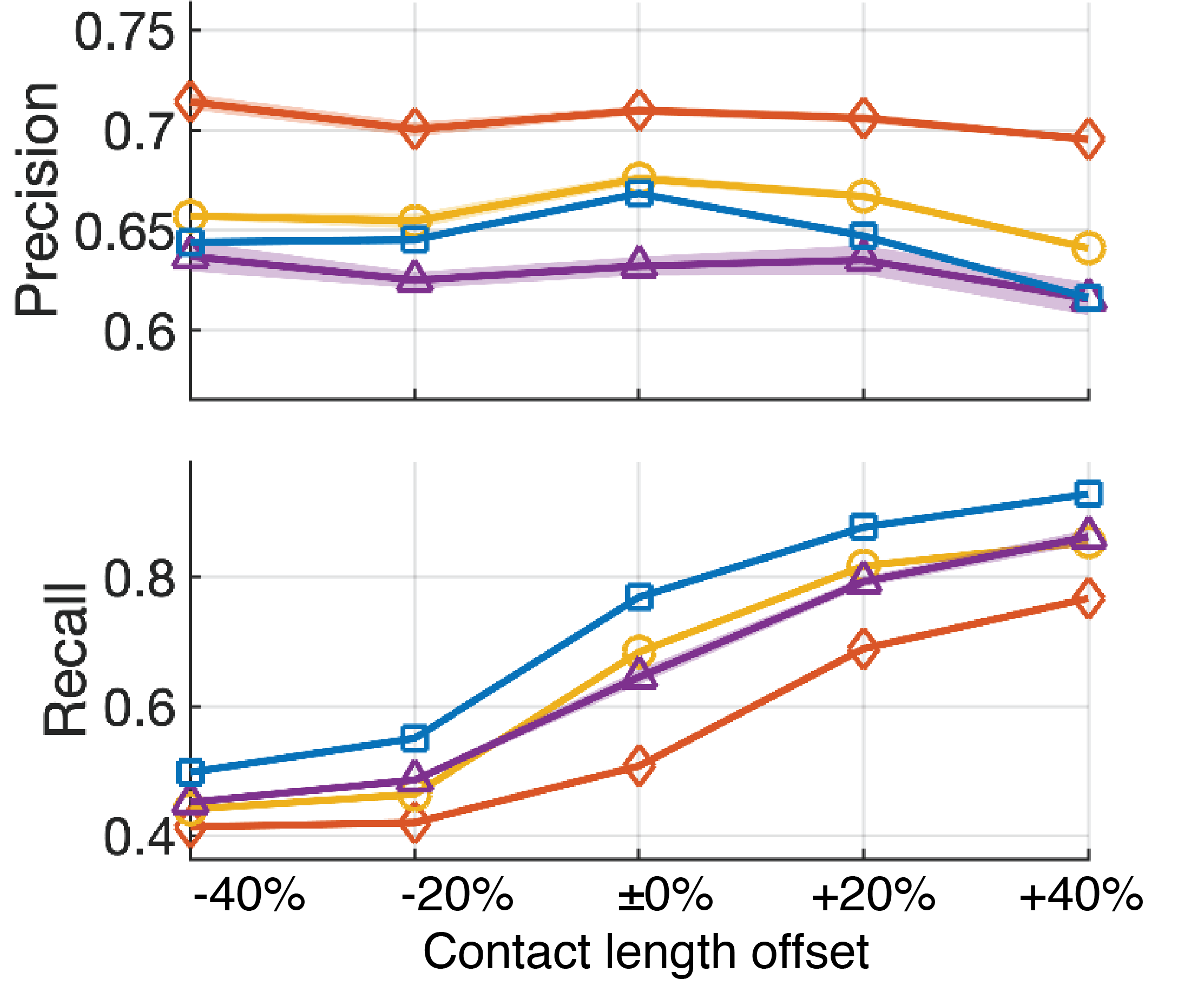} }
	\caption{{\small  Precision and recall for each contact model as a function of (a) the threshold pairs in increasing order, (b) the exponent $k$ of the power-law pressure model \cite{xydas.1999} with $k=10^6$ being close to a uniform distribution, (c) friction coefficient offset with $\pm0\%$ meaning the experimentally determined value $\mu = 0.3$, (d) contact length offset with $\pm0\%$ meaning the measured contact length $l_1$ or $l_2$ for the two contact types, respectively. The contact area increases with the contact length. }}
	\label{fig:sensitivity}
\end{figure*}

%% file: sections/13-discussion.tex
\label{sec:discussion}
The power-law pressure distribution described in \sref{sec:profile} is based on the assumption that the pressure is symmetric about the object center.
However, we observed in our FEM simulations and the results shown in \cite{shan2020modeling} that the pressure distribution can be asymmetric depending on the object pose relative to the fin-ray jaw.
With the vertical grasp direction shown in \fref{fig:setup} (left), an asymmetric pressure distribution leads to a component of the normal force that is parallel to the gravity direction, and therefore, affects the prediction results. 
Furthermore, contact profiles can change during the manipulation due to the jaws' deformation.
One way to address the two limitations is to relax the assumption of a constant symmetric contact profile and to constantly predict grasp success with updated profiles captured with deformable tactile sensors such as GelSlim~\cite{ma.2019} or the tactile fingertip sensors by Romero~et~al.~\cite{romero2020soft}.
A tactile sensor can further better detect relative motions between the grasped object and the jaws compared to tracking object poses using point clouds.

The contact surfaces evaluated in the physical experiments are restricted in (discrete) elliptic cylinders due to the design of the fin-ray jaws.
Using a tactile sensor that provides a 3D force field will likely improve the prediction accuracy for various contact surface geometries. 
Additionally, different pressure distributions, such as the radially-distributed model~\cite{arimoto2000dynamics} and a parallel-distributed model~\cite{inoue2008mechanics}, can be compared with the force field to better model the pressure of nonplanar surfaces. 

Prediction with updated contact profiles requires a real-time implementation of the algorithm. 
We note that finding a 6DLS model requires a minimization with many variables and is not real-time capable.
One way to enhance the computational speed is to apply common deep-learning techniques, so the fitting process can be completed at millisecond level. 
Specifically, a neural network can be trained to output a 6DLS model given a contact profile, as the current algorithm provides the ground truth for training.

%% file: sections/14-conclusion.tex
We propose the concept of a 6D limit surface to represent the 6D frictional wrench limit for a nonplanar surface contact.
We further generalize the quartic and ellipsoidal LS models from 3D to 6D to approximate a 6DLS.
Fitting results with parametric surfaces and FEM simulations show that the quartic and ellipsoidal 6DLS models have as low as 0.02 and 0.04 mean wrench error, respectively, which suggests that both models well describe frictional wrenches for a large variety of contacts.  

We further introduce an algorithm that builds a grasp wrench space with the 6DLS model for each jaw to predict multicontact grasp success. 
Physical experiments show that the proposed algorithm increases recall by up to 26\% over the existing contact models with similar precision, as well as improves $F_1$ score and accuracy by up to 16\% and 7\%, respectively. 
This suggests that the proposed algorithm is helpful in friction analysis for nonplanar surface contacts and in grasp success prediction with deformable jaws.    

In addition to addressing the limitations mentioned in Section~\ref{sec:discussion}, we intend to design a novel 6DLS-based grasp quality metric to plan grasps with deformable jaws in future work.

%% file: sections/15-acknowledgement.tex
This work has been funded, in part, by the Initiative Geriatronics by StMWi Bayern (Project X, grant no. 5140951).

We thank Prof. Ken Goldberg, Prof. Alberto Rodriguez, and Prof. Hao Su for the discussions and the constructive feedback. 
We also thank our colleagues who helped with the experiments and provided helpful suggestions, in particular Stefan Lochbrunner, Michael Danielczuk, Jeffrey Ichnowski, Matti Strese, Mojtaba Karimi, Dmytro Bobkov, Chongze Yu, Yeting Dong, and Kuo-Yi Chao.

%% file: sections/16-biography.tex

\begin{IEEEbiography}[{\includegraphics[width=1in,height=1.25in,clip,keepaspectratio]{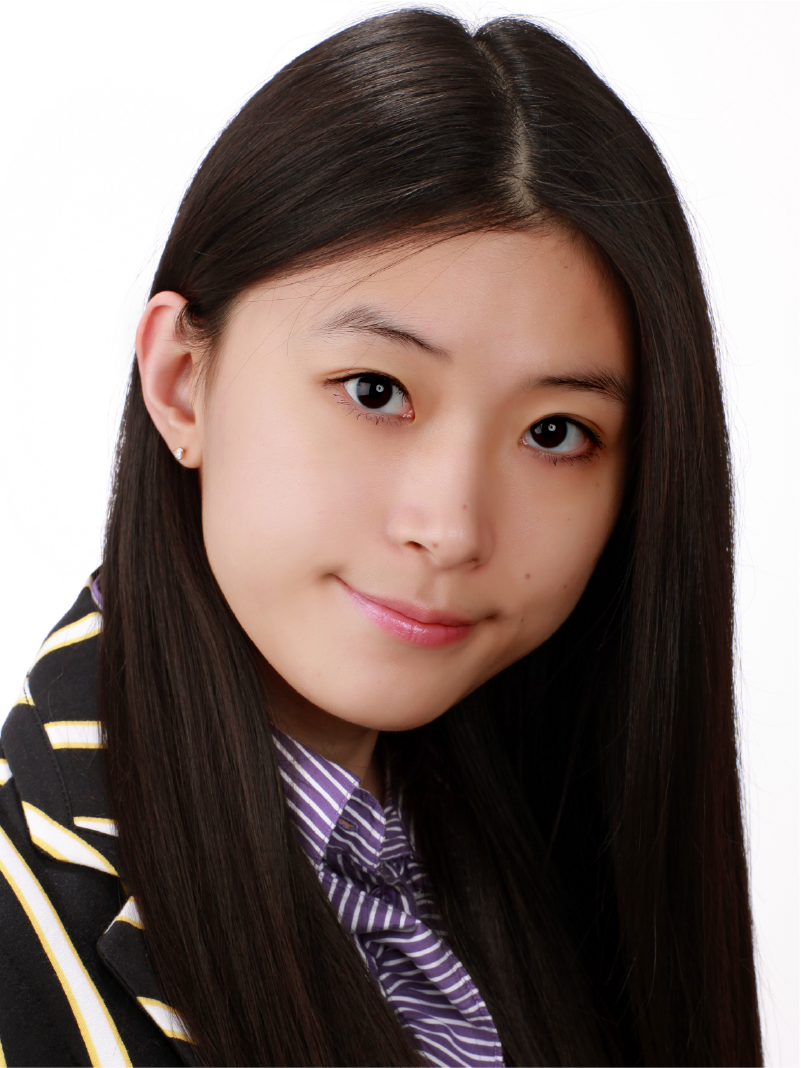}}]{Jingyi Xu}
studied Electrical Engineering and Information Technology at the Technical University of Munich (Germany). She received her B.S. and M.S. degrees (passed with high distinction) in 2012 and 2014, respectively. In November 2014, she joined the Chair of Media Technology at the Technical University of Munich as a member of the research associate.
From April to October 2019, she was a visiting PhD student with the AUTOLab at University of California, Berkeley. 
Her research focus is model-based grasp planning with deformable jaws.
\end{IEEEbiography}

\begin{IEEEbiography}[{\includegraphics[width=1in,height=1.25in,clip,keepaspectratio]{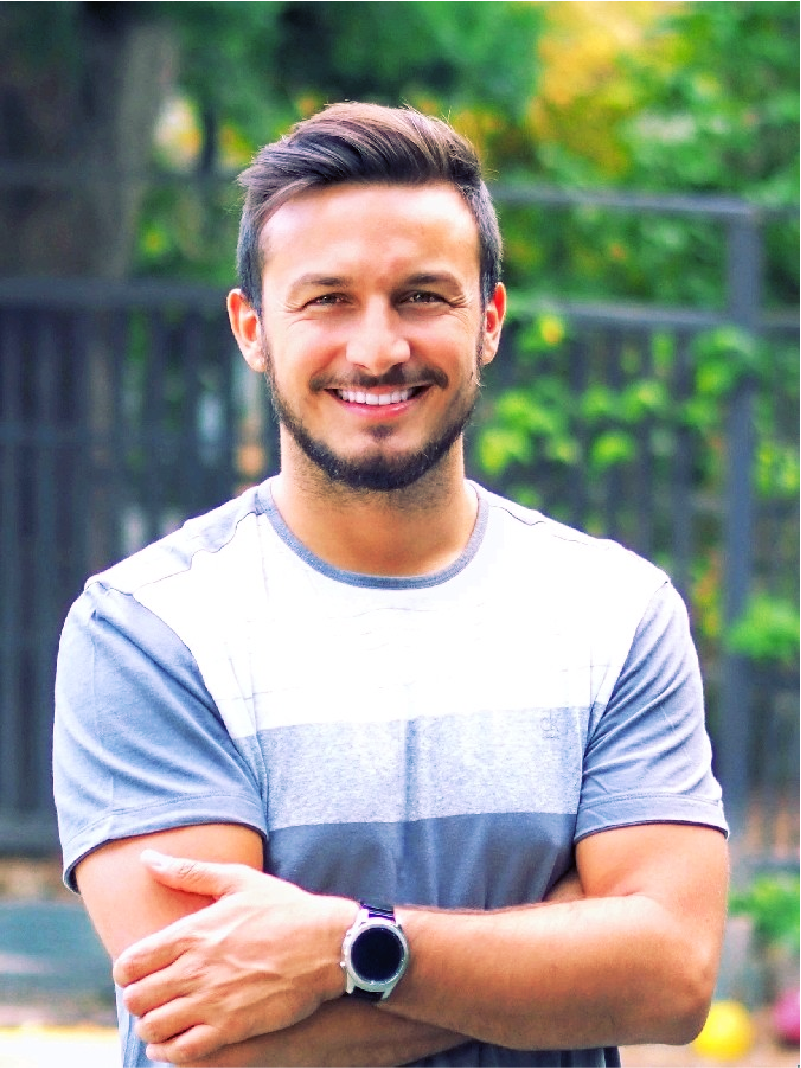}}]{Tamay Aykut} studied Electrical Engineering and Information Technology at the Technical University of Munich (Germany). He received his M.S. in 2016. In March 2016, he joined the Chair of Media Technology at the Technical University of Munich as a research associate and  received his Engineering Doctorate in August 2019. In September 2019, he joined Stanford University as Visiting Assistant Professor while leading the Visual Computing and Artificial Intelligence (VCAI) group at the Max-Planck Center for Visual Computing and Communication (MPC-VCC). His research interests comprise explainable artificial intelligence, visual computing and communication, as well as the extensive field of mobile robotics.
\end{IEEEbiography}

\begin{IEEEbiography}[{\includegraphics[width=1in,height=1.25in,clip,keepaspectratio]{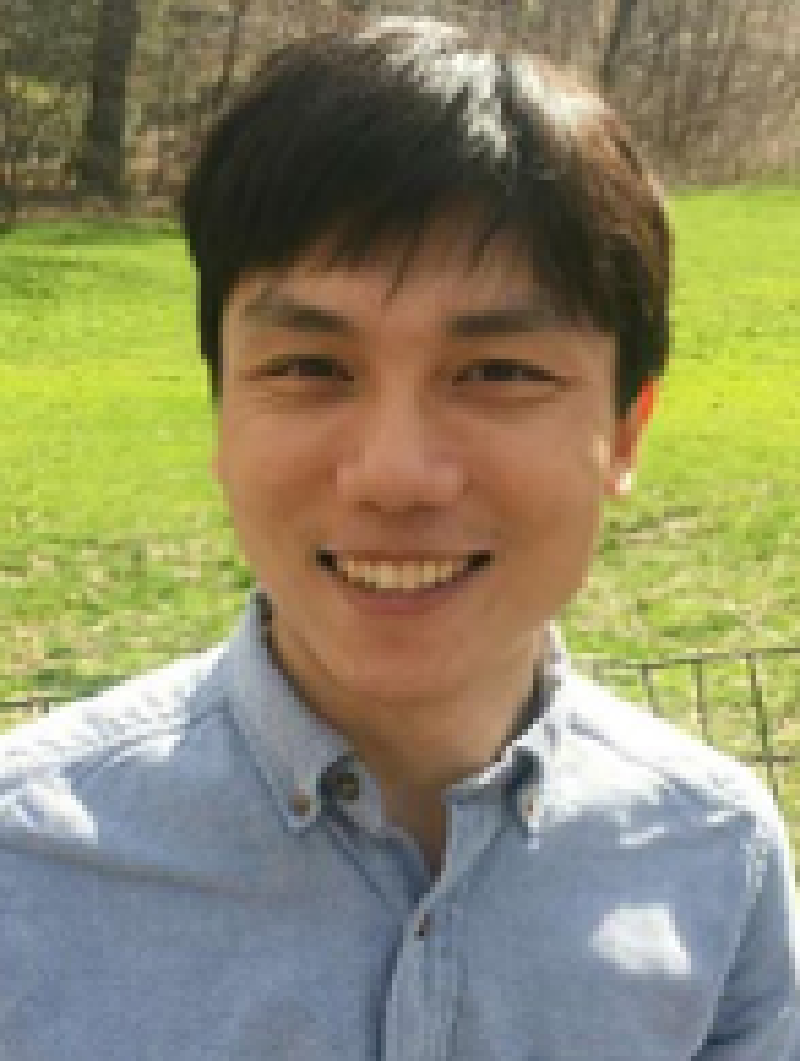}}]{Daolin Ma}
studied Theoretical and Applied Mechanics at Peking University (China). 
He received a Bachelor of Science (B.Sc.) degree in 2009 and a Ph.D. degree in 2015. In December 2016, he joined the Manipulation and Mechanism Lab at Massachusetts Institute of Technology (USA) as a post-doc. 
His research focus is contact modeling, tactile sensing and robotic manipulation.
\end{IEEEbiography}

\begin{IEEEbiography}[{\includegraphics[width=1in,height=1.25in,clip,keepaspectratio]{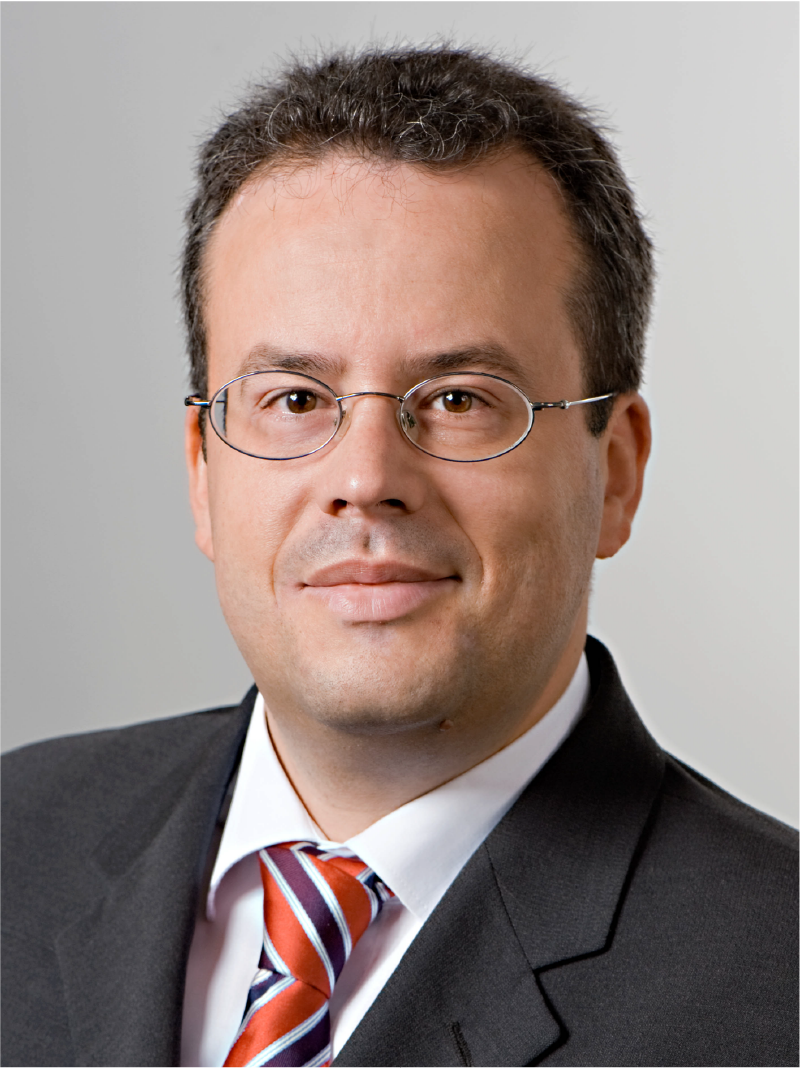}}]{Eckehard Steinbach}
studied Electrical Engineering at the University of Karlsruhe (Germany), the University of Essex (Great Britain), and ESIEE in Paris. From 1994 -- 2000 he was a member of the research staff of the Image Communication Group at the University of Erlangen-Nuremberg (Germany), where he received an Engineering Doctorate in 1999. From February 2000 to December 2001, he was a Postdoctoral Fellow with the Information Systems Laboratory of Stanford University. In February 2002, he joined the Department of Electrical Engineering and Information Technology of Technical University of Munich (Germany), where he is currently a Full Professor for Media Technology. His current research interests are in the area of audio-visual-haptic information processing and communication as well as networked and interactive multimedia systems.
\end{IEEEbiography}

%% file: sections/17-background.tex
We revisit the friction computation for planar contacts based on the concepts presented in \cite{goyal.1991,lee1991fixture,howe.1996}\cite[pp.~130--134]{mason2001mechanics}.
Friction depends on the relative motion between two bodies in contact. 
In two dimensions, the instantaneous motion of a body can be described as a rotation around a point defined as the \emph{center of rotation (COR)}.
A translation is considered as a rotation around a COR that is infinitely far away. 
The idea is to compute the frictional force of an infinitesimally small element by assuming a known COR. 
The friction of the contact area is computed by summing up the contribution of each element.
Possible frictional force and torque pairs can be obtained by sampling different CORs.

\fref{fig:planar-friction} shows a planar contact area $\Area$.
A rectilinear coordinate system is assumed to be fixed in the 2D plane, where its origin is located at the pressure center $\origin = [\originx,\originy]^T$ 
\begin{equation*}
	\originx = \frac{\int_{\Area} x \bigcdot  \pressure(x,y)\dA}{\int_{\Area}  \pressure(x,y)\dA}, \hspace{0.5em}
	\originy = \frac{\int_{\Area} y \bigcdot  \pressure(x,y)\dA}{\int_{\Area}  \pressure(x,y)\dA}
\end{equation*}
where $\pressure(x,y)$ is the pressure at $(x,y)$.

A representative COR is illustrated in \fref{fig:planar-friction}.
The velocity $\boldsymbol{v}(x,y)$ at the infinitesimally small contact area $\dA$ is perpendicular to the vector $\boldsymbol{d}(x,y)$, which is the vector from the COR to $\dA$.
The local frictional force $\diff\boldsymbol{f}(x,y)$ is opposite to $\boldsymbol{v}(x,y)$ and $\diff\boldsymbol{f}(x,y) = -\mu \bigcdot \pressure(x,y) \bigcdot \boldsymbol{v}(x,y)/\norm{\boldsymbol{v}(x,y)} \dA$.

By integrating the local frictional force and torque over $\mathcal{A}$, we obtain 
\begin{equation*}
	\boldsymbol{f} = [f_{x},f_{y}]^T = \int_\mathcal{A} \diff\boldsymbol{f}(x,y), \hspace{0.5em}
	\tau_z = \int_\mathcal{A}\vec{r}(x,y) \times \diff\boldsymbol{f}(x,y)
\end{equation*}
where $\vec{r}(x,y)$ is the torque arm of $\dA$.
The frictional wrench $\boldsymbol{w}$ of $\mathcal{A}$ is the vector composed of the frictional force and torque, and hence $\boldsymbol{w} = [f_{x},{f_{y}},\tau_{z}]^T$.
Goyal~et~al.~\cite{goyal.1991} showed that $\tau_{z}$ reaches the maximum when the COR is located at $\boldsymbol{\origin}$ since the torque arm of each element is perpendicular to the frictional force. 
$f_x$ reaches the maximum when the COR is infinitely far away along the $y$-axis.

%% file: sections/7-friction-discrete-surface.tex
\label{subsec:frictionDiscreteSurface}

Computing a 6D frictional wrench for a parametric surface described in \sref{sec:friction_computation} is inefficient due to the integral operation. Therefore, we introduce the frictional wrench computation for a discrete surface, which consists of $N_s$ convex polygonal elements.

Consider the $i$th element with the center $\centeri$ and the normal $\normali$ with $i \in \{1 \ldotsc N_s\}$, we compute the direction vector $\vri$ of the relative velocity at $\centeri$ given a unit twist parametrized as in \eref{eq:unittwist}
\begin{equation}
	\begin{aligned}
	 &\vel_i =  
    \begin{dcases*}
    h\linedir + \moment -\centeri \times \linedir & 
    if $ \normang \neq 0$\\
    \linedir &if $ \normang=0$
    \end{dcases*} \\
	&\vri = \frac{\left(\mat{I} -  \normali \hspace{0.1em} \normali^{\trans}\right)\hspace{0.1em} \vel_i}{\norm{\left(\mat{I} -  \normali \hspace{0.1em} \normali^{\trans}\right)\hspace{0.1em} \vel_i}}.
	\end{aligned}
\end{equation}

Denoting  $\pressurei$ and $\areai$ as the pressure and the area of the $i$th element, respectively, the pressure center $\vec{\origin}$ of the surface is
\begin{equation}
\begin{aligned}
\vec{\origin} = \begin{bmatrix}{\originx}\\{\originy}\\{\originz}\end{bmatrix}= \frac{\sum_ {i=1}^{N_s} \centeri \bigcdot \pressurei \bigcdot \areai}{\sum_ {i=1}^{N_s}  \pressurei \bigcdot \areai}. 
\end{aligned}
\end{equation}

The frictional force and torque of the discrete surface is
\begin{equation}
\begin{aligned}
\vec{f} &= -\mu \sum_{i=1}^{N_s} \pressurei \bigcdot \areai \bigcdot \vri  \\
\vec{\tau} &= -\mu \sum_{i=1}^{N_s}   \pressurei \bigcdot \areai \bigcdot [(\centeri- \vec{\origin}) \times \vri].
\end{aligned}
\label{eq:frictionDiscrete}
\end{equation}



\begin{figure}[t]
	\centering
	{{\includegraphics[width=\linewidth]{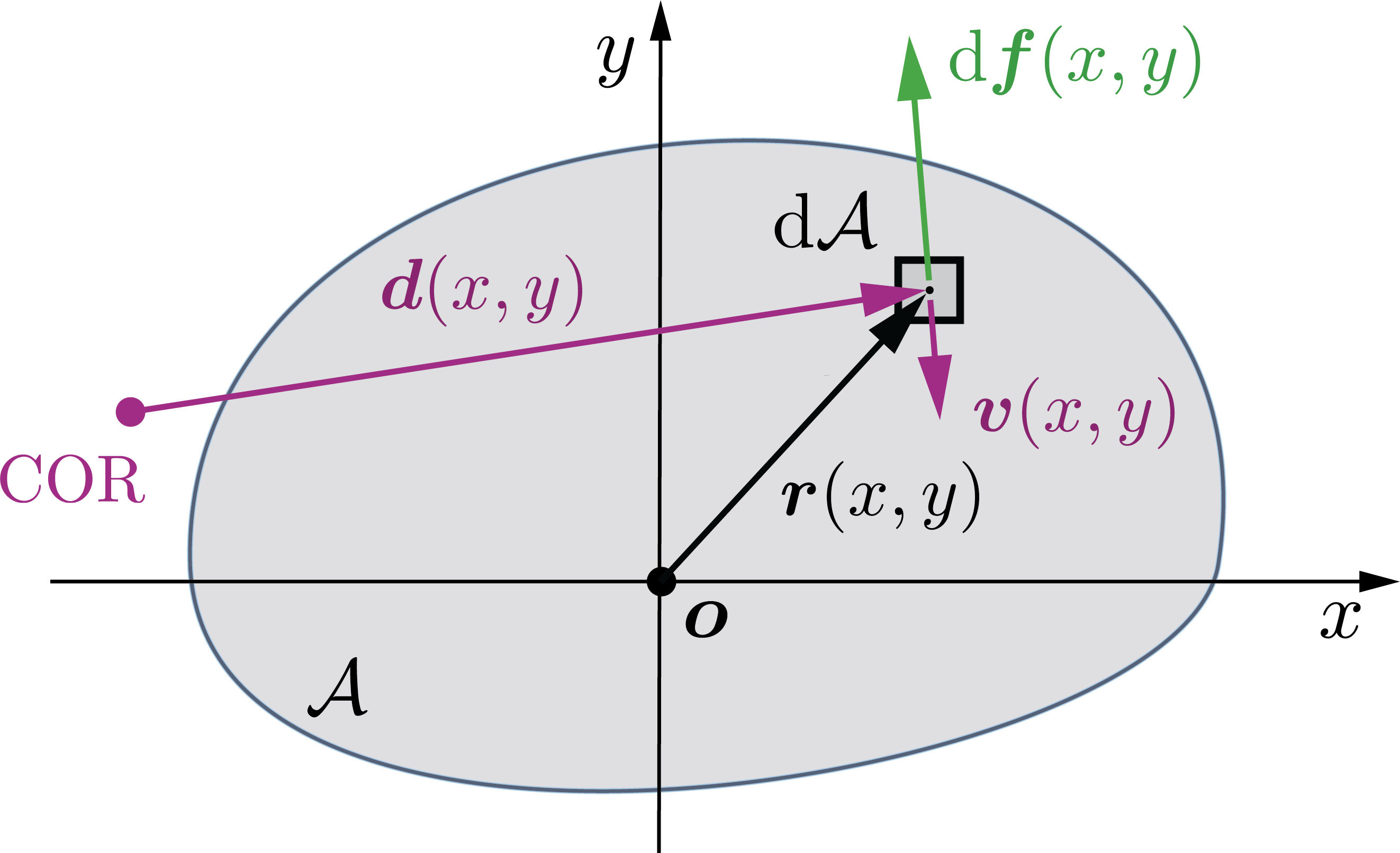} }}%
	\caption{\small Friction computation for a planar contact surface.}
	\label{fig:planar-friction}
\end{figure}

%% file: sections/20-wrench-example.tex
We provided an example in \sref{subsec:units} to compute the contact wrench for the surface shown in \fref{fig:curved-friction}.
We consider an elliptical cylinder with the parametric form $\boldsigma^O(u,v) = \left[0.02\cos u,0.02\sin u,v\right]^\trans,u\in\left[0.25\pi,0.75\pi\right],v\in\left[-0.05,-0.03\right]$ in the object frame $O$.
Given the pressure distribution $p(u,v)=10^3$ and $\mu=0.3$ as inputs, we computed the contact wrench with respect to the friction center in the local contact frame $C$. 
The frictional wrenches in $C$ do not depend on the jaw position relative to the object COM; therefore, can be precomputed and reused.

As the origin of $C$ is located at the pressure center $\origin^O$, we first compute $\origin^O$ in the object frame.
If the origin of $C$ is not located at $\origin^O$, one needs to first compute the equivalent contact wrenches with respect to to the origin of $C$, before transforming the wrenches to $O$.

With $\boldsigma^O(u,v)$, $\pressure(u,v)$, and $\diff \surface = \norm{\boldsigma^O_u \times \boldsigma^O_v}  \diff{u} \diff{v} = 0.02\diff{u} \diff{v}$, we compute $\origin^O$ using \eref{eq:origin} and obtain
\begin{equation*}
\begin{aligned}
    \origin^O &= \frac{\int_{-0.05}^{-0.03}\int_{0.25\pi}^{0.75\pi}10^3 \left[0.02\cos u,0.02\sin u,v\right]^\trans  0.02\dudv}{\int_{-0.05}^{-0.03}\int_{0.25\pi}^{0.75\pi}10^3\bigcdot 0.02\dudv} \\
    &\approx \left[{0},{0.018},{-0.04}\right]^\trans.
\end{aligned}
\end{equation*}
We compute $C$ by shifting $O$ by $ \left[{0},{0.018},{-0.04}\right]^\trans$.
The parametric form in $C$ is $\boldsigma(u,v) = \left[0.02\cos u,0.02\sin u,v\right]^\trans - \left[{0},{0.018},{-0.04}\right]^\trans$, where the ranges of $u$ and $v$ remain unchanged.
The superscript $C$ is omitted as the variables in \sref{subsec:frictionParametricSurface} and \sref{subsec:normal_wrench} are all in the local contact frame $C$.

%% file: sections/10-data-acquisition.tex
\section{Sampling Unit Twists}
\label{appendix:data}
As described in \sref{sec:wrenches_LS}, we sample the unit twists to obtain a finite set of the frictional wrenches that can be transmitted through a nonplanar surface contact. 
The sampling algorithm is introduced in the following. 

For unit twists that contain a rotation, we sample the triplet $\triplet$, where $\linedir$ is uniformly sampled from the unit sphere using the Fibonacci sphere algorithm, $\linepoint$ is randomly sampled within a radius $r$ of the pressure center $\origin$ of the contact, and the pitch $h$ is randomly sampled in the range $[-2,2]$.
The radius $r$ depends on the size of the contact surface.
If $r$ is too large or too small, one obtains very unevenly distributed frictional wrenches.
To determine $r$, we first find the smallest rectangular cuboid that contains the nonplanar surface.
Denoting $l_s$ as the longest side length of the cuboid, we experimentally determine the radius $r$ to be $0.25l_s$, which results in relatively evenly distributed frictional wrenches. 
The sampled $\linedir$ is also used to compute the unit twists that only contain a translation using \eref{eq:unittwist}.

%% file: main.bbl
\begin{thebibliography}{10}
\providecommand{\url}[1]{#1}
\csname url@samestyle\endcsname
\providecommand{\newblock}{\relax}
\providecommand{\bibinfo}[2]{#2}
\providecommand{\BIBentrySTDinterwordspacing}{\spaceskip=0pt\relax}
\providecommand{\BIBentryALTinterwordstretchfactor}{4}
\providecommand{\BIBentryALTinterwordspacing}{\spaceskip=\fontdimen2\font plus
\BIBentryALTinterwordstretchfactor\fontdimen3\font minus
  \fontdimen4\font\relax}
\providecommand{\BIBforeignlanguage}[2]{{%
\expandafter\ifx\csname l@#1\endcsname\relax
\typeout{** WARNING: IEEEtran.bst: No hyphenation pattern has been}%
\typeout{** loaded for the language `#1'. Using the pattern for}%
\typeout{** the default language instead.}%
\else
\language=\csname l@#1\endcsname
\fi
#2}}
\providecommand{\BIBdecl}{\relax}
\BIBdecl

\bibitem{ferrari.1992}
C.~Ferrari and J.~Canny, ``Planning optimal grasps,'' in \emph{{IEEE}
  International Conference on Robotics and Automation (ICRA)}, 1992, pp.
  2290--2295.

\bibitem{bicchi2000robotic}
A.~Bicchi and V.~Kumar, ``Robotic grasping and contact: A review,'' in
  \emph{{IEEE} International Conference on Robotics and Automation (ICRA)},
  2000, pp. 348--353.

\bibitem{okamura2000overview}
A.~M. Okamura, N.~Smaby, and M.~R. Cutkosky, ``An overview of dexterous
  manipulation,'' in \emph{{IEEE} International Conference on Robotics and
  Automation (ICRA)}, 2000, pp. 255--262.

\bibitem{miller.2004}
A.~T. Miller and P.~K. Allen, ``Graspit!: a versatile simulator for robotic
  grasping,'' \emph{{IEEE} Robotics \& Automation Magazine}, vol.~11, no.~4,
  pp. 110--122, 2004.

\bibitem{danielczuk2019reach}
M.~Danielczuk, J.~Xu, J.~Mahler, M.~Matl, N.~Chentanez, and K.~Goldberg,
  ``{REACH}: reducing false negatives in robot grasp planning with a robust
  efficient area contact hypothesis model,'' in \emph{International Symposium
  on Robotics Research (ISRR)}, 2019.

\bibitem{mahler2019learning}
J.~Mahler, M.~Matl, V.~Satish, M.~Danielczuk, B.~DeRose, S.~McKinley, and
  K.~Goldberg, ``Learning ambidextrous robot grasping policies,'' \emph{Science
  Robotics}, vol.~4, no.~26, 2019.

\bibitem{xu2018learning}
J.~Xu, A.~Bhardwaj, G.~Sun, T.~Aykut, N.~Alt, M.~Karimi, and E.~Steinbach,
  ``Learning-based modular task-oriented grasp stability assessment,'' in
  \emph{{IEEE/RSJ} International Conference on Intelligent Robots and Systems
  (IROS)}, 2018, pp. 3468--3475.

\bibitem{wall2017method}
V.~Wall, G.~Z{\"o}ller, and O.~Brock, ``A method for sensorizing soft actuators
  and its application to the {RBO} hand 2,'' in \emph{{IEEE} International
  Conference on Robotics and Automation (ICRA)}, 2017, pp. 4965--4970.

\bibitem{ciocarlie.2005}
M.~Ciocarlie, A.~Miller, and P.~Allen, ``Grasp analysis using deformable
  fingers,'' in \emph{{IEEE/RSJ} International Conference on Intelligent Robots
  and Systems (IROS)}, 2005, pp. 4122--4128.

\bibitem{ciocarlie.2007}
M.~Ciocarlie, C.~Lackner, and P.~Allen, ``Soft finger model with adaptive
  contact geometry for grasping and manipulation tasks,'' in \emph{{IEEE}
  Second Joint EuroHaptics Conference and Symposium on Haptic Interfaces for
  Virtual Environment and Teleoperator Systems}, 2007, pp. 219--224.

\bibitem{tsuji2014grasp}
T.~Tsuji, S.~Uto, K.~Harada, R.~Kurazume, T.~Hasegawa, and K.~Morooka, ``Grasp
  planning for constricted parts of objects approximated with quadric
  surfaces,'' in \emph{{IEEE/RSJ} International Conference on Intelligent
  Robots and Systems (IROS)}, 2014, pp. 2447--2453.

\bibitem{harada2014stability}
K.~Harada, T.~Tsuji, S.~Uto, N.~Yamanobe, K.~Nagata, and K.~Kitagaki,
  ``Stability of soft-finger grasp under gravity,'' in \emph{{IEEE}
  International Conference on Robotics and Automation (ICRA)}, 2014, pp.
  883--888.

\bibitem{goyal.1991}
S.~Goyal, A.~Ruina, and J.~Papadopoulos, ``Planar sliding with dry friction
  {P}art 1. {L}imit surface and moment function,'' \emph{Wear}, vol. 143,
  no.~2, pp. 307--330, 1991.

\bibitem{lee1991fixture}
S.~H. Lee and M.~Cutkosky, ``Fixture planning with friction,'' \emph{Journal of
  Manufacturing Science and Engineering}, vol. 113, no.~3, 1991.

\bibitem{zhou2018convex}
J.~Zhou, M.~T. Mason, R.~Paolini, and D.~Bagnell, ``A convex polynomial model
  for planar sliding mechanics: theory, application, and experimental
  validation,'' \emph{International Journal of Robotics Research (IJRR)},
  vol.~37, no. 2-3, pp. 249--265, 2018.

\bibitem{xu.2017}
J.~Xu, N.~Alt, Z.~Zhang, and E.~Steinbach, ``Grasping posture estimation for a
  two-finger parallel gripper with soft material jaws using a curved contact
  area friction model,'' in \emph{{IEEE} International Conference on Robotics
  and Automation (ICRA)}, 2017, pp. 2253--2260.

\bibitem{rimon_burdick_2019}
E.~Rimon and J.~W. Burdick, \emph{The Mechanics of Robot Grasping}.\hskip 1em
  plus 0.5em minus 0.4em\relax Cambridge University Press, 2019, ch.~4, pp.
  63--90.

\bibitem{kao2008contact}
I.~Kao, K.~Lynch, and J.~W. Burdick, ``Contact modeling and manipulation,'' in
  \emph{Handbook of Robotics}, B.~Siciliano and O.~Khatib, Eds.\hskip 1em plus
  0.5em minus 0.4em\relax Springer, 2008, ch.~27, pp. 647--669.

\bibitem{prattichizzo2016grasping}
D.~Prattichizzo and J.~C. Trinkle, ``Grasping,'' in \emph{Handbook of
  Robotics}, B.~Siciliano and O.~Khatib, Eds.\hskip 1em plus 0.5em minus
  0.4em\relax Springer, 2008, ch.~28, pp. 671--700.

\bibitem{li2001review}
Y.~Li and I.~Kao, ``A review of modeling of soft-contact fingers and stiffness
  control for dextrous manipulation in robotics,'' in \emph{{IEEE}
  International Conference on Robotics and Automation (ICRA)}, 2001, pp.
  3055--3060.

\bibitem{inoue2008mechanics}
T.~Inoue and S.~Hirai, \emph{Mechanics and control of soft-fingered
  manipulation}.\hskip 1em plus 0.5em minus 0.4em\relax Springer, 2008, ch.
  3--6, pp. 19--82.

\bibitem{kerr1986analysis}
J.~Kerr and B.~Roth, ``Analysis of multifingered hands,'' \emph{International
  Journal of Robotics Research (IJRR)}, vol.~4, no.~4, pp. 3--17, 1986.

\bibitem{tsuji2009easy}
T.~Tsuji, K.~Harada, and K.~Kaneko, ``Easy and fast evaluation of grasp
  stability by using ellipsoidal approximation of friction cone,'' in
  \emph{{IEEE/RSJ} International Conference on Intelligent Robots and Systems
  (IROS)}, 2009, pp. 1830--1837.

\bibitem{hertz1882uber}
H.~R. Hertz, ``{\"U}ber die {B}er{\"u}hrung fester elastischer {K}{\"o}rper und
  {\"u}ber die {H}{\"a}rte,'' \emph{Verhandlung des Vereins zur Bef{\"o}rderung
  des Gewerbeflei\ss{}es, Berlin}, 1882.

\bibitem{xydas.1999}
N.~Xydas and I.~Kao, ``Modeling of contact mechanics and friction limit
  surfaces for soft fingers in robotics, with experimental results,''
  \emph{International Journal of Robotics Research (IJRR)}, vol.~18, no.~9, pp.
  941--950, 1999.

\bibitem{tiezzi2007modeling}
P.~Tiezzi and I.~Kao, ``Modeling of viscoelastic contacts and evolution of
  limit surface for robotic contact interface,'' \emph{IEEE Transactions on
  Robotics}, vol.~23, no.~2, pp. 206--217, 2007.

\bibitem{fakhari2016development}
A.~Fakhari, M.~Keshmiri, and I.~Kao, ``Development of realistic pressure
  distribution and friction limit surface for soft-finger contact interface of
  robotic hands,'' \emph{Journal of Intelligent Robotic Systems}, vol.~82,
  no.~1, pp. 39--50, 2016.

\bibitem{arimoto2000dynamics}
S.~Arimoto, P.~T.~A. Nguyen, H.-Y. Han, and Z.~Doulgeri, ``Dynamics and control
  of a set of dual fingers with soft tips,'' \emph{Robotica}, vol.~18, no.~1,
  pp. 71--80, 2000.

\bibitem{arimoto2002stable}
S.~Arimoto, Z.~Doulgeri, P.~T.~A. Nguyen, and J.~Fasoulas, ``Stable pinching by
  a pair of robot fingers with soft tips under the effect of gravity,''
  \emph{Robotica}, vol.~20, no.~3, pp. 241--249, 2002.

\bibitem{johnson.1987}
K.~L. Johnson and K.~L. Johnson, \emph{Contact mechanics}.\hskip 1em plus 0.5em
  minus 0.4em\relax Cambridge university press, 1987.

\bibitem{roa2015grasp}
M.~A. Roa and R.~Su{\'a}rez, ``Grasp quality measures: review and
  performance,'' \emph{Autonomous Robots}, vol.~38, no.~1, pp. 65--88, 2015.

\bibitem{nguyen1988constructing}
V.-D. Nguyen, ``Constructing force-closure grasps,'' \emph{International
  Journal of Robotics Research (IJRR)}, vol.~7, no.~3, pp. 3--16, 1988.

\bibitem{ma.2019}
D.~Ma, E.~Donlon, S.~Dong, and A.~Rodriguez, ``Dense tactile force distribution
  estimation using {GelSlim} and inverse {FEM},'' in \emph{{IEEE} International
  Conference on Robotics and Automation (ICRA)}, 2019, pp. 5418--5424.

\bibitem{romero2020soft}
B.~Romero, F.~Veiga, and E.~Adelson, ``Soft, round, high resolution tactile
  fingertip sensors for dexterous robotic manipulation,'' in \emph{{IEEE}
  International Conference on Robotics and Automation (ICRA)}, 2020, pp.
  4796--4802.

\bibitem{pressley.2010.elementary}
A.~N. Pressley, \emph{Elementary differential geometry}.\hskip 1em plus 0.5em
  minus 0.4em\relax Springer Science \& Business Media, 2010.

\bibitem{murray.1994.mathematical}
R.~M. Murray, Z.~Li, and S.~S. Sastry, \emph{A mathematical introduction to
  robotic manipulation}.\hskip 1em plus 0.5em minus 0.4em\relax CRC Press,
  1994.

\bibitem{everingham2010pascal}
M.~Everingham, L.~Van~Gool, C.~K. Williams, J.~Winn, and A.~Zisserman, ``The
  pascal visual object classes (voc) challenge,'' \emph{International journal
  of computer vision}, vol.~88, no.~2, pp. 303--338, 2010.

\bibitem{howe.1996}
R.~D. Howe and M.~R. Cutkosky, ``Practical force-motion models for sliding
  manipulation,'' \emph{International Journal of Robotics Research (IJRR)},
  vol.~15, no.~6, pp. 557--572, 1996.

\bibitem{mason2001mechanics}
M.~T. Mason, \emph{Mechanics of robotic manipulation}.\hskip 1em plus 0.5em
  minus 0.4em\relax MIT Press, 2001.

\bibitem{kao1992quasistatic}
I.~Kao and M.~R. Cutkosky, ``Quasistatic manipulation with compliance and
  sliding,'' \emph{International Journal of Robotics Research (IJRR)}, vol.~11,
  no.~1, pp. 20--40, 1992.

\bibitem{magnani.2005}
A.~Magnani, S.~Lall, and S.~Boyd, ``Tractable fitting with convex polynomials
  via sum-of-squares,'' in \emph{{IEEE} Conference on Decision and Control},
  2005, pp. 1672--1677.

\bibitem{krug2017grasp}
R.~Krug, Y.~Bekiroglu, and M.~A. Roa, ``Grasp quality evaluation done right:
  How assumed contact force bounds affect wrench-based quality metrics,'' in
  \emph{{IEEE} International Conference on Robotics and Automation (ICRA)},
  2017, pp. 1595--1600.

\bibitem{alt.2016}
N.~Alt, J.~Xu, and E.~Steinbach, ``A dataset of thin-walled deformable objects
  for manipulation planning,'' in \emph{Workshop on Grasping and Manipulation
  Datasets, in conjunction with IEEE International Conference on Robotics and
  Automation ({ICRA})}, 2016.

\bibitem{ansys}
{ANSYS Academic Research Mechanical, Release 17.2}.

\bibitem{festo}
Festo {F}in-{R}ay finger. [Online]. Available:
  \url{https://www.festo.com/group/de/cms/10221.htm}. Last visited: March 2019.

\bibitem{pcl}
R.~B. {Rusu} and S.~{Cousins}, ``3d is here: Point cloud library (pcl),'' in
  \emph{{IEEE} International Conference on Robotics and Automation (ICRA)},
  2011, pp. 1--4.

\bibitem{super4pcs}
\BIBentryALTinterwordspacing
N.~Mellado, D.~Aiger, and N.~J. Mitra, ``{Super4PCS:} fast global pointcloud
  registration via smart indexing,'' \emph{Computer Graphics Forum}, vol.~33,
  no.~5, pp. 205--215, 2014. [Online]. Available:
  \url{http://dx.doi.org/10.1111/cgf.12446}
\BIBentrySTDinterwordspacing

\bibitem{Han2000Grasp}
{L. Han}, J.~C. {Trinkle}, and Z.~{Li}, ``Grasp analysis as linear matrix
  inequality problems,'' \emph{{IEEE} Transactions on Robotics and Automation},
  vol.~16, no.~6, pp. 663--674, 2000.

\bibitem{shan2020modeling}
X.~Shan and L.~Birglen, ``Modeling and analysis of soft robotic fingers using
  the fin ray effect,'' \emph{International Journal of Robotics Research
  (IJRR)}, pp. 1--20, 2020.

\end{thebibliography}
